\documentclass[pdflatex,sn-mathphys]{sn-jnl}

\jyear{2021}
\theoremstyle{thmstyleone}

\theoremstyle{thmstyletwo}%
\theoremstyle{thmstylethree}%
\usepackage{url}
\usepackage{soul}
\usepackage{array}
\usepackage{booktabs}
\usepackage{makecell}
\usepackage[figuresright]{rotating}
\usepackage{longtable}
\usepackage{bm}

\newcommand{\myref}[1]{Eq.(\ref{#1})}

\begin{document}

\title[Article Title]{Artificial Neural Networks for Finger Vein Recognition: A Survey}


\author[1]{Yimin Yin}
\equalcont{These authors contributed equally to this work.}

\author[2]{Renye Zhang}
\equalcont{These authors contributed equally to this work.}

\author[3]{Pengfei Liu}

\author[4]{Wanxia Deng}

\author[1]{Siliang He}

\author[5]{Chen Li}

\author*[6]{Jinghua Zhang}\email{zhangjingh@foxmail.com}

\affil[1]{School of Mathematics and Statistics, Hunan First Normal University, Changsha, 410205, Hunan, China}

\affil[2]{School of Computer Science, Hunan First Normal University, Changsha, 410205, China}

\affil[3]{College of Advanced Interdisciplinary Studies, National University of Defense Technology, Changsha,  410073, Hunan, China}

\affil[4]{College of Meteorology and Oceanography, National University of Defense Technology, 410073, Changsha, China}

\affil[5]{College of Medicine and Biological Information Engineering, Northeastern University, Shenyang,  \postcode{110016}, Liaoning, China}

\affil*[6]{College of Intelligence Science and Technology, National University of Defense Technology, Changsha, 410073, Hunan, China}


\abstract{Finger vein recognition is an emerging biometric recognition technology. Different from the other biometric features on the body surface, the venous vascular tissue of the fingers is buried deep inside the skin. Due to this advantage, finger vein recognition is highly stable and private. They are almost impossible to be stolen and difficult to interfere with by external conditions. Unlike the finger vein recognition methods based on traditional machine learning, the artificial neural network technique, especially deep learning, it without relying on feature engineering and have superior performance. To summarize the development of finger vein recognition based on artificial neural networks, this paper collects 149 related papers. First, we introduce the background of finger vein recognition and the motivation of this survey. Then, the development history of artificial neural networks and the representative networks on finger vein recognition tasks are introduced. The public datasets that are widely used in finger vein recognition are then described. After that, we summarize the related finger vein recognition tasks based on classical neural networks and deep neural networks, respectively. Finally, the challenges and potential development directions in finger vein recognition are discussed. To our best knowledge, this paper is the first comprehensive survey focusing on finger vein recognition based on artificial neural networks.}




\maketitle

\section{Introduction}\label{sec1}

\subsection{Biometric recognition}

\begin{figure}[htbp]
    \centering
    \includegraphics[width = 1.0\textwidth]{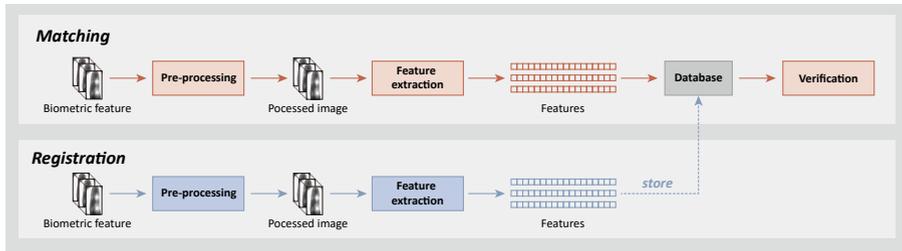}
    \caption{The general workflow of biometric identity verification system.}
    \label{biometric}
\end{figure}

The identity verification system is essential in many fields, such as account login, online payment, and automated teller machines. This technology aims to guarantee the security of user privacy information. The most widely used security mechanism is the classical password. Due to its time-consuming input process, potential leakage risk, and weak anti-attack ability, the classical password is inefficient. However, with the rapid development of information technology, biometric recognition systems are commonly applied in many authentication scenarios~\cite{1}. Biometric recognition aims at identifying a person based on physical and behavioral features, such as face~\cite{5}, voice~\cite{3}, fingerprint~\cite{2}, etc. The general workflow of the biometric identity verification system is shown in Fig.~\ref{biometric}. It usually contains the register and match parts. In the register part, the first step is encoding the original biometric data into feature representations through pre-processing operations, such as image processing techniques~\cite{6}. Then, these features will be registered as prototypes for their corresponding labels in the database. In the match process, the original data is encoded into the feature representation using the same or similar approach as the register part. The feature will be compared with the prototypes in the database to perform the identification task.

\begin{figure}[htbp]
    \centering
    \includegraphics[width = 1.0\textwidth]{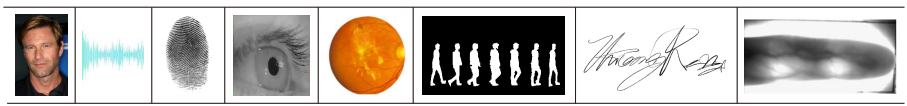}
    \caption{The images of biometric features. These images from left to right are face, voice, fingerprint, iris, retina, gait, signature, and finger vein~\cite{7, 8, 9}.}
    \label{B2}
\end{figure}

Biometric recognition technology is more efficient than the traditional secure identification process due to its convenience and steady security. With the increasing demand for digital security identification systems, biometric recognition demonstrates many advantages. For instance, the biometric recognition system, especially the fingerprint-based system, is widely used and efficient in many home security systems because it can make people released from having to remember passwords or carry keys~\cite{16}. Besides, biometric recognition technology, such as the face, handwriting, and voice recognition, also plays a vital role in crime investigation~\cite{18}. Furthermore, in some financial services, the biometric features, which include finger vein, face, fingerprint, etc., can guarantee that the individual attempting to access financial data is only an authorized user~\cite{19}. In addition to above mentioned biometric features, iris~\cite{4}, retina, and gait~\cite{14} are also widely used. The details of these biometric recognition technologies are provided in Tab.~\ref{Tab.1} to understand their characteristics better. The example image of these biometric features is given in Fig.~\ref{B2}.

\begin{table}[htbp]
    \centering
    \caption{Several biometric recognition technologies~\cite{14, 15, 22}. \textbf{N} represents non-contact. \textbf{C} represents contact. \textbf{RF} represents radio frequency. \textbf{NIR} represents near-infrared. }
    \label{Tab.1}
    \resizebox{\linewidth}{!}{
    \begin{tabular}{lllllcl}
        \toprule
        \textbf{Feature} & \textbf{Security} & \textbf{Obstruction} & \textbf{Data} & \textbf{Device} & \textbf{Contact} & \textbf{Cost} \\
        \hline
        \specialrule{0em}{4pt}{4pt}
        Face & Normal & Illumination & Image & Camera & N & Low \\
        \specialrule{0em}{4pt}{4pt}
        Voice & Normal & Noise & Audio & Microphone & N & Low \\
        \specialrule{0em}{4pt}{4pt}
        Fingerprint & Good & Skin surface & Image & \makecell[l]{Optical sensor \\ Thermal sensor \\ RF sensor} & C & Low \\
        \specialrule{0em}{4pt}{4pt}
        Iris & Superior & Glasses & Image & Special camera & N & High \\
        \specialrule{0em}{4pt}{4pt}
        Retina & Good & Glasses & Image & Special camera & N & Middle \\
        \specialrule{0em}{4pt}{4pt}
        Gait & Normal & \makecell[l]{Personal appearance \\ Filming angle} & \makecell[l]{Video \\ Foot pressure \\ Velocity \\ Frequency} & \makecell[l]{Camera \\ Floor sensor \\ Accelerometer \\ Radar } & N & Low \\
        \specialrule{0em}{4pt}{4pt}
        Signature & Normal & Randomness of writing & \makecell[l]{Image \\ Writing pressure \\ Writing posture} & \makecell[l]{Scanner \\ Electronic tablet \&\\ Electronic pen} & N & Low \\
        \specialrule{0em}{4pt}{4pt}
        Finger vein & Superior & Few & Image & \makecell[l]{NIR sensor \&\\ NIR sensitive camera} & N & Low \\
        \bottomrule
    \end{tabular}}
\end{table}

Although all the biometric features in Tab.~\ref{Tab.1} have particular applications in people's lives, they have different characteristics. Unfortunately, several representative biometric identification technologies are struck by some bottlenecks. For instance, the fingerprint recognition rate is significantly affected by the finger surface, and the fingerprint faces the risk of being forged. Voice recognition usually requires a relatively quiet environment. The recognition rate of iris systems is outstanding, but it requires expensive sensors and may be affected by device location and the environment around the iris. The face data obtained on the user side usually differs from registration in the face recognition system, which may hinder the recognition process~\cite{9}. 

\subsection{Finger vein recognition}
\begin{figure}[htbp]
    \centering
    \includegraphics[width = 1.0\textwidth]{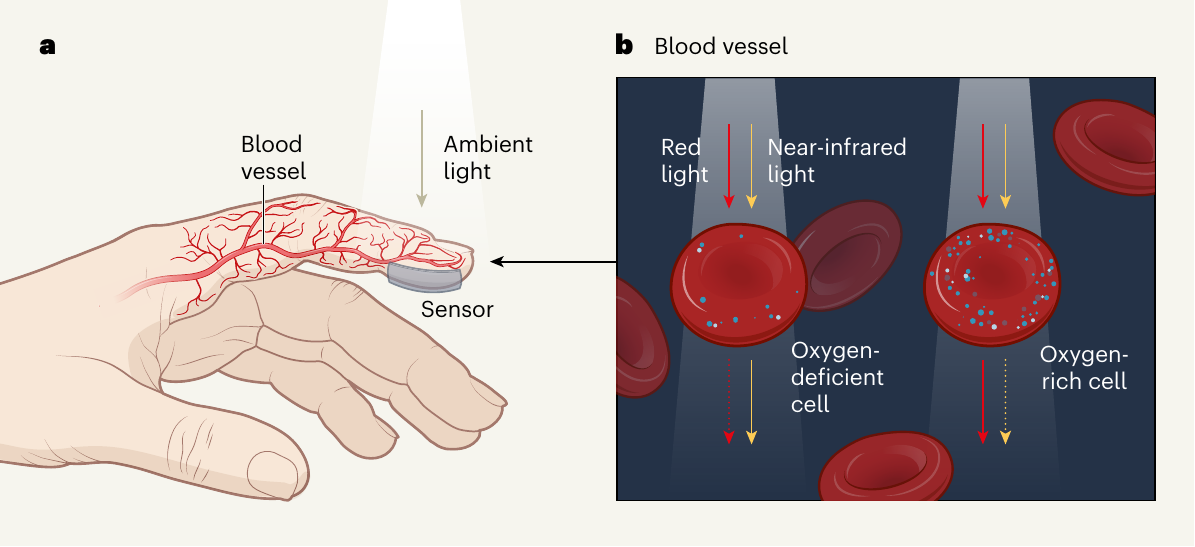}
    \caption{The specific principle of finger vein image capture~\cite{21}: The oxygenated and de-oxygenated Hemoglobin in the venous vessel absorbs the NIR light passing through the finger.}
    \label{Image capture}
\end{figure}

Different from the above biometric features, \emph{Finger Vein Recognition} (FVR) has many advantages since it utilizes the feature extracted from the intrinsic physiological structure of organisms. It matches the vascular feature extracted from the human finger with the previously registered prototypes to perform the recognition task. A Japanese scientist introduced the FVR technology in 2002~\cite{31}. FVR was initially used for medical diagnosis rather than biometric recognition, so early FVR researchers excessively focused on improving the quality of finger vein images, neglecting the practicality of FVR devices in daily life~\cite{31}. In 2006, Hitachi invented the first FVR device for identification~\cite{33}. FVR has been widely used in the biometric recognition field and created a boom in Japan, with financial institutions in hundreds of cities applying FVR to their security systems\cite{30, 32}. Since the finger veins are hidden deep beneath the skin surface, and it is usually observed by the \emph{Near-Infrared} (NIR) light~\cite{20} instead of the visible light~\cite{30}, this characteristic makes FVR-based security systems more private than other biometric-based recognition methods. To better understand the principle of FVR based on NIR, the details are illustrated in Fig.~\ref{Image capture}.

In addition to privacy, FVR also has the characteristic of uniqueness and stability. Even between identical twins, their finger vein structures are different from each other~\cite{26}. Besides, finger veins can maintain their structures with age~\cite{26}. Due to the above-maintained advantages, FVR is challenging to be affected by external factors. Compared with other biometric recognition approaches, finger veins are virtually impossible to be stolen, which brings FVR strong security. In addition, The FVR is more hygienic because its acquisition is non-contact, avoiding public health infections~\cite{26}, and more efficient due to the small size image it processes.

AI technology, especially DL technology, has developed rapidly in recent years. Compared with traditional image processing methods, DL achieves overwhelming performance in many tasks of computer vision, such as biometric recognition~\cite{35}, biomedical image analysis~\cite{36}, and autonomous driving~\cite{37}. The traditional FVR process usually includes image capture, image data pre-processing, feature extraction, and matching or other analysis tasks. The application of DL-based methods, especially \emph{Convolutional Neural Networks} (CNNs), dramatically changes the manual feature extraction process. The performance of conventional \emph{Machine Learning} (ML) approaches is significantly influenced by feature engineering, in which the feature selection is based on human domain knowledge. Nevertheless, CNNs can extract abstract but efficient features by supervised or semi-supervised learning. DL-based methods have highly simplified the recognition process. Due to this significant advantage of DL, DL-based methods are widely used in FVR tasks. For instance, \cite{121} uses GAN to generate data in the FVR task to suppress overfitting. \cite{146} employs a deep \emph{Convolutional Auto-Encoder} (CAE) structure and biohashing algorithm to encrypt the finger vein feature templates. These DL-based approaches have worked innovatively on the FVR tasks and achieved outstanding performance.

\subsection{The motivation about this survey}
The traditional feature extraction methods depend on prior knowledge, and designing a manual feature extraction algorithm for FVR usually requires the knowledge of finger vein anatomy, information coding, and computer vision~\cite{25}. These traditional feature extraction methods are complex and gradually bottlenecked due to the requirement of prior knowledge. Since the widespread use of ANN technology, especially DL, the traditional image feature extraction process has been dramatically changed. ANN-based FVR is attracting attention as a high-performance second-generation biometric technology~\cite{256}. To comprehensively describe the application of ANNs on FVR, this paper reviews classical neural networks and deep neural networks used in FVR. Although there are some existing surveys on FVR, none provide a comprehensive view of the application of ANNs. To figure out our contribution and the difference between our paper and other surveys, we discuss recent surveys on FVR~\cite{32, 25, 46} in the following parts. 

In~\cite{32}, the technology involved in each step of the traditional FVR workflow, such as pre-processing, feature extraction, and matching, is presented. Some traditional ML methods and DL methods for FVR are also discussed. \cite{25} summarizes the feature extraction methods commonly used in FVR, with vein patterns based-methods, dimensionality reduction-based methods, LBP based-methods, image transformations based-methods, and other feature extraction methods. Besides, this survey compares the characteristics between traditional feature extraction methods and feature learning methods. However, the ANN technique mentioned in these papers is not systematic and comprehensive enough, and the amount of literature presented is insufficient. \cite{46} focuses on the software development design, and the hardware development design of FVR is presented. Besides, this paper summarizes the challenges of FVR in several aspects. The solutions to these challenges are also processed based on existing research results. Nevertheless, this paper lacks a summary of related technical papers. These surveys are comprehensive and novel, providing summaries of FVR from different aspects. However, these surveys present a non-negligible drawback. ANN, a critical technology in FVR, is not comprehensively elaborated in these surveys, and the existing FVR surveys lack an overall summary of ANN's wide range of applications in FVR. Inspired by these papers, we have conducted this comprehensive survey of ANN-based FVR. From classical neural networks to deep neural networks, our survey provides a comprehensive summary of ANN applications in FVR.

\begin{figure}[htbp]
    \centering
    \includegraphics[width = 1.0\textwidth]{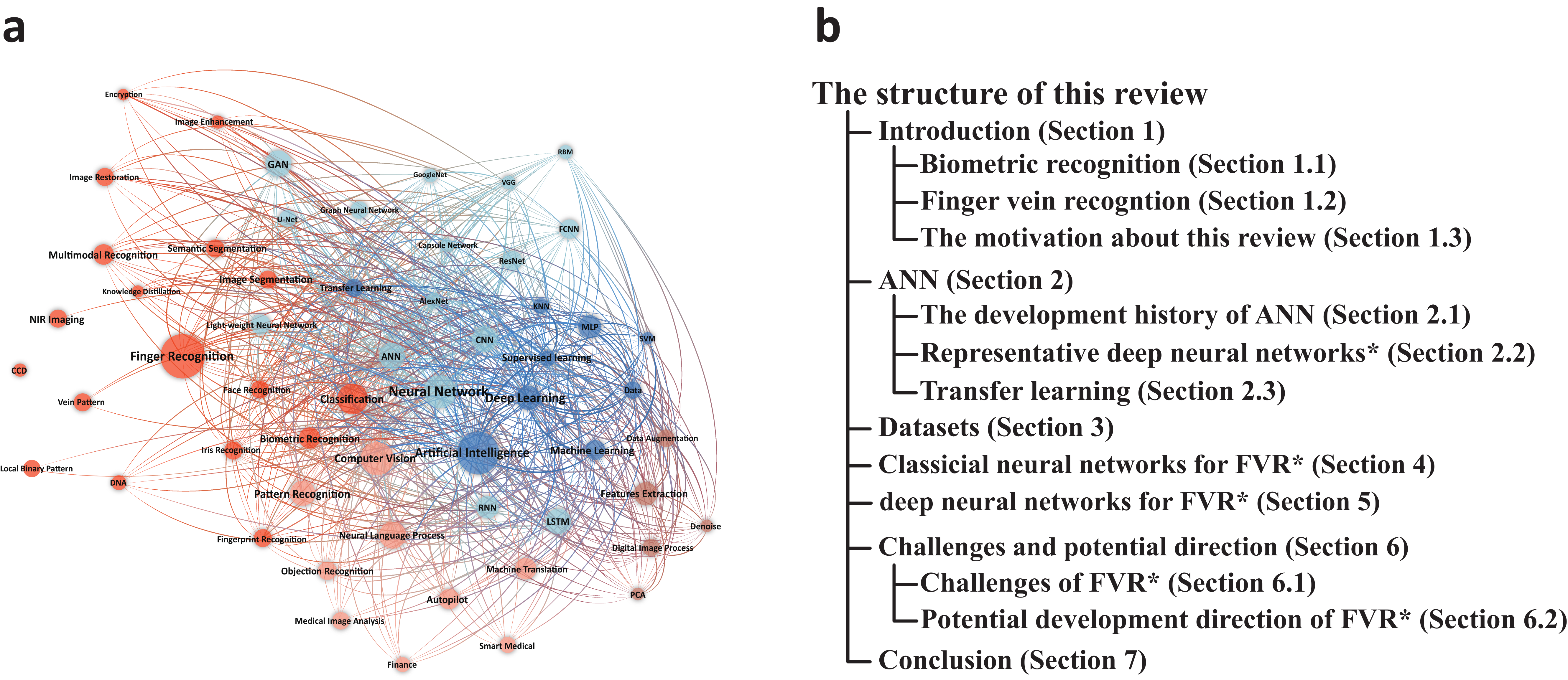}
    \caption{\textbf{(a)} The knowledge graph uses “Artificial Intelligence”, " Artificial Neural Network", "Deep Learning", "Finger Vein Recognition" as keywords, it elaborated on the ANN domain knowledge related to FVR, essentially DL. \textbf{(b)} The structure of this paper. \textbf{*} represents the details will be presented in the corresponding chapter.}
    \label{know}
\end{figure}

To conduct this paper, we summarized 149 papers in the field of ANN-related FVR from 2004 to 2022, covering tasks such as verification, image enhancement, segmentation, \emph{Presentation Attack Detection} (PAD), Template protection of finger vein images. These papers are collected from mainstream academic datasets or search engines, including IEEE Xplore, Springer, Elsevier, ACM, MDPI, World Scientific, and Google Scholar. We use \textbf{“finger vein image analysis” AND (“deep learning” OR “neural network” OR “ANN” OR “CNN” OR “GAN” OR “RNN” OR “LSTM”)} as the searching keywords. FVR is intimately connected with ANN technology, and the related keyword knowledge graph is shown in Fig.~\ref{know}a.

The contributions of this paper are as follows:
\begin{itemize}
    \item To the best of our knowledge, this is the first comprehensive survey summarizing the application of ANNs, including classical neural networks and deep neural networks in FVR. To conduct this survey, we discuss 149 relevant papers. In addition, we briefly describe the history of ANN development and summarize the commonly used public datasets in the FVR domain.
    \item We divide the involved ANNs into two types, classical neural networks and deep neural networks. In classical neural networks, we follow the traditional image process of the biometric recognition paradigm for analysis, including pre-processing, feature extraction, and matching.
    \item In deep neural networks, we follow the different image analysis tasks for summary, including verification, image enhancement, segmentation,  PAD, Template protection, multimodal biometric recognition, image quality assessment, feature extraction, and ROI extraction. 
    \item We have compiled some of the challenges encountered in FVR and provided potential directions for the development of FVR.
\end{itemize}

The structure of this survey is as follows: In Sec.~\ref{sec2}, we introduce the development history of ANN and highlight its groundbreaking achievements. We also introduce representative deep neural networks and transfer learning. The public datasets widely used for FVR are illustrated in Sec.~\ref{sec3}. In Sec.~\ref{sec4}, we summarize the application of classical neural networks on FVR according to the recognition workflow. Sec.~\ref{sec5} presents the summary of tasks of deep neural networks on FVR according to the different tasks. In Sec.~\ref{sec6}, we outline the potential directions of FVR. Sec.~\ref{sec7} summarized the entire paper. The specific content of each section in this survey is shown in Fig.~\ref{know}b. 

\section{ANN}\label{sec2}
\begin{figure}[htbp]
    \centering
    \includegraphics[width = 1.0\textwidth]{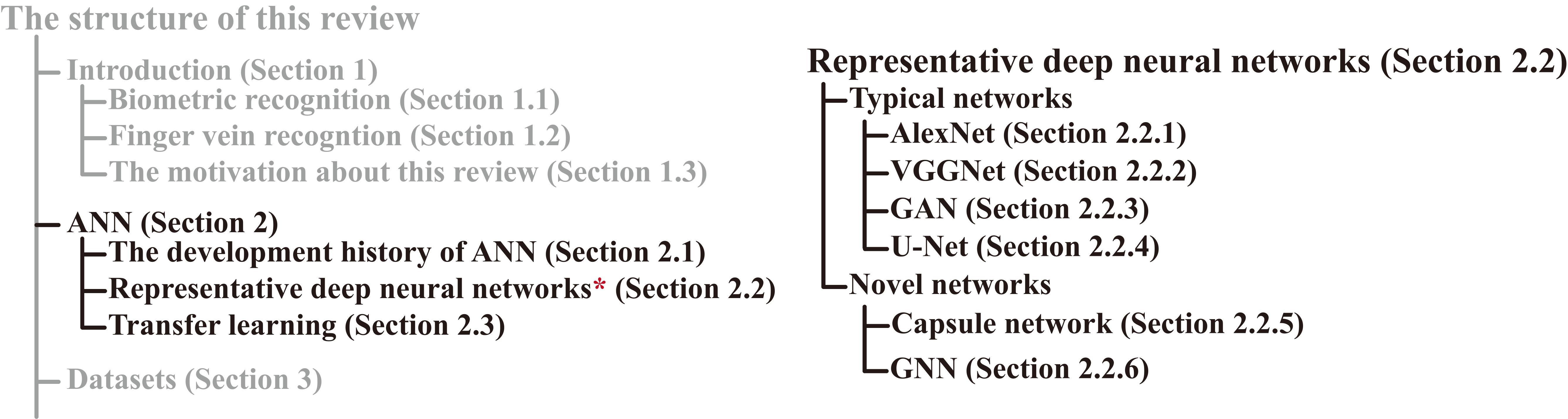}
    \caption{This section structure.}
    \label{S2}
\end{figure}

Since our work focuses on applying ANN to FVR, we briefly describe the development history and breakthrough discoveries of ANN and introduce the representative deep neural network structures widely used in FVR. At the same time, transfer learning, which is widely used in FVR, is also included in our discussion. The specific structure of this section is shown in Fig.~\ref{S2}.

\subsection{The development history of ANN}
\begin{figure}[htbp]
    \centering
    \includegraphics[width = 0.4\textwidth]{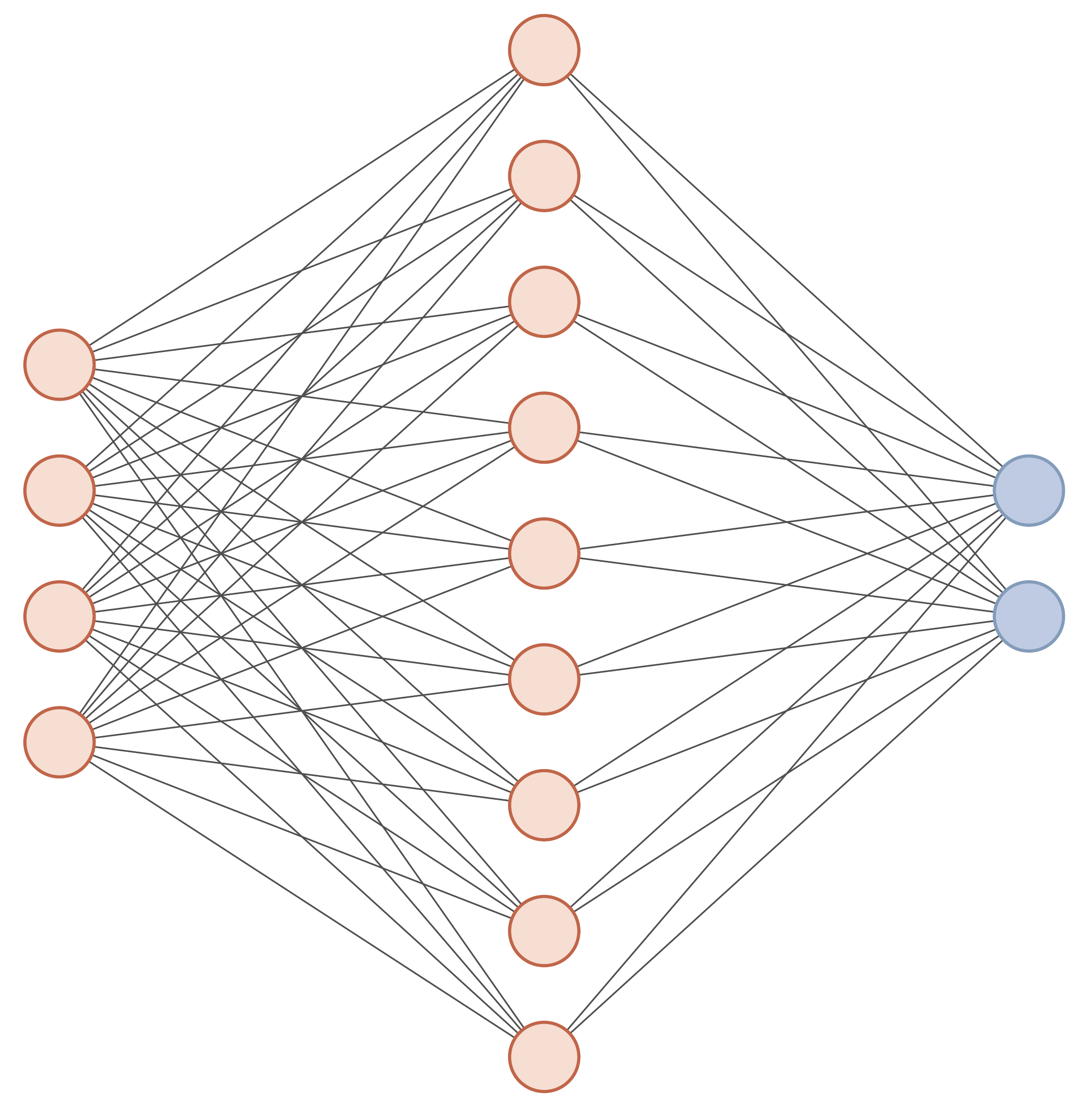}
    \caption{Typical ANN structure represented by MLP.}
    \label{MLP}
\end{figure}

ANN is a computational model designed to mimic biological neural networks. The structure of basic ANN, MLP, is shown in Fig.~\ref{MLP}. ANN consists of multiple nodes interconnected with each other, and nodes between two adjacent layers are given different connection weights to extract features from the original information. Each node in the ANN performs forward propagation by receiving the output from one or more nodes of the previous layer. ANN uses the interconnection between nodes to perform mathematical modeling to solve complex problems. 

\begin{figure}[htbp]
    \centering
    \includegraphics[width = 1.0\textwidth]{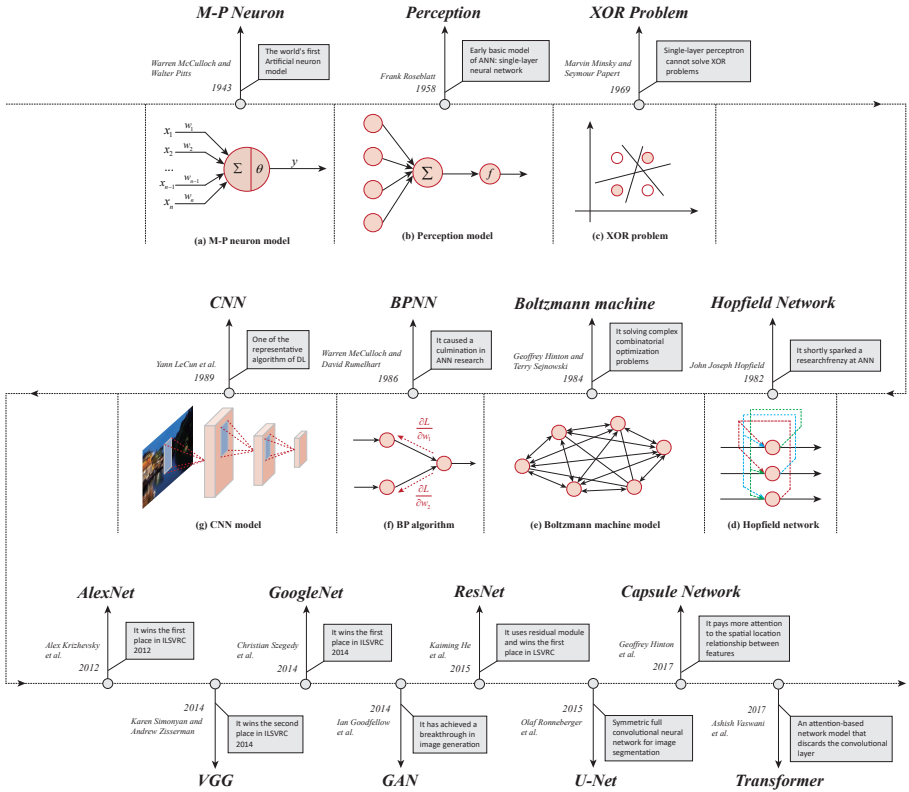}
    \caption{Representative results of ANN's development history.}
    \label{history}
\end{figure}

Nowadays, ANN has achieved great success on many tasks. However, these achievements have been obtained by many AI scientists through continuous research for up to half a century. To finger out the developing process of ANNs, we provide an overview of it in Fig.~\ref{history}.

\begin{figure}[htbp]
    \centering
    \includegraphics[width = 0.8\textwidth]{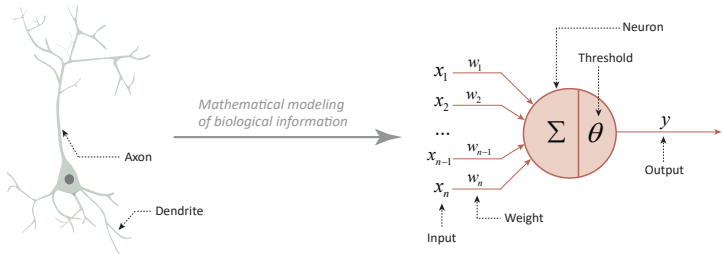}
    \caption{M-P neurons are modeled according to the physiological structure of the biological neurons, especially the function of dendrites and axons on biological neurons. Dendrites can receive stimuli and transmit excitation to the cell body, and axons can transmit their excitation to other neurons. This constitutes the basic function of artificial neurons.}
    \label{MP}
\end{figure}

The starting date of ANN research can be traced back to the 1940s. In 1943, McCulloch and Pitts were inspired by biological neurons to propose the first artificial neuron model, the M-P model,~\cite{55}. More details are shown in Fig.~\ref{MP}. This model is implemented through a physical network, consisting of resistors and other elements. It can be used to perform simple logical operations. Based on the M-P neuron. Roseblatt proposed the perception in 1958~\cite{56}. This model is capable of determining the weights between neuronal connections after training. Perception led to the first boom in ANN research. However, in 1969, Minsky and Papert pointed out that perception could not solve the linear indivisibility problem, XOR problem~\cite{57}. This finding brought ANN's research to a low ebb.

In 1983, Hopfield created a stir in the ANN field by proposing the \emph{Hopfield Neural Network} (HNN) that used associative memory~\cite{58}. HNN achieved the best result on the traveling salesman problem~\cite{263}. HNN has brought the previously lukewarm ANN back into the limelight of scientists. After that, Hinton proposed the famous Boltzmann machine, a randomized HNN~\cite{59}. However, the emergence of \emph{Back Propagation} (BP) algorithms led to the second wave of ANN research. In 1986, McCulloch and Rumelhart proposed the BP algorithm~\cite{60} and applied it to \emph{Multi-layer Perceptron} (MLP) that can solve the XOR problem. In 1989, LeCun et al. introduced convolutional layers into ANN by the biological primary visual cortex. They also introduced the BP algorithm into the network and achieved great success in \emph{Handwritten Digit Recognition} (HDR) tasks~\cite{51, 61}. The BP algorithm is one of the most successful and fundamental ANN algorithms. Even now, it is still an essential element in the training process of ANN.

Although ANN can model complicated patterns and forecast issues by increasing the number of hidden layers and neurons, the computer performance at that time was so limited that it was almost impossible to train a large-scale ANN. Additionally, the overfitting problem also hindered the development of ANN. With the popularity of ML methods, especially \emph{Support Vector Machine} (SVM)~\cite{63}, the research related to ANN gradually fell into a depression for the second time.

Despite the dilemma of ANN research, Hinton and Bengio et al. still focused on ANN research~\cite{64, 65, 66, 67, 68, 69, 70, 71, 72, 73, 52}. In 2006, \cite{66} proposed a \emph{Deep Belief Network} (DBN), which employed the Layer-wise pre-training strategy. After the pre-training, the weights in DBN were fine-tuned with the BP algorithm. This approach made it possible to train a deep ANN at that time. Besides, The deep neural network based on the pre-training strategy achieved significant breakthroughs in speech recognition tasks~\cite{73}. Meanwhile, the remarkable AlexNet designed by Hinton and Krizhevsky~\cite{52} won first place in ImageNet ILSVRC-2012. Benefiting from these outstanding achievements, the ANN technologies represented by DL attracted researchers' attention again. Recently, with the significant improvement in computer performance, the training cost of deep neural networks has decreased. In addition, the amount of data available for network learning has increased greatly compared to before, which can well avoid the problem of overfitting caused by limited data. With the above progress, ANN has once again reached a climax.

\subsection{Representative deep neural networks}
\begin{figure}[htbp]
    \centering
    \includegraphics[width = 1.0\textwidth]{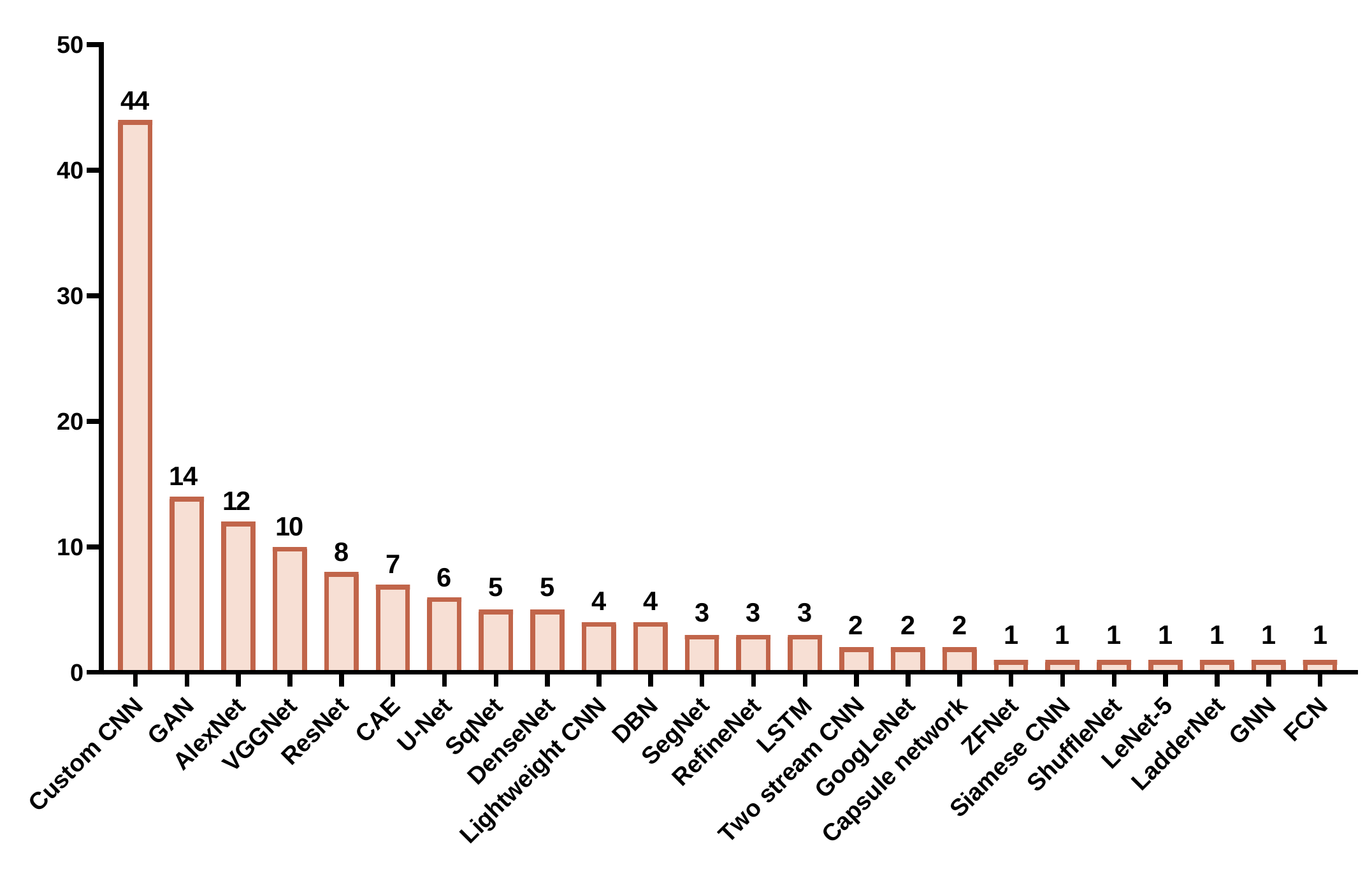}
    \caption{The frequency of deep neural networks on FVR tasks.}
    \label{EDNN}
\end{figure}

As mentioned above, training a complex neural network is no longer difficult due to the development of computer hardware and ANN technology. To figure out the application trend of deep neural networks in FVR, we perform frequency statistics based on the literature we summarized. The details are provided in Fig.~\ref{EDNN}. To better understand these popular networks, their characteristics are briefly summarized. Besides, since U-Net is widely used in the finger vein image segmentation tasks, we also introduce the stricture of U-Net. Some novel networks that have great potential to be applied to FVR tasks are also introduced in this section, including the capsule network and \emph{Graph neural network} (GNN).

\subsubsection{AlexNet}
\begin{figure}[htbp]
    \centering
    \includegraphics[width = 1.0\textwidth]{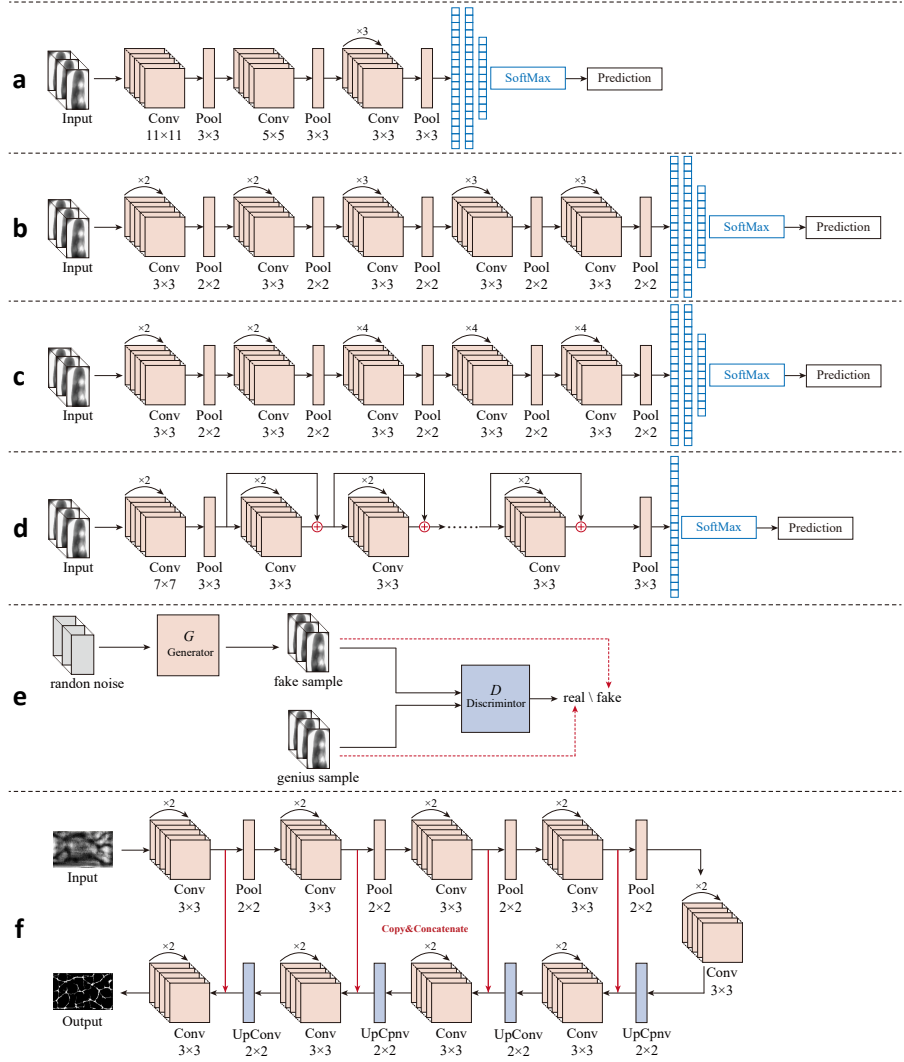}
    \caption{The structure of representative networks. \textbf{(a)} AlexNet. \textbf{(b)} VGG-16. \textbf{(c)} VGG-19. \textbf{(d)} ResNet. \textbf{(e)} GAN. \textbf{(f)} U-Net.}
    \label{DNN}
\end{figure}

AlexNet is a milestone network because it is the first CNN that won first place in the ILSVRC 2012. Before it, the development of neural network technology was at a low ebb for many years. Since the success of AlexNet, deep CNNs have become the mainstream technology in many computer vision tasks~\cite{78}. AlexNet has 600 million parameters and 650,000 neurons. As the structure of AlexNet shown in Fig.~\ref{DNN}, it contains five convolutional layers and three fully connected layers with 4096, 4096, and 1000 neurons, respectively~\cite{52}. AlexNet uses overlapping pooling instead of convolutional pooling and uses the \emph{Rectifed Linear Unit} (ReLU) as the activation function because it is faster in gradient descent than a saturated nonlinear function. The equation of ReLU is formulated in \myref{ReLU}. Additionally, AlexNet was trained on multi-GPUs, since the performance of GPU was limited at that time~\cite{52}. To avoid overfitting, AlexNet utilized data augmentation and dropout operations. 

\begin{equation}\label{ReLU}
    f(x)=\left\{\begin{matrix}
    0\quad x<0\\
    x\quad x\ge 0
\end{matrix}\right.
\end{equation}

\subsubsection{VGGNet}
Compared with AlexNet, the significant contribution of VGGNet is only using $3\times3$ convolutional kernels instead of large convolutional kernels used in AlexNet to compose the network structure. This innovation reduces the parameters while enhancing the nonlinear fitting ability of the network. There are various structures of VGGNet in the earliest research of~\cite{53}. Among them, VGG-16 and VGG-19 are the most widely used. The detailed structures are shown in Fig.~\ref{DNN}, VGG-16 consists of 13 convolutional layers, five max-pooling layers, and three fully connected layers. VGG-19 has three more convolutional layers. There have 138 million and 144 million parameters in VGG-16 and VGG-19, respectively~\cite{53}.

\subsubsection{ResNet}
In theory, the deeper the network structure is, the more rewarding the performance has. This is because deep CNNs are capable of extracting more efficient features. However, a too deep network structure is prone to lead to the problem of gradient explosion or vanishing~\cite{82}. To overcome this problem, \cite{76} proposed \emph{Residual Network} (ResNet) by introducing residual learning in 2015, which won first place in ILSVRC 2015. As shown in Fig.~\ref{DNN}, ResNet improves the efficiency of information propagation by adding shortcut connections between the convolutional layers. The combination structure is called the residual unit, and ResNet is a deep neural network composed of several residual units.

In the residual unit, the input can be denoted as $x$, and the feature map learned for $x$ by the residual unit can be represented by $H(x)$. As shown in~\myref{ResEq}, $H(x)$ is split into $x$ and $H(x)-x$, which called identity mapping and residual mapping respectively. Since the $x$ is invariant, the learning target of the network changes from $H(x)$ to the residual mapping $F(x)=H(x)-x$. Benefiting identity mapping, ResNet can build the network deeper than other networks without gradient vanishing during gradient backpropagation. This deep network structure allows ResNet to learn more useful features than shallow networks, which usually promote better performance.
\begin{equation}\label{ResEq}
    H(x)=\underbrace{x}_{Identity\;mapping} + \underbrace{[H(x)-(x)]}_{Residue\;mapping}
\end{equation}

\subsubsection{Generative adversarial network}
\emph{Generative adversarial network} (GAN) proposed by~\cite{74} is the hottest DL algorithm in the image generation task. The specific workflow of the algorithm is shown in Fig.~\ref{DNN}. Unlike the neural networks used for classification, GAN is composed of a generator and a discriminator. The purpose of the generator is to generate fake samples that can fool the discriminator. The task of the discriminator is to distinguish samples between true and fake.

In the training process of GAN, the generator is first initialized by random noise. Then, the samples obtained by the generator are combined with the real samples to train the discriminator to distinguish samples between true and fake. The predicted probability of the discriminator is then used to promote the updating of the generator by the loss function. After sufficient training rounds, when the generator can fool the discriminator, the network parameters will be fixed and used to generate the image. The generative adversarial process can be modeled in the form of~\myref{gan}. $V$ is the objective function of the entire model. $D$ is discriminator, $G$ is generator, $E_{x\sim P_{data}}$ represents the true data distribution, $E_{z\sim P_{z}(z)}$ represents random noise distribution. The entire formula reveals the GAN optimization process.

\begin{equation}\label{gan}
    \underset{G}{min}\;\underset{D}{max}V(G,D)=E_{x\sim P_{data}}[logD(x)]+E_{z\sim P_{z}(z)]}[log[1-D(G(z))]
\end{equation}

\subsubsection{U-Net}
U-Net is a popular fully convolutional network for image segmentation, initially used for biomedical image segmentation. As shown in Fig.~\ref{DNN}, the network structure of U-Net is symmetrical and includes a compressed path to extract features and an extensive path to perform up-sampling. This network structure can achieve precise segmentation with few images since it takes advantage of data augmentation.

\subsubsection{Capsule network}
CNNs learn global information by deepening the number of layers. However, they cannot capture spatial information. This limitation hinders the network from considering the feature-to-feature interrelationships and positional relationships. Meanwhile, pooling operations widely used in CNNs are susceptible to information loss. Based on these shortcomings of CNNs, \cite{264} proposed the capsule network that is a more effective image processing network. The capsule network uses capsules instead of ordinary artificial neurons to store feature information. A capsule is a carrier that contains multiple neurons and is expressed as a vector. The mode value of the capsule vector indicates the degree to which the capsule recognizes the features, and the direction indicates the posture information of the feature. The parameters of the capsule vector can reflect the variation of feature locations. The capsule network can store spatial information of the features. In addition, the capsule network uses the dynamic routing algorithm to train the weights among the capsules rather than the BP algorithm. The algorithm difference between capsule network and traditional CNN is shown in Tab.~\ref{DB}.

\begin{table}[htbp]
\centering
\caption{Difference between capsule network and CNN~\cite{264}.}
\label{DB}
\renewcommand\arraystretch{1.2}
\begin{tabular}{ccccll}
\cline{1-4}
                                                     & \multicolumn{1}{c}{~}                                  & Capsule network & CNN    &  &  \\ \cline{1-4}
\multicolumn{2}{c}{Input from low-level}  & vector $\bm{u}_i$        & scalar $x_i$ &  &  \\  \cline{1-4}
\multicolumn{1}{c|}{\multirow{3}{*}{Operation}}      & \multicolumn{1}{c}{Transform}               & ${\hat{\bm u}}_{j\mid i} = \bm{W}_{ij}\bm{u}_i$ & \textbf{-}    &  &  \\
\multicolumn{1}{c|}{}                                & \multicolumn{1}{c}{Weighted summation}      & $\bm{s}_j=\sum\nolimits_i {{c_{ij}{\hat{\bm u}}_{j\mid i}}}$ & ${a_j} = \sum\nolimits_i {{w_i}{x_i} + b} $    &  &  \\
\multicolumn{1}{c|}{}                                & \multicolumn{1}{c}{Activation}              & ${\bm{v}_j} = \frac{{{{\left\| {{\bm{s}_j}} \right\|}^2}}}{{1 + {{\left\| {{\bm{s}_j}} \right\|}^2}}}\frac{{{\bm{s}_j}}}{{\left\| {{\bm{s}_j}} \right\|}}$             & $h_j=f(a_j)$    &  &  \\ \cline{1-4}
\multicolumn{2}{c}{Output}                                                                         & vector $\bm{v}_i$         & scalar $h_i$ &  &  \\ \cline{1-4}
\end{tabular}
\end{table}

\subsubsection{Graph neural network}
As a data structure consisting of vertices and edges, the graph can store rich information and represent relationships between information~\cite{265}. To learn the graph structure, GNN is proposed as a variant of CNN~\cite{266}. The structure of GNN is defined by the node and connection matrix. GNN represents a new node by aggregating adjacent feature vectors and extends the traditional convolutional layers to the non-Euclidean space to capture the information in a graph structure. Finger vein features can be considered a graph structure because of the interlocking venous vascular lines embedded in images.

\subsection{Transfer learning}
Since most ML methods require the training and testing data from the same label space~\cite{91}, it usually needs to reconstruct a proprietary dataset when the learning model needs to perform different tasks. However, re-collecting and labeling the data in many real application scenarios is time-consuming and costly. However, Semi-supervised learning~\cite{93} can partly solve this problem by relaxing the annotation requirements of the data. It is sometimes challenging to collect enabled raw samples in large quantities. To reduce the cost of reconstructing datasets from scratch, transfer learning is widely used in various tasks, including FVR.

\begin{figure}[htbp]
    \centering
    \includegraphics[width = 0.65\textwidth]{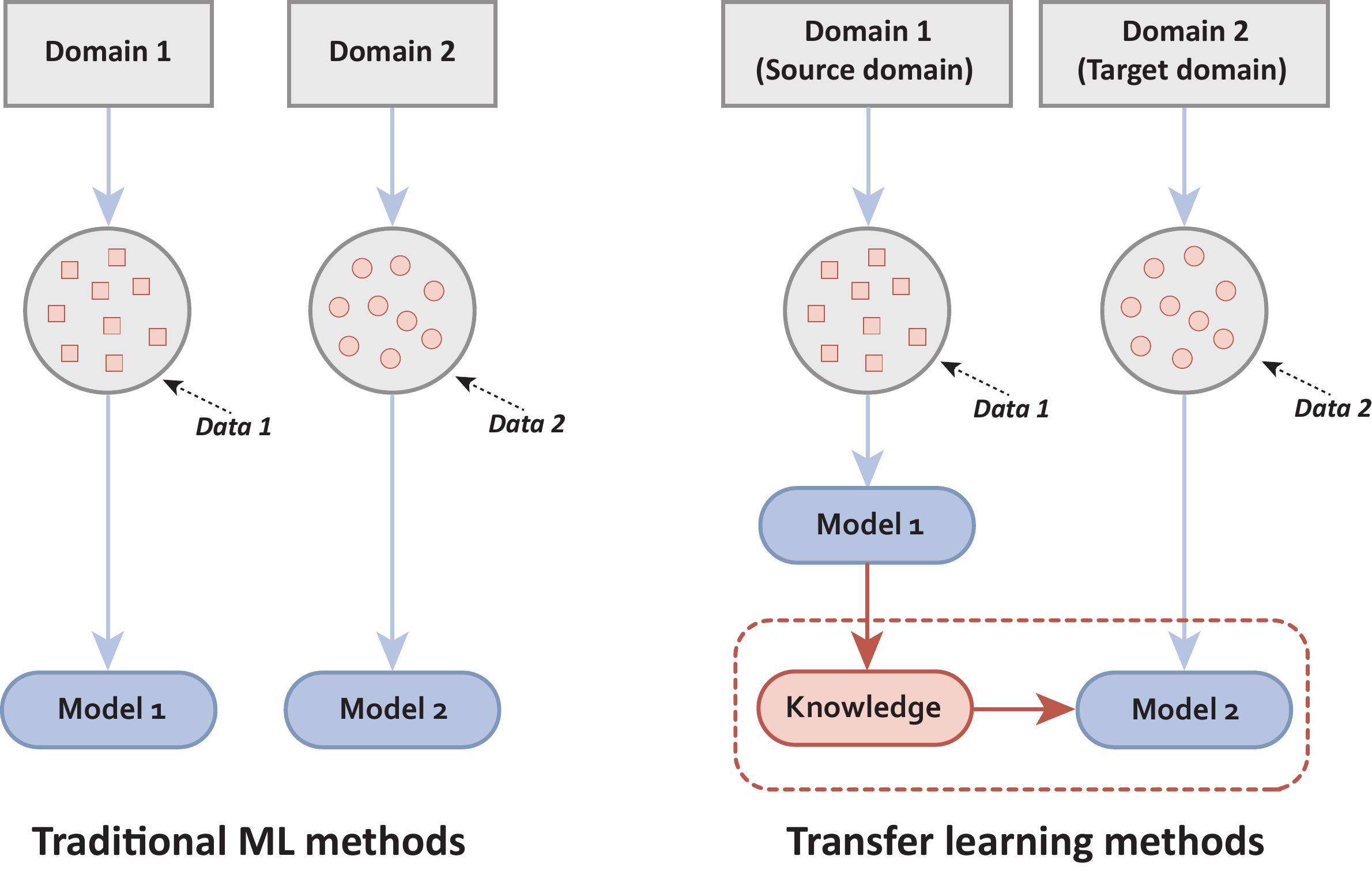}
    \caption{Transfer learning method.}
    \label{Transfer}
\end{figure}

In transfer learning, the training and test samples are called the source and target domain, and transfer learning resolves target tasks by transferring knowledge from the source domain to the target domain, thereby reducing the cost of reconstructing data for new tasks. Transfer learning in the ML domain is proposed by the knowledge transfer in the human knowledge domain. For instance, someone who has studied piano may have an easier time learning violin, and someone who can ride a bicycle will seem more proficient on a motorcycle. As shown in Fig.~\ref{Transfer}, the goal of transfer learning is to utilize knowledge from source domains to increase the learning performance of the target task by minimizing the number of labeled examples required for the target domain. Due to the insufficient number of samples on the public datasets in the FVR domain, transfer learning has been used extensively in the process of model training of FVR tasks. Pre-training network module such as AlexNet and VGGNet on ImageNet~\cite{95} before performing the FVR significantly improves the training process of the models. 

\section{Datasets}
\label{sec3}

As DL occupies an important position in the research field of AI, it is also widely used in FVR task~\cite{77}. DL is a data-driven learning paradigm that aims to learn effective features based on abundant training data to perform the analysis task. Therefore, the dataset plays an essential role in the development of DL. In this section, we introduce the commonly used datasets in FVR.
By investigating relevant papers, we conduct a statistic to summarize the usage frequency of various FVR datasets. The details are demonstrated in Fig.~\ref{data}. It can be found that the popular datasets for FVR mainly include SDUMLA-HMT~\cite{84}, FV-USM~\cite{87}, HKPU~\cite{85}, MMCBNU-6000~\cite{86}, UTFVP~\cite{96, 97}, THU-FVFDT~\cite{98, 89}, SCUT~\cite{131}, and IDIAP~\cite{119}. The fundamental information of these datasets is provided as follows:

\begin{figure}[htbp]
    \centering
    \includegraphics[width = 1.0\textwidth]{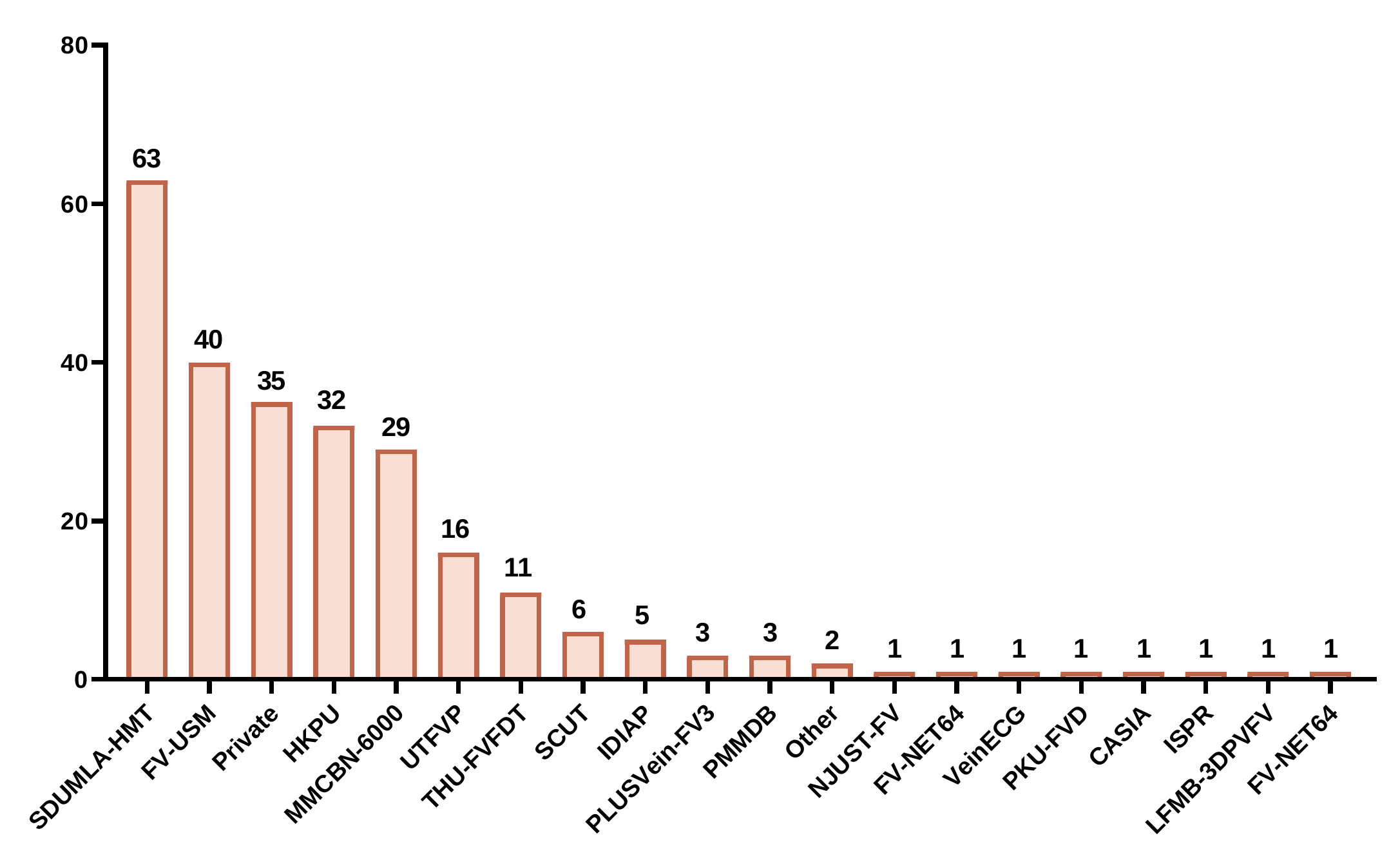}
    \caption{The frequency of all datasets on FVR tasks.}
    \label{data}
\end{figure}

\textbf{SDUMLA-HMT} is a homologous multi-modal traits database containing multiple biometric features such as face, finger veins, gait, iris, and fingerprints. The finger vein part of SDUMLA-HMT is the first publicly available finger vein dataset, consisting of 3816 images. These images were collected from each of the six fingers of 106 people, and six images were collected from each finger.

\textbf{FV-USM} contains 5904 images obtained from 123 volunteers, including 93 males and 40 females, ranging in age from 20 to 50. The image collection process was divided into two stages. The time gap between these two stages is more than two weeks. Each person provided four fingers for image capture. For each image collection stage, six images were taken for each finger.

\textbf{HKPU} contains 6264 images acquired from 156 subjects. Half of these images are finger vein images, and the rest are finger texture images. 93\% of the subjects are younger than 30 years old. Images were acquired in two separate sessions with a minimum interval of one month and a maximum interval of six months. The average interval is 66.8 days. In each session, every subject provided six samples. Each sample contains one vein image and one finger texture image.

\textbf{MMCBNU-6000} contains 6000 finger vein images collected from 100 volunteers from 20 different countries. These volunteers have different skin tones. Each subject provided their index finger, middle finger, and ring finger, and each finger was photographed ten times in an office environment (rather than a dark environment).

\textbf{UTFVP} contains 1440 vascular pattern images obtained from 60 volunteers. These images were captured in two sessions. The average time gap between these sessions is 15 days. The vascular pattern of the six fingers from each subject was taken two times.

\textbf{THU-FVFDT} contains two versions. The first version, THU-FVFDT1, contains 440 finger vein images from 220 subjects. The second version, THU-FVFDT2, contains 2440 finger vein and finger dorsal texture images from 610 subjects. Both datasets were acquired with only one finger of each subject, and their image acquisition process was finished in two sessions.

\textbf{SCUT} contains 10800 images acquired from 100 subjects. Each subject provided six fingers, and each finger was photographed 18 times. For each finger, the first six images were taken in a normal posture, and the last 12 images were taken at a rotational angle of less than $20\,^{\circ}$.

\textbf{IDIAP} consists of 880 cropped and full versions of real and faked images from 110 subjects, and these subjects were from different races. Half of these images are real acquisitions, and half are fake. The fake images are created based on some images of the VERA dataset after the simple pre-processing. This dataset is mainly used for PAD.

To summarizes the above-mentioned datasets, the crucial information of these datasets is provided in Tab.~\ref{dataset}. 

\begin{table}[htbp]
    \renewcommand\arraystretch{1.6}
    \centering
    \caption{Detail of public datasets that widely used in FVR. \textbf{SN} represents subject number. \textbf{IN} represents image number. \textbf{M} represents middle finger. \textbf{I} represents index finger. \textbf{R} represents ring finger. \textbf{No. FS} represents number of finger for each subject.}
    \label{dataset}
    \resizebox{\linewidth}{!}{
    \begin{tabular}{ccclcc}
        \toprule
        \textbf{Dataset}  & \textbf{SN} & \textbf{IN} & \textbf{No. FS} &  \textbf{Resolution} & \textbf{URL} \\
        \hline
        SDUMLA-HMT~\cite{84} & 106 & 3816 & 6 (both M, I, R) & $320\times240$ & http://mla.sdu.edu.cn/info/1006/1195.htm \\
        FV-USM~\cite{87} & 123 & 5904 & 4 (both M, I) & $640\times480$ & http://drfendi.com/fv\_usm\_database/\\
        HKPU~\cite{85} & 156 & 6264 & 2 (left M, R) & $513\times256$ & http://www4.comp.polyu.edu.hk/~csajaykr/fvdatabase.htm \\
        MMCBNU-6000~\cite{86} & 100 & 6000 & 6 (both M, I, R) & $640\times480$ & http://multilab.jbnu.ac.kr/MMCBNU\_6000 \\
        UTFVP~\cite{97} & 60 & 1440 & 6 (both M, I, R) & $672\times380$ & https://pythonhosted.org/bob.db.utfvp/ \\
        THU-FVFDT1~\cite{98} & 220 & 440 & 1 (left I) & $200\times100$  & https://www.sigs.tsinghua.edu.cn/labs/vipl/thu-fvfdt.html\\
        THU-FVFDT2~\cite{89} & 610 & 2440 & 1 (left I) & $200\times100$ & https://www.sigs.tsinghua.edu.cn/labs/vipl/thu-fvfdt.html\\
        SCUT~\cite{131} & 100 & 10800 & 6 (both M, I, R) & $640\times480$ & https://github.com/SCUT-BIP-Lab/SCUT-RIFV \\
        IDIAP~\cite{119} & 110 & 440 & 4 & $665\times250$ & https://www.idiap.ch/dataset/vera-fingervein/index\_html\\
        \bottomrule
    \end{tabular}}

\end{table}

\section{Classical neural network for finger vein recognition}\label{sec4}
\begin{figure}[htbp]
    \centering
    \includegraphics[width = 1.0\textwidth]{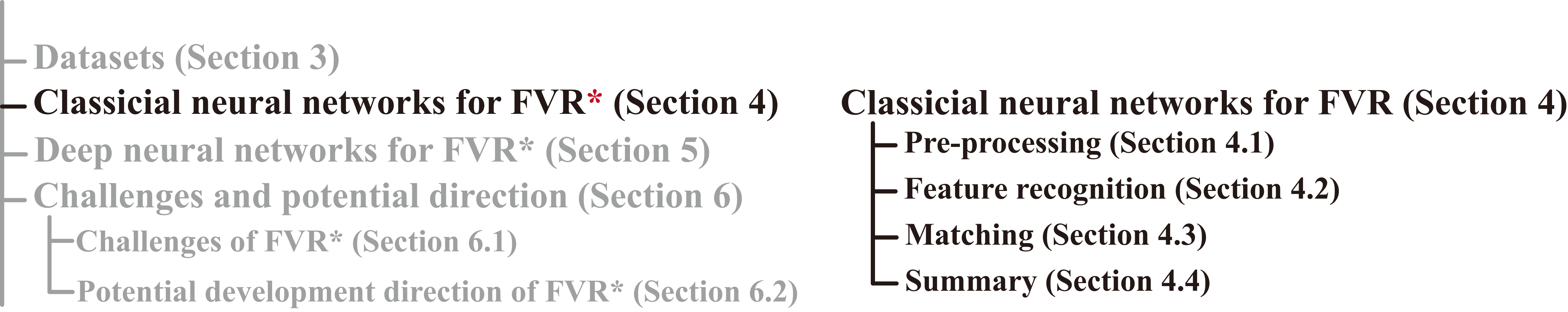}
    \caption{This section structure.}
    \label{S4}
\end{figure}

In this section, we follow the biometric recognition process to summarize the application of classical neural networks on FVR, including pre-processing, feature extraction, and matching. These ANNs transmit information through weighted connections between artificial neurons instead of convolutional layers. The specific structure of this section is shown in Fig.~\ref{S4}.

\subsection{Pre-processing}
The purpose of image pre-processing is to enhance the effectiveness of the features of the object and suppress the features of other factors. Typical pre-processing methods include image enhancement, filtering, segmentation, etc. In the FVR tasks using classical neural networks, it is vital to pre-process the finger vein images to make the finger vein features more clear. \cite{134} uses multi-scale self-adaptive enhancement transform based on wavelet to denoise and enhance the finger vein feature in the finger vein images. The wavelet transform is good at extracting point features. However, the finger vein images exhibit more significant linear features than point features. To improve the previous work, \cite{135} proposes the multi-scale self-adaptive enhancement transform based on curvelets. The method applies curvelets decomposition to finger vein images making the pre-processing strategy not only have the local time-frequency analysis capability of wavelets but also the differentiation capability of direction selection and the identification capability of linear features. The pre-processing steps in~\cite{136} are divided into four parts: vein-region division, Gamma enhancement, Gary-scale, and contrast enhancement. Firstly, the vein region is divided in the original image. After the feature of the vein region is Gamma enhanced to obtain better image tones, followed by the conversation of RGB image to grayscale using the Gary-scale method, and finally, the contract of the image is enhanced. The above pre-processing method is complex, but the pre-processing of some studies is straightforward, and it also helps improve the network's recognition performance. In~\cite{140}, the range of pixel values is normalized, followed by ROI extraction and image enhancement. \cite{141} uses image cropping to highlight the finger vein region.

\subsection{Feature extraction}
\begin{figure}[htbp]
    \centering
    \includegraphics[width = 0.75\textwidth]{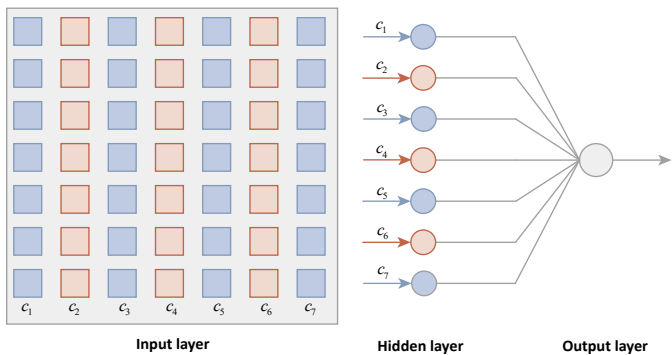}
    \caption{The structure of the local interconnection neural network~\cite{134, 135}.}
    \label{7NNN}
\end{figure}

Extracting venous features from finger vein images is a critical step in finger vein verification. Although pre-processing techniques can improve the quality of images, the processed images still contain some irrelevant information, which is inefficient to be used for model training directly. Feature extraction techniques focus on extracting the required information from a large amount of information and removing the irrelevant information to the maximum extent possible. \cite{134, 135} construct a local interconnection neural network to extract features from pre-processed finger vein images. As shown in Fig.~\ref{7NNN}, this network has seven columns with seven nodes in each column in the input layer, seven nodes in the hidden layer, and one node in the output layer. Nodes in one column of the input layer form a local interconnection structure with a node in the hidden layer. This network reduces the computational effort of fully connected networks, and both local and global information can be considered. 

\cite{136} uses Radon transform to extract the feature. Radon transform is a mathematical projection method that condenses the image's information in a few high-value coefficients on the transformed domain. In~\cite{140}, repeated line tracking, Gabor filter, and image segmentation are used to extract the feature. The repeated line tracking method clears the irrelevant information by removing irregular shadows generated by the thickness of the finger bones. The Gabor filter is mainly used to detect the finger vein images' length and width; subsequently, the finger vein contour is segmented from the entire image.

In addition to these feature extraction methods mentioned above, PCA and \emph{Linear Discriminant Analysis} (LDA) are also applied to FVR~\cite{137, 138, 139, 141, 142}. PCA is a common data dimensionality reduction method, which finds the most significant features from the high-dimensional features for retention, thus realizing the dimensionality reduction of features and simplifying the computation. \cite{139} introduces LDA to extracting features. Analogous to PCA, LDA is a data dimensionality reduction method that separates two or more classes by finding a linear combination of features.

\subsection{Matching}
In the matching phase, the feature extracted from the row data is compared with the finger vein information stored in the registration database for identity verification. Since the finger vein structure contains feature points formed by the intersection of vein lines, template matching based on pixel value is suitable for FVR. \cite{134, 135} address the lack of robustness of traditional template matching methods by identifying the blurred areas around the vein vessels and ignoring the slick misalignments between vein patterns. The method is implemented by relabeling the vein track space in the blurred region with pixel values between 46 and 170.

\begin{figure}[htbp]
    \centering
    \includegraphics[width = 1.0\textwidth]{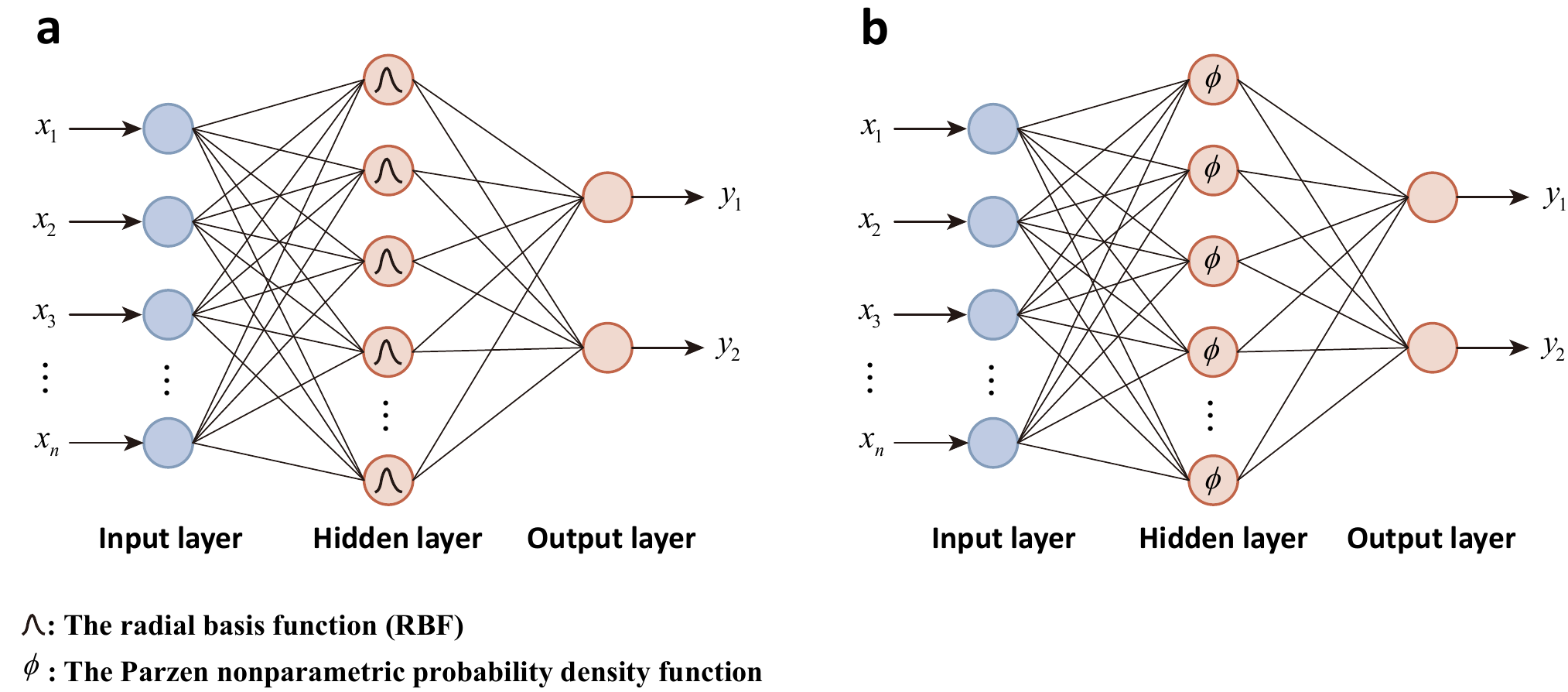}
    \caption{The structure of RBFNN and PNN~\cite{136}. \textbf{(a)} RBFNN. \textbf{(B)} PNN.}
    \label{M1}
\end{figure}

ANNs show a powerful performance in matching step~\cite{136, 137, 138, 139, 140, 141, 142}. \cite{137, 138} use \emph{Adaptive Neuro-Fuzzy Inference Systems} (ANFIS) to match the finger vein features. The structure of ANFIS is a merger of an adaptive network and a fuzzy inference system, which inherits the interpretability of the fuzzy inference system and the learning ability of the adaptive network. The ANFIS can change the system parameters based on prior knowledge to make the output closer to optimization~\cite{152}. These papers also conduct comparative experiments using MLP based on the BP algorithm, and the experimental results show that ANFIS significantly outperforms MLP. \cite{136} uses \emph{Radial Basis Function Neural Network} (RBFNN) and \emph{Probabilistic Neural Network} (PNN) to perform the classification of finger vein images. The structures of RBFNN and PNN are shown in Fig.~\ref{M1}. In RBFNN, the hidden layers provide the clustering ability in the classification process because its nodes consist of RBF that realize the nonlinear transformation from the input space to the hidden space. PNN is a supervised feed-forward neural network whose network structure is built directly through Parzen nonparametric probability density function. The experimental results show that both RBFNN and PNN have rewarding performance, achieving identification rates of 98.3\% and 99.2\%, respectively. Additionally, the training time of PNN is shorter.

\subsection{Summary}
In the papers we summarized, the classical neural networks for FVR are most used for classification, and only~\cite{134, 135} are used for feature extraction. This may be because that early ANNs had limited ability to extract venous features since the networks were too shallow. Although their ANN-based feature extraction methods achieve great experimental results, the samples used in their experiments are deficient. Experiments conducted on the small samples fail to demonstrate the practical application of the model. Unlike classical neural networks, the PCA-based feature extraction method requires few pre-processing steps and eventually achieves excellent performance, indicating that PCA has a solid ability to abstract complex features. In the classification process, the MLP based on the BP algorithm that is widely used in other tasks performs poorly on FVR tasks, with \emph{Accuracy} (ACC) rates lower than 50\% in~\cite{137, 138}. Conversely, ANFIS achieves satisfactory performance. We have provided a summary table for these papers in Tab.~\ref{sc}.

\begin{sidewaystable}[htbp]
    \centering
    \caption{Summary of the FVR based on classical neural networks.}
    \label{sc}
    \resizebox{\linewidth}{!}{
    \begin{tabular}{ccclllcc}
        \toprule
        \multirow{2}*{\textbf{Year}} & \multirow{2}*{\textbf{Reference}} & \multirow{2}*{\textbf{Task}} & \multicolumn{3}{c}{\textbf{Method}} & \multirow{2}*{\textbf{Dataset}} & \multirow{2}*{\textbf{Results}} \\
        \cline{4-6}
        ~ & ~ & ~ & \textbf{Per-processing} & \textbf{Feature extraction} & \textbf{Matching} & ~ & ~ \\
        \hline
        \specialrule{0em}{3pt}{3pt}
        2005 & \cite{134} & Verification & \makecell[l]{Multiscale self-adaptive enhancement transfornm \\ based on wavelet} & Local interconnection NN & Template & Private & EER = 0.130\% \\
        \specialrule{0em}{3pt}{3pt}
        2006 & \cite{135} & Verification & \makecell[l]{Multiscale self-adaptive enhancement transfornm \\ based on curvelets} & Local interconnection NN & Template & Private & EER = 0.128\% \\
        \specialrule{0em}{3pt}{3pt}
        2009 & \cite{136} & Verification & \makecell[l]{Vein-region Segmentation \\ Image enhancement \\ Gray-scale } & Random transform & \makecell[l]{RBFNN \\ PNN} & Private & \makecell[c]{ACC = 98.3\% \\ ACC = 99.2\%} \\
        \specialrule{0em}{3pt}{3pt}
        2010 & \cite{137} & Verification & $\backslash$ & PCA & \makecell[l]{MLP \\ ANFIS} & Private & \makecell[c]{ACC = 48\% \\ ACC = 99\%} \\
        \specialrule{0em}{3pt}{3pt}
        2011 & \cite{138} & Verification & $\backslash$ & PCA & \makecell[l]{MLP \\ ANFIS} & Private & \makecell[c]{ACC = 48\% \\ ACC = 99\%} \\
        \specialrule{0em}{3pt}{3pt}
        2011 & \cite{139} & Verification & $\backslash$ & \makecell[l]{PCA \\ LDA} & \makecell[l]{SVM \\ ANFIS} & Private & \makecell[c]{ACC = 98\% \\ ACC = 98\%} \\
        \specialrule{0em}{3pt}{3pt}
        2015 & \cite{140} & Verification & \makecell[l]{Image normalization \\ ROI extraction \\ Image enhancement} & \makecell[l]{Repeated line tracking \\ Gabor filter \\ Image segmentation} & ANN & $\backslash$ & $\backslash$ \\
        \specialrule{0em}{3pt}{3pt}
        2017 & \cite{141} & Verification & Image cropping & PCA & MLP & Private & ACC = 81\% \\
        \specialrule{0em}{3pt}{3pt}
        2017 & \cite{142} & Verification & $\backslash$ & PCA & ANN & Ptivate & \makecell[c]{FAR = 6.53\% \\ FRR = 0.71\% \\ ACC = 93.27\%} \\
        \specialrule{0em}{3pt}{3pt}
        \bottomrule
    \end{tabular}}
\end{sidewaystable}

\section{Deep neural network for finger vein recognition}\label{sec5}
\begin{figure}[htbp]
    \centering
    \includegraphics[width = 1.0\textwidth]{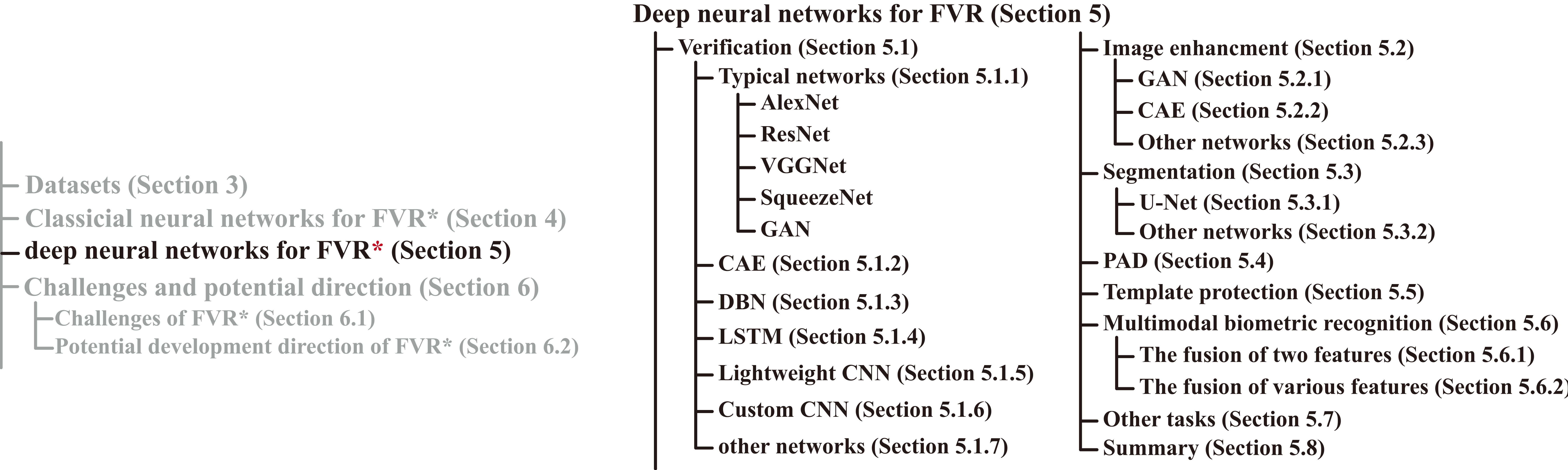}
    \caption{This section structure. \textbf{BI} represents biometric information.}
    \label{S5}
\end{figure}

This section introduces the FVR task based on deep neural networks. Unlike classical neural networks, deep neural networks have been used in a wide range of applications on FVR. We will discuss the literature based on the tasks of the paper, which are verification, image enhancement, segmentation, PAD, template protection, and other tasks. In addition, multi-biometric recognition, including finger vein, is also discussed. The specific structure of this section is shown in Fig.~\ref{S5}

\subsection{Verification}

\subsubsection{Verification based on typical CNN}

\paragraph{AlexNet}
Since the pre-processing process of the traditional ML-based FVR method is too complicated, \cite{99} proposes an AlexNet-based finger vein verification system to solve the problem. The AlexNet is used to extract feature vectors, and Euclidean distance is used to calculate the distance between two feature vectors for verification. The experimental results show that the method achieves an \emph{Equal Error Rate} (EER) of 0.21\% on the private dataset, indicating that the network can effectively discriminate the distance between intra-class and inter-class. In~\cite{191}, the finger vein texture features are first extracted by a local coding method constructed by a set of fixed sparse predefined binary convolution filters. The features extracted by the local encoding method are robust to rotation and illumination changes. These features are fed to pre-trained modified AlexNet for further learning. Finally, the SVM is used for classification. The experimental results show that this method obtains an average ACC of 98.78\% on the SDUMLA-HMT dataset.

\cite{101, 100} design the modified AlexNet network for end-to-end learning. In~\cite{101}, The ROI extracted images are directly fed into AlexNet for classification. The experiment achieves a correct rate of 99.53\% on the SDUMLA-HMT dataset. Compared with~\cite{101}, the network in~\cite{100} is more lightweight and can be deployed on Android platforms. Meanwhile, this system incorporates the ResNet module and SENet module to enhance the ability of feature extraction. This system reaches a recognition rate of 94.53\%.

\cite{104} proposes a network structure inspired by AleNxet. The research introduces the densely connected layers to the base structure of AlexNet. The study also compares the performance of two loss functions, Softmax and triplet-loss. Based on the network structure designed in this research, Softmax and triplet-loss achieve the ACCs of 97.00\% and 97.35\% on FV-USM, respectively. As AlexNet was an early successful CNN for image classification, there are related transfer learning methods on FVR using pre-trained AlexNet networks. \cite{102, 103} use a pre-trained and fine-tuned AlexNet network for classification. Using transfer learning instead of initial tuning of network weights makes network training faster. 

\begin{figure}
    \centering
    \includegraphics[width = 1.0\textwidth]{ 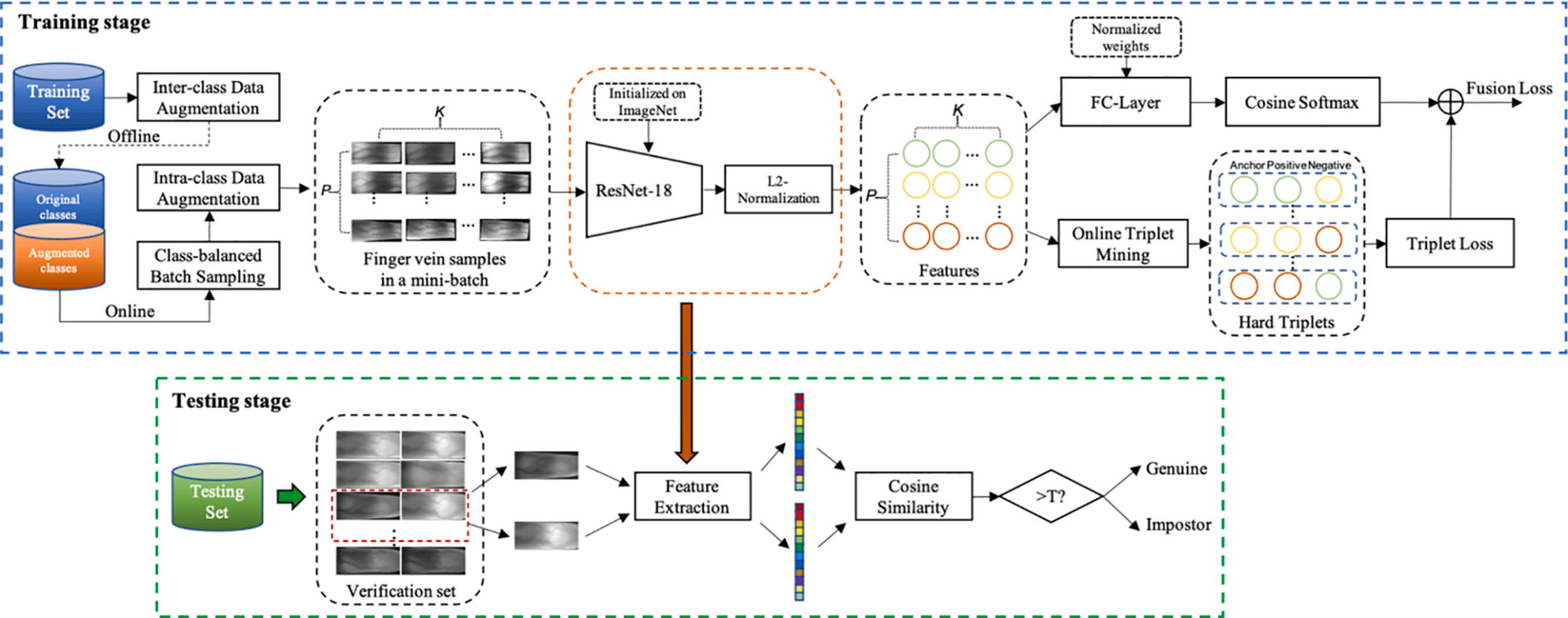}
    \caption{The workflow of the real-time FVR system~\cite{105}.}
    \label{res1}
\end{figure}

\begin{figure}[htbp]
    \centering
    \includegraphics[width = 0.7\textwidth]{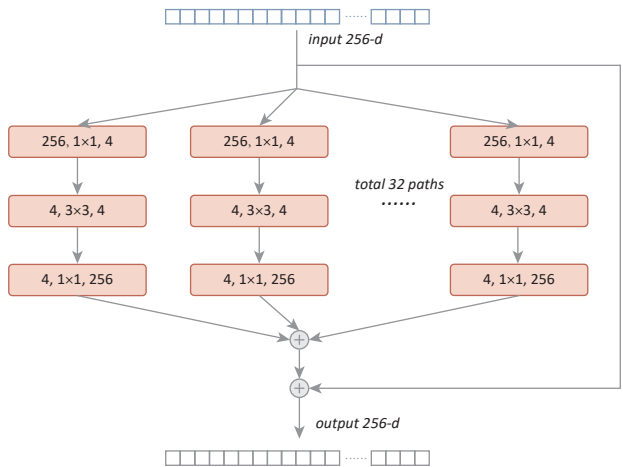}
    \caption{The split-transform-merge strategy of ResNext~\cite{112}.}
    \label{resnext}
\end{figure}

\paragraph{ResNet}
\cite{105} proposes a real-time FVR system that employs a fusion loss to learn more robust features by combining classification loss and metric learning loss, and an inter-class data augmentation technique is used to solve the lack problem of training data. In this system, ROI extraction and alignment are performed by the method of~\cite{106}, and the ResNet-18 based on the transfer learning technique is used for feature extraction. This network employs cosine similarity as the metric for matching. The specific flow of the whole system is shown in Fig.~\ref{res1}. The experimental results show that the method achieves an EER of 0.48\% on FV-USM, which is significantly lower than other DL methods. \cite{107} proposes a FVR method based on ResNet and U-Net. Both network models are trained by the end-to-end approach. The proposed method introduces bias field correction and spatial attention mechanisms. In this system, the contrast of the original images is first adjusted by the bias field correction model, followed by the inversion of the pixel values. The processed images are fed into the U-Net-based spatial attention model for enhancing the selective region information. Finally, ResNet-50 classifies the enhanced images. The method reduces the impact of low-quality images and classifies finger veins using global features, and makes the network can extract more significant features. The experimental results show that the method has a rank-one verification rate of 99.53\% on SDUMLA-HMT and 98.20\% on THU-FVFDT2. \cite{108, 109, 110} propose improving CNN structures with the help of the residual idea of ResNet rather than just using the original ResNet for image analysis tasks. \cite{108} proposes \emph{Efficient Channel Attention Residual Network} (ECA-ResNet) to enhance the practical application ability on FVR tasks. The ECA~\cite{111} can bring significant benefits to the model with a small number of parameters, breaking the paradox between performance and complexity. \cite{109} proposes a new ResNet-based architecture, called \emph{FV2021}. The network is compact and suitable for installation on mobile devices. FV2021 uses separable convolutional layers instead of normal convolution layers to reduce the model complexity. In~\cite{110}, a novel ResNet-based network architecture, ResNext, is proposed. This network is employed for classification, it has a homogeneous multi-branch structure with only a few hyperparameters to tune, and this structure uses the split-transform-merge strategy for scaling any large number of transformations without the need for specialized design. Meanwhile, this model uses \emph{cutout}~\cite{112} as the data augmentation strategy. The specific structure of ResNext is shown in Fig.~\ref{resnext}. The model still outperforms the original ResNet even with the same model complexity. As the extension of this method, \cite{195} uses a neural architecture search network for FVR. This network uses a controller neural network to sample subnetworks with different structures. This approach is used to update the parameters of the controller network to generate a better architecture. The performance of this network is better than the previous ResNext.

\paragraph{VGGNet}
\cite{113, 114} use VGG-16-based CNN for finger vein verification. \cite{113} resizes the detected finger vein ROI to a $224\times224$ pixel image and then obtains the difference image between the input finger vein ROI image and the registered finger vein ROI image. The difference image is fed to the VGG-16-based CNN model to directly obtain the matching results without the redundant image pre-processing step. The experimental results show that the adjusted VGG-16 network achieves EERs of 0.396\%, 1.275\%, and 3.906\% on three datasets with different image quality (from high to low), respectively. VGG-16 also inspires the network of~\cite{114}, and this study uses a wide line detectortor~\cite{106} to extract finger vein features from the normalized images in addition to simple ROI extraction. The method achieves an EER of 0.42\% on the private dataset.

\paragraph{SqNet}
SqNet has fewer parameters without losing accuracy than other deep neural networks and is also used in some FVR tasks. \cite{116} proposes a lightweight SqNet that can be deployed on hardware platforms with limited computational power and memory. The study uses \emph{3 Channel} (3C) images as the input of the SqNet. The 3C images are obtained by the different operation between the input images and registered images. The SqNet can be fine-tuned to achieve the most accurate performance by feeding 3C images into a network that has been pre-trained on ImageNet. The experimental results fully demonstrate that the method achieves high recognition rates while simplifying the network structure, whose EERs of 1.889\% and 4.906\% on MMCBNU-6000 and SDUMLA-HMT, respectively. \cite{117} uses pre-trained SqNet to extract the features of the left finger vein and right finger vein and then uses SVM for classification. This method significantly decreases the feature dimension. The experimental results show that this feature fusion method achieves an ACC of 99.81\% on SDUMLA-HMT and an ACC of 99.36\% in one section of FV-USM. The longitudinal rotation of fingers in the acquisition process can affect recognition performance. To solve this problem, \cite{118} proposes a neural network with robustness to longitudinal finger rotation by training the neural network using finger vein images with different angles. These images from different angles are from two places. One is captured from different angles, and another uses a data augmentation strategy to simulate the longitudinal rotation of the fingers. The study uses two networks, Triplet-SqNet and DenseNet-161, to train these images separately. Experimental results on the PLUSVein dataset show that training with images acquired from different angles leads to a more robust model, especially Triplet-SqNet.

\begin{figure}[htbp]
    \centering
    \includegraphics[width = 0.8\textwidth]{ 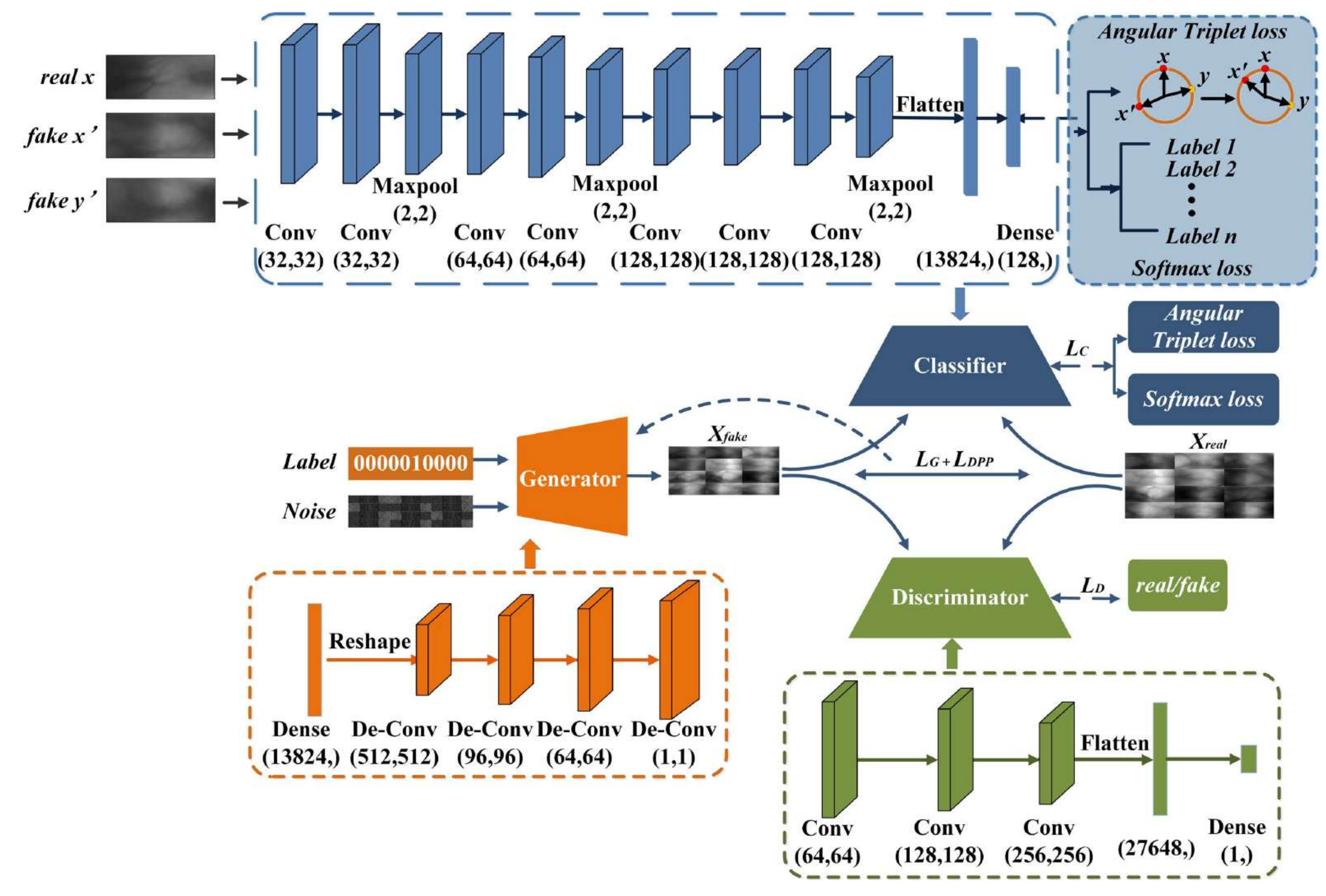}
    \caption{Architecture of the triplet-classifier GAN~\cite{121}.}
    \label{gan1}
\end{figure}

\paragraph{GAN}
\cite{120} proposes a novel structure termed FV-GAN based on \emph{Cycle-consistent Adversarial Network} (CycleGAN) to extract features and perform the verification of finger veins. The generator in the FV-GAN consists of an image generator based on U-Net and a pattern generator based on an encoder-decoder network. The pattern generator extracts the finger vein patterns from the image and outputs the probability of each pixel belonging to the vein pattern, and a binary discriminator is used for verification. In addition, FV-GAN uses a fully convolutional structure to reduce the cost of computation. The experimental results show that the EER of FV-GAN is 0.94\% on  SDUMLA-HMT and 1.12\% on THU-FVFDT2. \cite{121} proposes a GAN-based structure termed triplet-classifier GAN, which combines a conditional generator and an angular triple loss-based classifier. As shown in Fig. \ref{gan1}, the triplet-classifier GAN is used for data augmentation and classification, and the data augmentation strategy can enhance the training effect of the network. In addition, the cosine similarity is used to replace the Euclidean distance in the designed angular triple loss to improve the feature extraction ability. The experimental results show that this model achieves the EERs of 0.05\%, 0.14\%, and 0.15\% on SDUMLA-HMT, FV-USM, and HKPU, respectively. \cite{122} employs a \emph{Deep Convolutional Generative Adversarial Network} (DCGAN) to identify genuine and fake finger vein images. The generator generates fake images from genuine images, and the discriminator is used to distinguish between genuine and fake images. The experimental results show that DCGAN significantly improves the performance of finger vein verification. The method can be used for forensic identification and biometric verification of the finger vein. Traditional FVR systems are trained using one type of data and have serious performance degradation problems when the trained model is applied to different types of data. To improve the recognition performance of the network on heterogeneous datasets, \cite{158} proposes a FVR system incorporating the domain adaptation technique based on CycleGAN. All input samples are pre-processed and fed into CycleGAN to generate composite images so that data from different domains have some similarity. Due to the advantage of this method, the model generality is enhanced. The composite image is fed to DenseNet-161 for classification. This experiment is trained on SDUMLA-HMT and tested on HKPU, and an EER of 0.85\% is obtained.

\subsubsection{Verification based on CAE}
\cite{123} proposes a finger vein verification system that integrates CAE-based feature learning methods with CNN. This system first extracts features from the finger vein images using CAE. Then, these features are fed into the CNN to further depth feature extraction and classification. Using this method, the recognition rate for the FV-USM is 99.16\%, and the EER is 0.16\%. The workflow of~\cite{124} is similar to~\cite{123}. CAE is used to learn features of a particular distribution from original finger vein images, and SVM is used for classification. The experimental results show that the method achieves an EER of 0.12\% on FV-USM and 0.21\% on SDUMLA-HMT. \cite{125} uses CAE to extract preliminary features from the image, and then the parameters of the CAE are used to initialize the parameters of the convolutional layers of a deep CNN. This deep CNN is used to classify the finger vein images. Meanwhile, the \emph{Extreme Learning Machine} (ELM) layers are used to replace the fully connected layers after the training is completed. Because the ELM has faster convergence and better generalization ability than the traditional BP algorithm. The experimental results show that this method achieves EERs of 98.88\% and 98.58\% on FV-USM and SDUMLA-HMT.

\subsubsection{Verification based on DBN}
\cite{126, 127} proposes a feature fusion FVR algorithm based on DBN and CNN. It uses the features extracted from DBN and CNN and then uses the similarity measure for matching. To reduce the learning and detection time of the model, feature points based on endpoint and intersection are extracted using the feature extraction method based on eight neighborhoods. These feature point sets are then used as inputs to the network, simplifying the computation of the network and making it adaptable to mobile terminals. The ACC of the method on the private dataset reaches 99.6\%. \cite{128} proposes a FVR algorithm based on DBN and uniform LBP operator. The texture features are extracted from the sub-blocks of finger vein curvature gray images using the uniform LBP operator. This makes the learned feature contain more vein information. Then, the histogram of the sub-block features is computed and integrated into an overall histogram for training the DBN. The experimental results show that the recognition rate of this method is 97.4\% on FV-USM.

\subsubsection{Verification based on LSTM}
\cite{143} proposes a network model for feature extraction by combining CNN and LSTM. CNN represents the vein texture features in the local region, and LSTM is used to capture the spatial location relationship within the region. This way of extracting features considers the spatial location relationship between features and makes the model more robust. Finally, the Hamming distance is used for matching. The proposed LSTM-based network achieves an EER of 0.95\% on HKPU. \cite{144} designs a bidirectional LSTM-based verification system. The system uses ROI extraction and Gaussian filtering for pre-processing, followed by feature extraction using bidirectional LSTM and shark smell optimization algorithm to optimize the hyperparameters. Finally, Euclidean distance is used for matching. The method surpassed the earlier methods, and the maximum ACC is 99.93\%. The traditional FVR system requires the users to hold their finger for a few seconds to complete the verification process. \cite{145} conducts a real-time verification system of finger veins. The system can obtain the user's finger vein feature dynamically. The system extracts finger vein image sequences from recorded videos and uses CNN to extract features from the pre-processed sequences. Then, LSTM is used to find and track the temporal dependencies within the input feature sequences for verification. The system achieves an ACC of 99.13\% on the collected dataset. The images in this dataset were taken from different exposure times.

\subsubsection{Verification based on Lightweight CNN}
\begin{figure}[htbp]
    \centering
    \includegraphics[width = 1.0\textwidth]{ 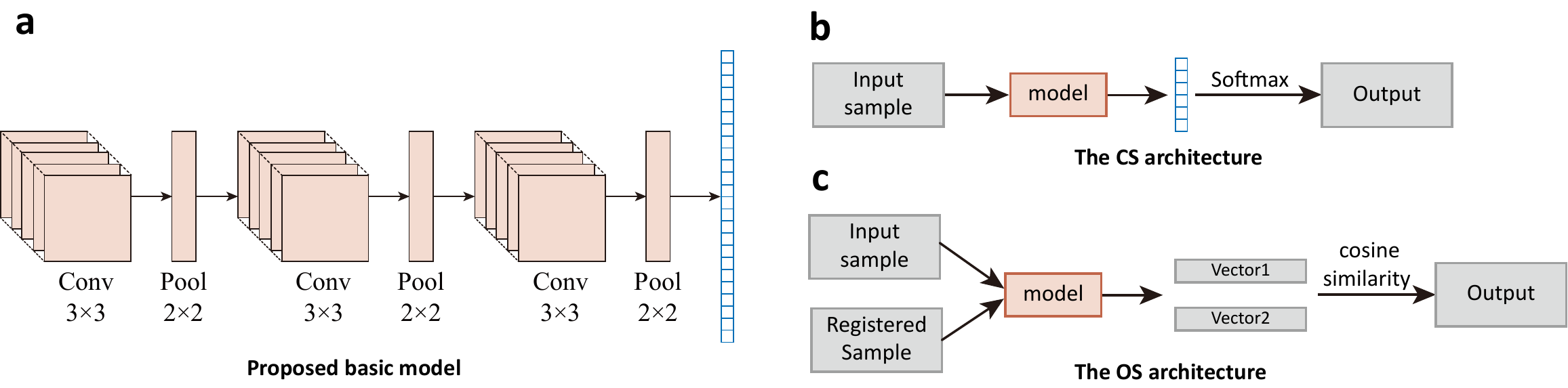}
    \caption{The details of the framework~\cite{155}. \textbf{(a)} The Proposed lightweight CNN. \textbf{(b)} CS architecture. \textbf{(c)} OS architecture.}
    \label{L1}
\end{figure}

Some deep neural networks achieve excellent performance by stacking layers but lose the ability to be applied well because the models are too complex. To let the networks perform verification tasks on mobile terminals, it is necessary to explore the development of lightweight CNN on FVR tasks. \cite{153, 154} use lightweight CNN for FVR, and these network structures have a few convolutional and fully connected layers. The network of~\cite{153} is end-to-end, and it is trained by the joint supervise based on the center loss function and the Softmax loss function, which can obtain highly discriminative features for FVR. \cite{154} proposes a lightweight CNN to perform the classification task along with feature extraction and optimization method of maximum curvature finger vein features based on Gaussian filter, which reduces the influence of image noise on recognition. The lightweight CNN proposed in~\cite{155} has two structures by using different loss functions, and these two architectures are \emph{Closed-set} (CS) architecture and \emph{Open-set} (OS) architecture. The specific process is shown in Fig.~\ref{L1}. The CS architecture uses Softmax to predict the class of the input samples. The OS architecture in this study outputs the feature vectors of the input samples and registered samples. The experimental results show that the network achieves EERs of 2.29\% and 0.47\% on SDUMLA-HMT and MMCBNU-6000. \cite{156} designs a lightweight CNN model using a partially pre-trained MobileNetV2~\cite{157} as a backbone. The model achieves high performance on FVR tasks by using pre-trained auxiliary blocks and customized auxiliary blocks while simplifying the training process. This network achieves the CIR of 96.98\% on public datasets.

\subsubsection{Verification based on custom CNN}
To perform FVR tasks in a targeted manner, many studies use customized deep network structures to analyze finger vein images. The researchers design the number of layers and parameters of these networks according to the FVR task without the help of representative deep neural networks such as AlexNet, VGGNet, and ResNet.

\cite{178, 179} use \emph{Curvature Gray Images} (CGIs) instead of original finger vein images as the input of the network. Their CGIs are obtained from 2D Gaussian templates, and CGIs are directly fed into customized CNN for classification. These studies employ the improved activation function instead of ReLU. The improved activation function has both the ReLU function's sparsity and the Softplus function's smoothness. The ReLU and Softplus functions are shown in~\myref{ReLU} and~\myref{SP}. \myref{IAC} illustrate this improved activation function. This method is effective to resist noise interference and improve the recognition rate. \cite{190} uses two relatively independent sub-convolutional networks with different granularity for FVR and LeakyReLU as the activation function. The experimental results show that the two sub-convolutional networks can extract feature more effectively and achieves an ACC of 95.1\% on the small dataset.

\begin{equation}\label{SP}
    S(t)=ln(1+e^t)
\end{equation}

\begin{equation}\label{IAC}
S(t) = \left\{ {\begin{array}{*{20}{c}}
0&{t < 0}\\
{\ln (1 + {e^t}) - \ln 2}&{t \ge 0}
\end{array}} \right.
\end{equation}

Researchers have designed some neural networks to perform only one part of feature extraction or classification. \cite{180} removes the bottom structure from a pre-trained CNN model and retains the front structure to extract features from finger vein images. These images are matched using a template matching strategy. \cite{181} uses PACNet~\cite{183} instead of CNN for feature extraction and ridge regression classifier~\cite{182} for classification. The PCANet filter is generated based on the correlation between the original image and the grayscale image. The best performance of these two methods on the public dataset reaches an EER of 0.04\% and an ACC of 100\%, respectively. \cite{184} uses PCA for feature extraction and then uses a custom CNN model for classification, achieving a recognition rate of 98.53\% on FV-USM.

\begin{figure}[htbp]
    \centering
    \includegraphics[width = 1.0\textwidth]{ 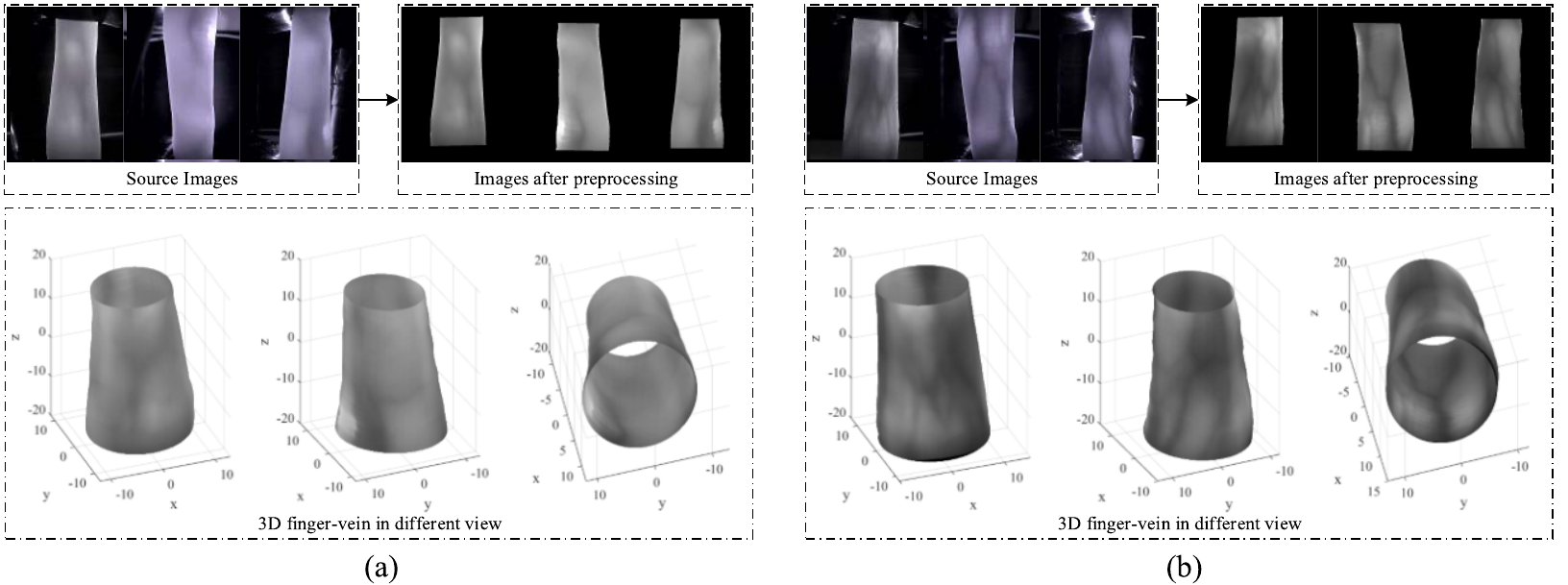}
    \caption{The 3D finger vein image~\cite{188}.}
    \label{CC1}
\end{figure}

\begin{figure}[htbp]
    \centering
    \includegraphics[width = 1.0\textwidth]{ 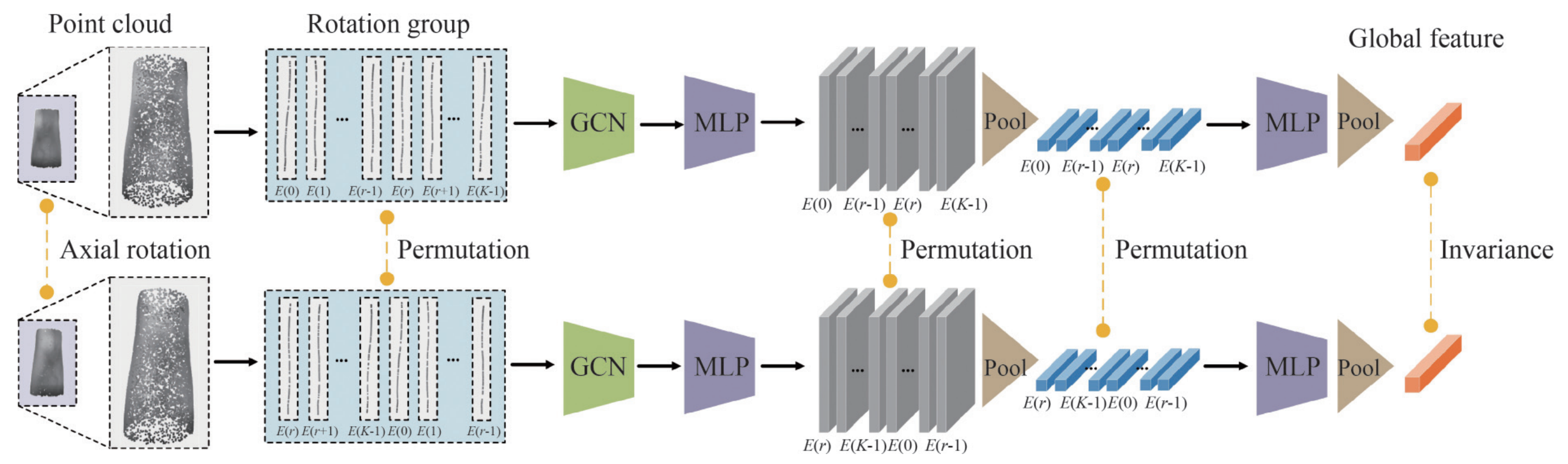}
    \caption{The structure of 3DFVSNet~\cite{132}. \textbf{GCN} represent graph convolutional neural network.}
    \label{CC2}
\end{figure}

To solve the problem that traditional 2D finger vein images are easily affected by finger position and posture changes during acquisition, \cite{188} constructs 3D finger vein images and uses these images for detection. This study uses three cameras to perform the imaging process. The 3D images usually include more sufficient vein information than 2D. This research uses a 3D reconstruction algorithm and a corresponding texture mapping algorithm to create a 3D image based on three 2D finger vein images, and 3D finger vein images are shown in Fig.~\ref{CC1}. Finally, the lightweight CNN with depth-wise separable convolution is used for feature extraction and matching. The experimental results are shown that this method achieves EERs of 0.94\%, 1.69\%, and 2.40\% on FV-USM, SDUMLA-HMT, and HKPU. As an extension of previous work, \cite{132} uses a contour-based optimization model for 3D FVR and a corresponding acceleration strategy to obtain 3D point clouds of finger vein structures. A custom CNN structure, 3DFVSNet, is used to extract rotationally invariant features, and the specific network structure is shown in Fig.~\ref{CC2}. Cosine similarity distance is used for verification. The Experimental results show that 3DFVSNet has powerful robustness to axial rotation.

Some structures use Gabor filters~\cite{187} instead of convolutional layers. Gabor filter is a wavelet with good transform properties of time and frequency domains. \cite{185, 186} use Gabor convolutional layers instead of the normal convolutional layers in the CNN structure, and the whole model is end-to-end. The method maintains the advantages of the Gabor filter in FVR and has the excellent feature extraction ability of CNNs. They all achieve favorable recognition results on public datasets. \cite{169} proposes a novel \emph{Trilateral Filterative Hermitian Feature Transformation based Deep Perceptive Fuzzy Neural Network} (TFHFT-DPFNN) model for improving the verification accuracy of FVR. In this system, the Hermitian Hat wavelet is first used to decompose and extract features from the noise-reduced finger image, and then the features are transferred to the next hidden layer. These features are matched by the Jaccard similarity index. Finally, the fuzzy membership function of the output layer is applied to output the validation results. The TFHFT-DPFNN achieves an ACC of 98\% on SDUMLA-HMT.

The attention mechanism has been introduced into some custom CNN structures. \cite{193} designs a network structure consisting of a convolutional layer and three \emph{Joint Attention} (JA) models. The JA model can improve the model's discrimination of low-contrast finger vein images by exploiting the interdependence between the spatial locations of channels and feature maps. In addition, this network introduces GeM pooling layers to enhance the feature. Numerous experiments demonstrate the validity of each network component, and the network has an EER of only 0.34\% on FV-USM. \cite{196} proposes an improved FVR model based on the residual attention mechanism. The main part of this network is the residual attention blocks, and this network is divided into the trunk branch and the soft mask branch. The trunk branch extracts finger vein features from feature maps that are generated by the previous layers, and the soft mask branch learns global vein information using the hourglass network~\cite{197}. Benefiting the advantages of this structure, this network can learn more abundant finger vein information than other simple networks without branching structures. This method achieves the ACCs of 98.58\% and 97.54\% on FV-USM and MMCBNU-6000. \cite{194} designs a network based on bilinear pooling to extract the second-order features of finger veins, and the complexity of this network is reduced by replacing conventional convolution with DSC. Meanwhile, this method designs a dimensional interactive attention mechanism to enhance the correlation between channels and space, further increasing the model's recognition accuracy. This method reaches an ACC of 100\% on FV-USM.

In~\cite{214}, The vein areas are extracted from original finger vein images using Sobel operator and polygon ROI, and then the dark vein lines are generated using double \emph{Contrast Limited Adaptive Histogram Equalization} (CLAHE). Finally, these processed images are fed into a 20-layer CNN for feature extraction and verification. The application of CLAHE enhances the finger vein image quality. The experimental results show that this method achieves an ACC of 94.88\% on SDUMLA-HMT.

\cite{171} proposes a new FVR method that uses two different fingers instead of a single finger to perform verification, and the method is called \emph{Re-enforced DL} (RDL). The CNN model of RDL stores the weights of both the index and middle finger veins. In the RDL, the index finger is used for the first verification, and the middle finger for enhanced verification. The RDL method effectively utilizes the venous features of both fingers and achieves an ACC of 91.19\% on HKPU. \cite{198} performs finger vein verification with the ensemble DL system. In this system, several CNNs are first trained as weak classifiers, and then these classifiers are ensembled to obtain an integrated classifier for recognition. The input of the ensemble DL system is the feature maps from other CNNs, Gabor filters, and LBP. The ensemble DL system obtains ACCs of 92.11\% and 94.17\% on HKPU and FV-USM. 

\subsubsection{Verification based on other networks}

LeNet-5 is one of the most representative structures of early CNNs. It defined the main structure of the CNN, convolutional layer, pooling layer, and fully connected layer. \cite{130} designs a MATLAB-based FVR system. This system uses a modified LeNet-5 as the backbone. In this system, the image pre-processing and other steps are performed on different platforms, and these two parts are connected using MATLAB's MEX files. The experimental results show that this FVR system recognizes ten subjects within less than ten seconds and achieves an ACC of 96\%. \cite{159, 173} use the network proposed by~\cite{160} for FVR. This network is more lightweight than LeNet-5, with only two convolutional layers and two fully connected layers. To reduce the number of parameters, \cite{159} cuts off the connections between some neurons in the fully connected layers. In addition, this network uses the stochastic diagonal Levenburg-Marquardt algorithm to ensure fast convergence. The network achieves recognition rates of 100\% and 99.38\% on 50 and 81 subjects, respectively. ZFNet is also one of the classical CNNs. In~\cite{149}, the pre-processed images were fed to ZFNet for classification. The first convolutional layer of ZFNet uses small convolutional kernels and strides, which can retain more features of the original information. This network achieves an ACC of 86\% on one section of SDUMLA-HMT.

\cite{115} compares the performance of three well-known deep CNN structures, AlexNet, SqNet, and GoogLeNet, on FVR tasks. These networks have been pre-trained on ImageNet. The image noises are removed using a wiener filter, and these networks are trained with these enhanced images. The experiment results show that GoogLeNet outperforms other networks in terms of classification. \cite{129} introduces the DSC into a pre-trained Xecption model, replaying the traditional convolutional layers with DSC. The DSC enables Xception to learn more robust features from images and achieves superior classification performance than normal convolutional layers. This method achieves an ACC of 99\% and an F1-score of 98\% on the SDUMLA-HMT dataset, and an ACC of 90\% and an F1-score of 88\% on the THU-FVFDT2. \cite{147} uses discrete supervised hashing sequences and triplet loss function to train the network. Discrete supervised hashing can reduce the size of stored feature templates, and this method can improve the matching speed. This paper compares a lightweight CNN with the improved VGG-16, and the experimental results show that lightweight CNN has a superior performance.

DenseNet~\cite{163} has a more dense connection mechanism than ResNet. Each layer of DenseNet accepts the output of all its preceding layers as its additional input. This design greatly suppresses the gradient vanishing. \cite{161, 162} use DenseNet-based network architecture to perform FVR tasks. In~\cite{161}, three images are simultaneously fed into a pre-trained DenseNet-161. The input image, the registered image, and their composite image. This method solves the problem that the normal different image is susceptible to noise, and the composite input of three images makes the recognition performance more accurate. The method achieves EERs of 0.33\% and 2.35\% on HKPU and SDUMLA-HMT, respectively. \cite{162} is the first study to use both venous texture and finger shape features for FVR. The texture image and the segmented shape image are fed into the DenseNet-161 and then output their respective matching scores. Finally, the verification result is presented by score-level fusion. The method effectively utilizes multiple features of finger vein images and enhances the noise resistance of the model.

ShuffleNet~\cite{164, 166} maintains a balance between speed and accuracy by introducing pointwise group convolution and channel shuffle. \cite{165} uses ShuffleNet V2 as the backbone and removes the first pooling layer to generate larger feature maps, which helps to retain fine-grained features. In addition, this model uses Triplet and Softmax-based fusion loss function instead of the original Softmax loss function. The network achieves an EER of 0.05\% on the public dataset. 

\cite{131} transfers the knowledge of a pre-trained CNN model to a more lightweight Siamese CNN by knowledge distillation. This Siamese CNN uses a new modified contrastive loss function to improve the discriminative ability for features. The experimental results show that the EERs of this Siamese CNN on MMCBNU-6000, FV-USM, and SDUMLA-HMT are 0.08\%, 0.11\%, and 0.75\%. \cite{168} proposes a FVR system based on a two-branch network incorporating joint Bayesian loss. In this system, finger shape images and ROI vein images are fed into the two-branch network to extract more effective features from finger vein images. Joint Bayesian loss is used to train the network. The experimental results show that the system obtains the EERs of 0.17\% and 0.94\% on MMCBNU-6000 and SDUMLA-HMT. In~\cite{192}, two finger vein images from the same finger are fed into two CNNs that have parallel structures, and the two outputs are combined using the CLAHE method and the Gabor filtering method. This method utilizes the vein information from both images instead of a single image, and achieves an ACC of 99.56\% on THU-FVFDT2.

Different from the traditional CNNs, some novel network structures are used in FVR. \cite{170} uses a capsule network for FVR. Capsule network extract features of finger vein in a more reasonable way than CNNs by their translational and rotational invariance. In addition, the special structure of GNN determines that it can effectively learn the graph structure features of finger veins. \cite{172} applies GNN to perform FVR task for the first time. The method without pre-processing steps and data augmentation strategy. In this study, the images are first fed into a small CNN to extract feature vectors, and then an edge features learning network is used to model the relationships between all node pairs. Finally, the results are fed into a GNN to perform the classification. The experimental results show that GNN achieves high performance on FVR tasks with fewer parameters and faster model convergence.

\subsection{Image enhancement}

\subsubsection{Image enhancement based on GAN}
Finger vein images may be blurred due to low temperature, vein stretching, and illumination change. These factors seriously affect the quality of the images. Some GAN-based methods can repair these defects of impaired images to make these images greatly recognized by ANN. \cite{199} proposes a GAN based on \emph{Neighbors-based Binary Patterns} (NBP) to recover the finger vein images. This model uses NBP texture loss between the input image and the generated image to train the generator network, and the loss function is able to enhance the deblurring ability of the network. Meanwhile, residual connections are added to the generator network to prevent overfitting. The \emph{Peak Signal-to-Noise Ratio} (PSNR) of this GAN based on NBP reaches 30.42. dB. \cite{200} proposes a novel finger vein image restoration method, and this method also uses the NBP texture loss function. Meanwhile, the method uses Possion fusion in the input process to reconstruct the finger vein images to make the image boundary connection of the image more natural. The discriminator network consists of two Wasserstein GAN with Gradient Penalty modules to ensure consistency between the global and local information of the restored image. The experimental results show that adding texture loss can better recover the veins. \cite{201} and~\cite{202} are focus on the optical blurring and the motion blurring of finger vein images, respectively. \cite{201} uses a modified conditional GAN to restore optically blur red finger vein images. Previous conditional GANs use random noise in the form of dropout to ensure the randomness of noise. However, this method may change the information of the restored finger vein image. This study proposes a modified conditional GAN without dropout because of the need for deterministic output. A comparison with the original conditional GAN shows that the approach performs better image restoration. As an extension of the previous work, \cite{202} reduces the number of residual modules in the generator network to speed up the model inference and reduce the network parameters.

To overcome the obstacle of illumination, \cite{204} designs a \emph{GAN for The Illumination Normalization of Finger Vein Images} (INF-GAN). The structure of INF-GAN is designed based on Pix2Pix-HD~\cite{205}. The residual model is used as the generator and PatchGAN~\cite{206} as the discriminator. The residual image generation block can highlight the vein textures distorted by serving uneven illumination. \cite{207} proposes a finger vein images denoise method based on GAN, which called \emph{Custom Sample Texture Conditional GAN} (CS-TCGAN). This approach designs a joint loss function that combines adversarial loss, content loss, and texture loss to obtain more abundant vein information from the image than Softmax loss. Meanwhile, the CS-TCGAN describes the rough finger vein structures by the de-convolutional layers operation and then fills the details to generate the image. To make the training samples simulate the real noise distribution, this study also designs a dataset with a mixture of Gaussian, Poisson, and speckle noise. The CS-TCGAN shows a great denoising performance on a private dataset.

\subsubsection{Image enhancement based on CAE}
\cite{208, 209} use the CAE network for finger vein image enhancement. \cite{208} proposes a finger vein image with spots and stains repair solution. The scheme first removes the effect of illumination changes on the image by the Gabor filter and Weber’s low descriptor. Then, the images are fed into an encoder network structured as AlexNet to mark the smeared pixels and learn shallow features. Finally, the marked images are fed into the decoder network to recover the image. The workflow of~\cite{209} is similar like~\cite{208}, which uses the adaptive thresholding approach to detect contaminated areas in finger vein images before feeding them into CAE for image recovery. These two CAE-based image enhancement methods have great performance on private datasets.

\subsubsection{Image enhancement based on other networks}
\begin{figure}[htbp]
    \centering
    \includegraphics[width = 0.85\textwidth]{ 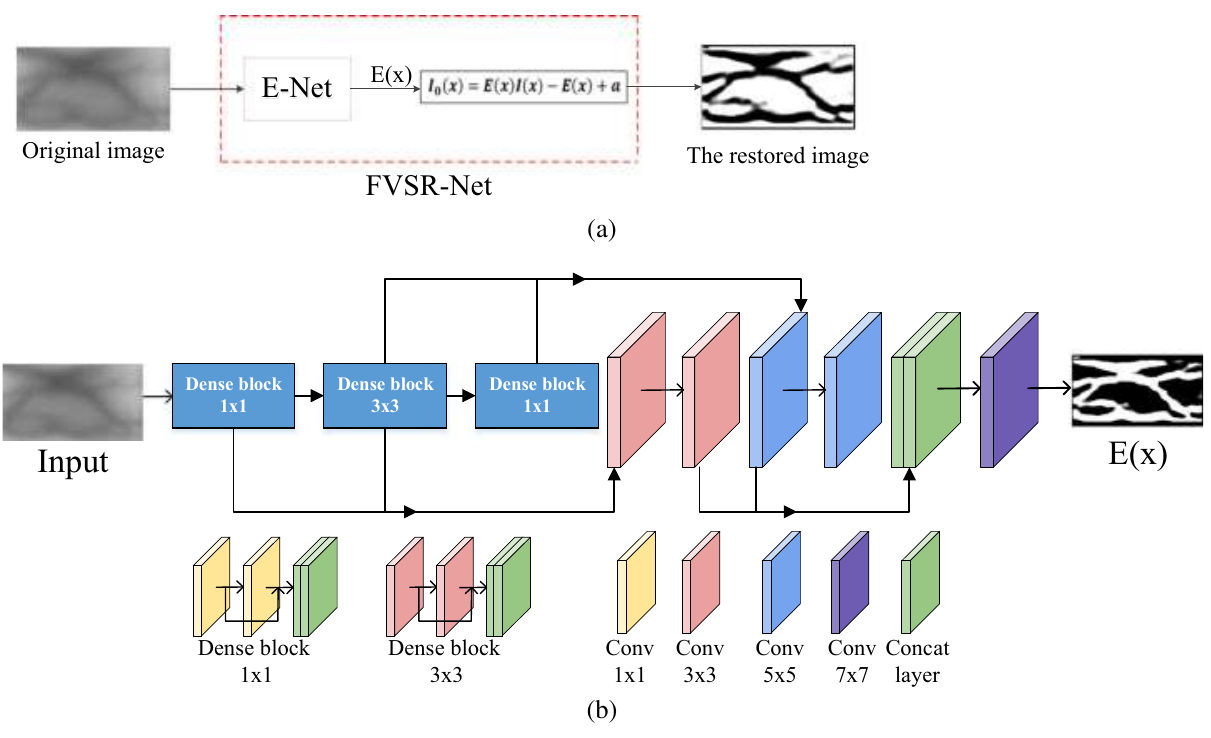}
    \caption{The framework of the proposed scattering removal model~\cite{212}. \textbf{(a)} The structure of the FVSR-Net. \textbf{(b)} The structure of the E-Net.}
    \label{EN1}
\end{figure}

\cite{210} pre-trains a lightweight VGG-16, and the pre-processed images are fed into this network for image enhancement. This modified VGG-16 structure includes 13 convolutional layers and five max-pooling layers. Meanwhile, The VGG-16 removes the fully connected layer. The ACC of more than 99\% is achieved using enhanced images by this network for recognition. \cite{211, 212} uses custom CNN structures for finger vein image enhancement. In~\cite{211}, the captured images are directly fed into the CNN for image enhancement. This network contains three convolutional layers and two deconvolutional layers. The convolutional layers are used to learn the distribution of noise features and generate feature maps, and the deconvolutional layers use the feature maps to reconstruct noise-free finger vein images. The model achieves a PSNR of 29.638 dB on HKPU. \cite{212} builds an end-to-end \emph{Finger Vein Image Scattering Removal Network} (FVSR-Net) by combining an optical scattering model and a multi-scale CNN named E-Net. The specific wolkflow of this method is shown in Fig.~\ref{EN1}, and the theory of the optical scattering model is shown in~\myref{OSRM}. $I$ and $I_0$ represent the original and restored images. The $E(x)$ and $a$ represent the output of E-Net and bias. The FVSR-Net achieves the PSNR of 13.5929 dB on SDUMLA-HMT.

\begin{equation}\label{OSRM}
    I_0(x)=E(x)I(x)-E(x)+a
\end{equation}

In addition to these CNN-based structures mentioned above, a novel neural network is used to enhance the finger vein images. \cite{213} uses a modified \emph{Pulse Coupled Neural Network} (PCNN) for finger vein image enhancement. PCNN has the characteristics of synchronous pulse release and global coupling to extract sufficient information from complex backgrounds. This study proposes a parameters tuning scheme based on the original PCNN to adjust the parameters without any empirical correlation automatically. The experimental results show that the PCNN-enhanced images can produce a rewarding recognition performance. 

\subsection{Segmentation}

\subsubsection{Segmentation based on U-Net}
\cite{215, 216, 217} compare the segmentation performance of three fully CNNs: U-Net, RefineNet, and SegNet. These studies use the manual labeling method and automatic labeling method to generate the ground truth images from the original datasets, and these ground truth images are used to train these networks. Finally, the segmentation performance is evaluated by calculating the correlation between the input images and the ground truth images. Numerous experiments illustrate that labeled images generated by automatic labeling methods can improve the segmentation accuracy of the network, and U-Net is more sensitive to the input image quality, while RefineNet and SegNet are more stable. 

The existing large finger vein segmentation networks are not suitable for implementation in mobile terminals since they are too deep. To solve this problem, \cite{223} proposes a lightweight network for finger vein segmentation. First, the DSC is introduced to the original U-Net to reduce the model's parameters, and the Ghost model~\cite{224} is introduced to the network to compress the network further. In addition, channel shuffling is introduced to the model to shuffle and reorganize all feature channels uniformly. Finally, this network obtains better segmentation performance and shorter segmentation time by using filter pruning via geometric median. Experimental results demonstrate that the network achieves great segmentation performance.

\cite{222} proposes a LadderNet-based segmentation method for finger vein images. The LadderNet is an improved network based on the conventional U-Net, which can fuse multi-path transmission information to obtain complex vein features. In this study, the venous features are extracted by the local maximum curvature detection method. These features are used to train LadderNet. The experimental results show that the LadderNet obtains a AUC of 91.56\% and 92.91\% on SDUMLA-HMT and MMCBNU-6000. 

\subsubsection{Segmentation based on other networks}
\begin{figure}[htbp]
    \centering
    \includegraphics[width = 0.8\textwidth]{ 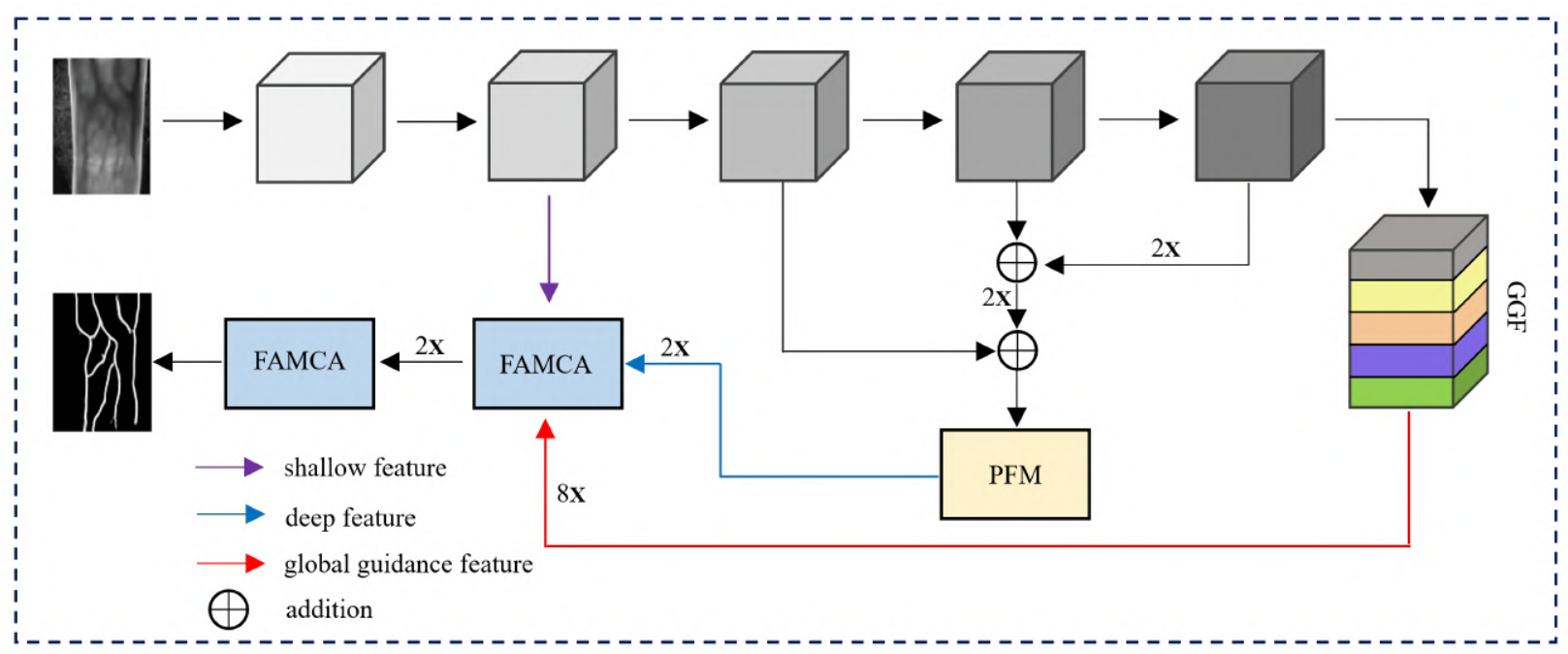}
    \caption{The framework of the proposed multi-scale feature representation method~\cite{225}.}
    \label{S1}
\end{figure}

\cite{218} evaluates three \emph{State-Of-The Art} (SOTA) semantic segmentation networks on finger vein image segmentation task, and they are Mask RCNN~\cite{219}, CCNet~\cite{220}, and HRNet~\cite{221}. In this research, the pre-processing steps only contain the resizing and normalization of the input images. The experimental results show that the performance of these networks is unstable when tested on different public datasets. \cite{225} designs a CNN-based multi-scale feature representation method for finger vein image segmentation, and the specific flow is shown in Fig.~\ref{S1}. The \emph{Global Guiding Feature} (GGF) model is used to extract multi-scale feature information and enables the vein features to be better separated from the background. The GGF model comprises four \emph{Local Similarity Pyramid} (LSP) models based on different scales. In addition, the network uses the \emph{Pyramid Fusion Module} (PFM) to enhance the multi-scale features, and this approach avoids the contextual information loss between different sub-regions. In the feature aggregation phase, this network uses the \emph{Feature Aggregation Module with Channel Attention} (FAMCA) to retain the important feature mappings and ignore the irrelevant ones. The network can automatically exploit multi-scale features in finger vein images to improve the segmentation performance by fusing GGF, PFM, and FAMCA models.

\subsection{Presentation attack detection}
\cite{226, 227} use CNN structure based on AlexNet for finger vein PAD. \cite{226} uses a pre-trained CNN model to extract features from the pre-processed finger vein images and uses PAC for feature extraction. Finally, the SVM is used to verify the authenticity of the images. This PAD method achieves zero error on IDIAP. \cite{227} extends the original AlexNet with seven layers to enrich the robustness of the model. Meanwhile, the study uses patches corresponding to real and fake images to fine-tune the network to suppress overfitting. This method performs well for inkjet printed artifacts with 3.48\% APCER and 0\% BPCER. 

\cite{228} proposes a DSC with residual structure and Linear SVM for finger vein PAD tasks. This is the first time DSC has been used for feature extraction in the finger vein PAD tasks. This feature extraction method allows more comprehensive processing of real-time scenes and makes the network more lightweight. In addition, residual modules are added to the DSC to prevent the gradient vanishing. Linear SVM is used for classification. The experimental results show that the error rate of this model on both IDIAP and SCUT is 0.00\%.

\cite{229, 230} perform PAD using a custom CNN structure. \cite{229} designs a lightweight CNN that consists of f two convolutional layers and pooling layers followed by two fully-connected layers. This network is not pre-trained in any way, but the training samples of the network are enriched by data augmentation directly on the original images. The method achieves an ACC of 100\% on both public datasets. \cite{230} uses a multi-task learning approach to integrate the recognition task and the PAD task into a united CNN model and designs an FVR system. In this system, the image with the most obvious vein information is selected by applying a multi-intensity illumination strategy, and if it passes the anti-spoofing detection, the features of that image are used for subsequent registration and recognition. The experiments illustrate the excellent performance of the system even on challenging databases with images depicting axial rotation.

\subsection{Template protection}
\cite{231} presents a template generation framework based on random projection and DBN, which is termed FVR-DLRP. In the FVR-DLRP, the features extracted from the original finger are transform the high-dimensional space to the low-dimensional space by random projection while generating a protected template by combining randomly generated keys. These templates are trained on the DBN, and the complexity of the DBN structure ensures the safety of these templates. The experimental results show that the FVR-DLRP achieves a recognition rate of 96.9\% and a \emph{False Acceptance Rate} (FAR) of 1.5\% on FV-NET64. \cite{232, 233} use CAE structure to learn deep features from the feature map generated by conventional FVR methods. Biohash generates a protection template for these features. This method performs better than the original biohash protection template in the stolen scenario. \cite{237} encodes the finger vein images into one-dimensional vectors, then encrypts the vectors and re-encodes the vectors into images using the \emph{Rivest–Shamir–Adleman} (RSP) algorithm, and finally feeds it into a modified ResNet containing the Squeeze-and-Excitation block for feature extraction and recognition. \cite{234} proposes a biometric protection algorithm based on the \emph{Binary Decision Diagram} (BDD)~\cite{235} and a \emph{Multi-Layer ELM} (ML-ELM)~\cite{236}. This BDD-based secure template generation algorithm extracts features from the pre-processed finger vein images using Gabor filters with LDA. These features are transformed into binary-based features, and BDD generates the protected templates of these features. Finally, these templates are fed into the ML-ELM for training. ML-ELM is a multi-layer ANN structure with a faster learning speed than other deep CNNs. The BDD-based protected template generation method can be applied to any binary-valued feature vector, and this method requires only a little storage space. The experimental results show that the method achieves CIRs of 93.09\%, 98.70\%, and 98.61\% on SDUMLA-HMT, MMCBNU-6000, and UTFVP.
 
\subsection{Multimodal biometric recognition}

\subsubsection{The fusion of two biometric information}
\cite{238, 239, 240,  241} use finger vein and another biometric information for recognition. \cite{238} uses finger shape and finger vein for recognition, where the finger shape image is a two-dimensional spectrogram image that expresses the change in the frequency component of the finger thickness based on the horizontal position of the finger. The finger shape image and the finger vein image are fed into a more lightweight ResNet with only four residual blocks to output the matching scores separately. Score level fusion is used for multimodal recognition of finger shape and finger vein. The experimental results show this multimodal recognition framework obtains EERs of 3.509\% and 1.706\% on SDUMLA-HMT and HKPU. \cite{241} performs score level fusion recognition of face and finger vein. The CNN structures used here are based on AlexNet, and the CNN for faces has two more convolutional layers and one more pooling layer than that for finger veins. This method achieves an ACC of 99.78\%.

\begin{figure}[htbp]
    \centering
    \includegraphics[width = 1.0\textwidth]{ 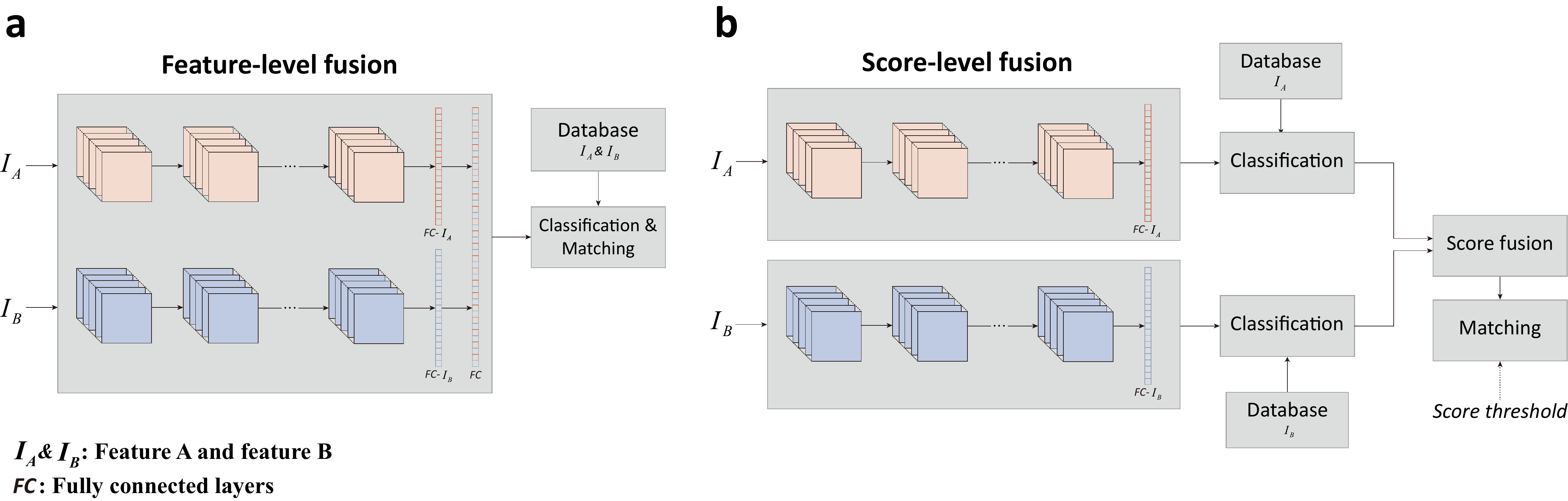}
    \caption{Two fusion methods on multimodal biometric recognition~\cite{239}. \textbf{(a)} feature level fusion method. \textbf{(b)} score level fusion method.}
    \label{MV1}
\end{figure}

\cite{239} proposes two methods for multimodal recognition regarding finger knuckle and finger vein. The respective specific processes of the two methods are shown in Fig.~\ref{MV1}. This study compares the recognition performance of three pre-trained CNN structures for finger knuckle print image and finger vein image: AlexNet, VGG-16, and ResNet-50. Among these networks, ResNet-50 achieves the best experimental results, with the ACC of 99.89\% in the score level fusion method and 98.84\% in the feature level fusion method. \cite{240} presents a multimodal biometric recognition system for \emph{Electrocardiogram} (ECG) and finger veins. A custom CNN containing nine convolutional layers is used to extract deep features, and the K-Nearest Neighbors is used for classification. In addition, this system uses multi-canonical correlation analysis to express deep features in low-dimensional space and accelerate the validation. The experimental results show this approach achieves EERs of 1.40\% on the score level fusion and 0.12\% on the feature level fusion.

\subsubsection{The fusion of various biometric information}
\cite{242, 243} performs multimodal biometric recognition using multiple features, including finger veins. \cite{242} aggregates fingerprints, finger veins, and faces for recognition using the score level fusion method. The CNN structure used for feature extraction and the three network structures used to detect the three biometric information are the same. The method related to finger vein uses random forest~\cite{244} as the classifier, and other biometric information uses Softmax. This multimodal recognition method achieves an ACC of 99.49\% on SDUMLA-HMT. \cite{243} designs a score level fusion recognition approach with iris, face, and finger vein, and the experiments of this study are also performed on SDUMLA-HMT. It uses pre-trained VGG-16 as the CNN structure to perform the classification, and this method achieves an ACC of 99.39\%.

\subsection{Other tasks of FVR} 
Finger vein image quality assessment is an essential part of FVR, and the finger vein images of unsatisfactory quality may lead to verification failure. \cite{245, 246, 247, 248} performs the quality assessment of finger vein images based on deep neural networks. \cite{245, 246} extracts binary finger vein patterns from grayscale images, feeds them into a deep CNN for training, and outputs the features using the last fully connected layer in this deep CNN. This deep feature is fed into the SVM to generate image quality scores. Experiments on two public datasets show that the CNN + SVM scheme can accurately discriminate between high and low-quality finger vein images. \cite{247} uses a lightweight CNN with only two convolutional layers and two fully connected layers for image quality assessment. In this approach, the segmented image sub-blocks are fed into the network instead of the entire image, and the network generates the respective quality scores. Finally, the quality score of the whole image is the average of all sub-blocks scores. \cite{248} uses the histogram of competitive Gabor responses~\cite{249} to label the training samples. Meanwhile, to compensate for the insufficient number of low-quality images in the training data, it uses the SMOTH method to perform the data augmentation. A custom CNN model trained by these samples achieves an ACC of 98.3\% on MMCBNU-6000.

\cite{251, 252} focus on the feature extraction step in FVR. \cite{251} proposes a finger vein feature extraction model based on fully CNN and \emph{Conditional Random Field} (CRF). DSC is added to fully CNN in this model to capture complex vein features by adaptively adjusting the received field. In addition, residual recurrent convolution is used to mine the deep features further. The CRF-RNN module is embedded in the model to output the feature maps. \cite{252} proposes a new loss function. It can dynamically adjust margins of different types to obtain more representative features. Experiments prove that the features extracted by this method have stronger geometric interpretations. \cite{253, 254} focus on ROI extraction of finger vein images. \cite{253} uses VGG-16 for ROI extraction of finger vein images. This research compares the feature maps outputted by each convolutional layer, and the experiment results show that the feature map output by the third convolutional layer converges the fastest and has the best performance. \cite{254} uses a capsule network to perform the ROI extraction task of FVR. The dynamic routing algorithm between capsules replaces the pooling layer in CNN, thus avoiding the loss of location information due to pooling operations. This method not only improves the efficiency of ROI extraction but also simplifies the network model compared to deep CNN.

\cite{255} proposes a GAN-based image generation method to compensate for the lack of finger vein images in public datasets by generating finger vein images. This network generates new finger vein images from the segmented finger vein image by convolution and deconvolution operations, adding residual blocks to suppress overfitting. The images generated by this method are clearer than the original images in the datasets. \cite{257} uses a modified Pix2Pix model to generate grayscale finger vein images from binary templates. The generator network uses a U-Net structure to generate the vein images and adds residual blocks between the up-sampled and down-sampled convolution layers. This method can be used for data reconstruction across datasets rather than just intra-class data augmentation.

\cite{259, 260} explore the similarities between finger veins from the same people. To investigate the vein similarity between the left and right hand symmetrical fingers, a pair of symmetrical fingers from the same subject is considered a class in~\cite{259}. DenseNet-201 is used to train the manually extracted features. This paper experimentally considers that although there is some similarity between symmetrical fingers of the same people, this similarity is not sufficient for applicants on FVR. As an extension work, \cite{260} introduces the triple loss into the CNN model, enabling vein similarities between symmetric fingers to be successfully detected.

\subsection{Summary}
Compared with classical neural networks, deep neural networks perform more tasks in the field of FVR. DL is now the mainstream technology used in the FVR. To illustrate the application of deep neural networks on FVR, we provide a summary table of related papers in Tab.~\ref{DNNT}.

Verification tasks are the most frequent tasks involved in FVR. Typical structures, AlexNet, VGGNet, ResNet, etc. have achieved excellent results in the verification task. In addition, many custom CNN structures have also obtained satisfactory performance. The construction of 3D finger vein images provides a desirable prescription for finger posture change problems. GANs are the mainstream network in image enhancement tasks on FVR due to their powerful image generation ability. The CAEs are also performed image enhancement by the convolutional layers in the encoder and de-convolutional layers in the decoder. Like many other computer vision tasks, U-Net is the most commonly used network for performing segmentation tasks in FVR. However, the number of papers related to segmentation tasks is scarce, and this research direction needs further exploration. The main task of the PAD on FVR is to identify the printed finger vein images and the real collected finger vein images. Many research works have achieved complete success on this task. The study of the template protection tasks helps prevent data leakage from the FVR system and fully safeguards users' privacy. 

\begin{sidewaystable}[htbp]
    \centering
    \caption{Summary of the FVR based on deep neural networks.}
    \label{DNNT}
    \resizebox{\linewidth}{!}{\begin{tabular}{ccllll}
        \toprule
        \textbf{Year} & \textbf{Reference} & \textbf{Task} & \textbf{Network} & \textbf{Dataset} & \textbf{Performance} \\
        \specialrule{0em}{3pt}{3pt}
        \hline
        \specialrule{0em}{4pt}{4pt}
        2015 & \cite{126} & Verification & DBN & Private (IN:6000 SN:64) & ACC = 96.9\% \; EER = 1.5\% \\
        \specialrule{0em}{4pt}{4pt}
        2015 & \cite{245} & Image Q-Assessment & Custom & \makecell[l]{HKPU \\ FV-USM} & \makecell[l]{ACC = 88.89\% (HI) \; ACC = 88.18\% (LI) \\ ACC = 74.98\% (HI) \; ACC = 70.07\% (LI)} \\
        \specialrule{0em}{4pt}{4pt}
        2016 & \cite{130} & Verification & LeNet-5 & Private & ACC = 96.78\% \\
        \specialrule{0em}{4pt}{4pt}
        2016 & \cite{159} & Verification & Constom & UTFVP & ACC = 100\% \\
        \specialrule{0em}{4pt}{4pt}
        2017 & \cite{99} & Verification & AlexNet & Private & ACC = 99.4\% \; EER = 0.21\% \\
        \specialrule{0em}{4pt}{4pt}
        2017 & \cite{113} & Verification & VGG-16 & \makecell[l]{SDUMLA-HMT \\ Private1 (IN:1200 SN:20) \\ Private2 (IN:1980 SN:33)} & \makecell[l]{EER = 0.396\% \\ EER = 1.275\% \\ EER = 3.906\%} \\
        \specialrule{0em}{4pt}{4pt}
        2017 & \cite{114} & Verification & VGG-16 & \makecell[l]{Private1 (Outdoor continuous acquisition)  \\ Private2 (Indoor intermittent acquisition)  \\ Private3 (Indoor continuous acquisition ) \\ } & \makecell[l]{EER = 0.42\% \\ EER = 1.41\% \\ EER = 2.14\%} \\
        \specialrule{0em}{4pt}{4pt}
        2017 & \cite{229} & PAD & Custom & \makecell[l]{IDIAP \\ SCUT} & \makecell[l]{ACER = 0.00\% \\ ACER = 0.00\%} \\
        \specialrule{0em}{4pt}{4pt}
        2017 & \cite{101} & Verification & AlexNet & \makecell[l]{SDUMLA-HMT \\ Private (IN:2970 \; SN:198)} & \makecell[l]{EER = 0.80\% \\ EER = 0.079\%} \\
        \specialrule{0em}{4pt}{4pt}
        2017 & \cite{127} & Verification & DBN Custom & Private (IN:960 SN:15) & ACC = 99.6\% \\
        \specialrule{0em}{4pt}{4pt}
        2017 & \cite{246} & Image Q-Assessment & Custom & \makecell[l]{FV-USM \\ HKPU} & \makecell[l]{ACC = 71.01\% (HI) \; ACC = 73.57\% (LI) \\ ACC = 87.08\% (HI) \; ACC = 86.36\%} \\
        \specialrule{0em}{4pt}{4pt}
        2017 & \cite{258} & Image enhancement & Custom & \makecell[l]{HKPU \\ FV-USM} & \makecell[l]{EER = 2.70\% \\ EER = 1.42\%} \\
        \bottomrule
    \end{tabular}}
\end{sidewaystable}

\begin{sidewaystable}[htbp]
    \renewcommand{\thetable}{5}
    \caption{Continued: Summary of the FVR based on deep neural networks.}
    \centering
    \resizebox{\linewidth}{!}{\begin{tabular}{ccllll}
        \toprule
        \textbf{Year} & \textbf{Reference} & \textbf{Task} & \textbf{Network} & \textbf{Dataset} & \textbf{Performance} \\
        \specialrule{0em}{3pt}{3pt}
        \hline
        \specialrule{0em}{4pt}{4pt}
        2017 & \cite{226} & PAD & \makecell[l]{AlexNet \\ VGG-16} & \makecell[l]{IDIAP \\ ISPR} & \makecell[l]{APCER = 0.2018\% \; BPCER = 0.1863\% \; ACER = 0.1940\% \\ APCER = 0.0000\% \; BPCER = 0.1240\% \; ACER = 0.0620\%} \\
        \specialrule{0em}{4pt}{4pt}
        2017 & \cite{227} & PAD & AlexNet & Private & \makecell[l]{APCER = 3.48\% \; BPCER = 0\% (Inkjet printed) \\ APCER = 0\% \; BPCER = 0\% (Laserjet printed)} \\
        \specialrule{0em}{4pt}{4pt}
        2018 & \cite{208} & Image enhancement & CAE & Private (IN:5850) & \makecell[l]{PSNR = 22.86 dB (Single irregular-region) \\ PSNR = 21.65 dB (Multiple irregular-region) \\ PSNR = 24.01 dB (Square-region)} \\
        \specialrule{0em}{4pt}{4pt}
        2018 & \cite{167} & Verification & Two stream CNN & \makecell[l]{MMCBNU-6000 \\ SDUMLA-HMT} & \makecell[l]{EER = 0.10\% \\ EER = 0.47\%} \\
        \specialrule{0em}{4pt}{4pt}
        2018 & \cite{174} & Verification & Custom & \makecell[l]{HKPU \\ FV-USM \\ SDUMLA-HMT \\ UTFVP} & \makecell[l]{ACC = 95.32\% \\ ACC = 97.53\% \\ ACC = 97.48\% \\ ACC = 95.56\%} \\
        \specialrule{0em}{4pt}{4pt}
        2018 & \cite{175} & Verification & Custom & Private (IN:1200) & ACC = 100\% \\
        \specialrule{0em}{4pt}{4pt}
        2018 & \cite{238} & Verification & ResNet & \makecell[l]{SDUMLA-HMT \\ HKPU} & \makecell[l]{EER = 2.4088\% \\ EER = 0.8255\%} \\
        \specialrule{0em}{4pt}{4pt}
        2018 & \cite{215} & Segmentation & \makecell[l]{U-Net \\ RefineNet \\ SegNet} & \makecell[l]{UTFVP \\ SDUMLA-HMT} & \makecell[l]{EER = 0.46\% \; FMR = 0.60\% \; ZFMR = 1.75\% (U-Net) \\ EER = 0.27\% \; FMR = 0.27\% \; ZFMR = 0.46\% (RefineNet) \\ EER = 1.20\% \; FMR = 2.87\% \; ZFMR = 4.44\% (SegNet) \\ EER = 6.15\% \; FMR = 13.37\% \; ZFMR = 16.80\% (U-Net) \\ EER = 2.45\% \; FMR = 5.87\% \; ZFMR = 8.65\% (RefineNet) \\ EER = 5.50\% \; FMR = 12.63\% \; ZFMR = 16.34\% (SegNet)} \\
        \specialrule{0em}{4pt}{4pt}
        2018 & \cite{231} & Template protection & DBN & FV-NET64 & GAR = 96.9\% \; FAR = 1.5\% \\    
        \specialrule{0em}{4pt}{4pt}
        \bottomrule
    \end{tabular}}
\end{sidewaystable}

\begin{sidewaystable}[htbp]
    \renewcommand{\thetable}{5}
    \caption{Continued: Summary of the FVR based on deep neural networks.}
    \centering
    \resizebox{\linewidth}{!}{\begin{tabular}{ccllll}
        \toprule
        \textbf{Year} & \textbf{Reference} & \textbf{Task} & \textbf{Network} & \textbf{Dataset} & \textbf{Performance} \\
        \specialrule{0em}{3pt}{3pt}
        \hline
        \specialrule{0em}{4pt}{4pt}
        2018 & \cite{123} & Verification & CAE & FV-USM & ACC = 99.49\% \; EER = 0.16\% \\
        \specialrule{0em}{4pt}{4pt}
        2018 & \cite{102} & Verification & AlexNet & Private & ACC = 95\% \\
        \specialrule{0em}{4pt}{4pt}
        2018 & \cite{180} &Verification & Custom & \makecell[l]{MMCBNU-6000 \\ FV-USM \\ SDUMLA-HMT} & \makecell[l]{EER = 0.04\% (Closed-set) \; EER = 0.30\% (Open-set) \\ EER = 0.06\% (Closed-set) \; EER = 0.76\% (Open-set) \\ EER = 0.46\% (Closed-set) \; EER = 1.20\% (Open-set)} \\
        \specialrule{0em}{4pt}{4pt}
        2018 & \cite{247} & Image Q-Assessment & Lightweight CNN & \makecell[l]{MMCBNU-6000 \\ SDUMLA-HMT} & \makecell[l]{ACC = 71.95\% \\ ACC = 74.63\%} \\
        \specialrule{0em}{4pt}{4pt}
        2019 & \cite{176} & Verification & Custom & \makecell[l]{SDUMLA-HMT \\ HKPU \\ FV-USM \\ MMCBNU-6000} & \makecell[l]{ACC = 99.21\% \\ ACC = 99.61\% \\
        ACC = 98.44\% \\ ACC = 98.58\%} \\
        \specialrule{0em}{4pt}{4pt}
        2019 & \cite{211} & Image enhancement & Custom & HKPU & PSNR = 29.638 dB \\
        \specialrule{0em}{4pt}{4pt}
        2019 & \cite{147} & Verification & \makecell[l]{Lightweight CNN \\ Custom1 (Triplet loss) \\ Custom2 (Joint Bayesian) \\ VGG-16} & HKPU & \makecell[l]{EER = 0.1497\% \\ EER = 0.1316\% \\ EER = 0.1327\% \\ EER = 0.1223\%} \\
        \specialrule{0em}{4pt}{4pt}
        2019 & \cite{170} & Verification & Capsule network & \makecell[l]{SDUMLA-HMT \\ UTFVP \\ HKPU \\ MMCBNU-6000} & \makecell[l]{100\% \\ 94\% \\ 88\% \\ 100\%} \\
        \specialrule{0em}{4pt}{4pt}
        2019 & \cite{178} & Verification & Custom & Private (IN:3000 SN:1000) & ACC = 96.8\% \\
        \specialrule{0em}{4pt}{4pt}
        2019 & \cite{261} & Verification & Custom & \makecell[l]{UTFVP \\ SDUMLA-HMT} & \makecell[l]{EER =  0.0614\% \\ EER = 0.0395\%} \\
        \specialrule{0em}{4pt}{4pt}
        \bottomrule
    \end{tabular}}
\end{sidewaystable}

\begin{sidewaystable}[htbp]
    \renewcommand{\thetable}{5}
    \caption{Continued: Summary of the FVR based on deep neural networks.}
    \centering
    \resizebox{\linewidth}{!}{\begin{tabular}{ccllll}
        \toprule
        \textbf{Year} & \textbf{Reference} & \textbf{Task} & \textbf{Network} & \textbf{Dataset} & \textbf{Performance} \\
        \specialrule{0em}{4pt}{4pt}
        \hline
        \specialrule{0em}{4pt}{4pt}
        2019 & \cite{213} & Image enhancement & PCNN & \makecell[l]{NJUST-FV \\ SDUMLA-HMT \\ THU-FVFDT1 \\ HKPU} & \makecell[l]{
        Gradient in spatial = 0.6087 \; Grey contrast = 31.5743 \\ Information capacity = 9.2709 \; Deep evaluator = 6.4869 \\ 
        Gradient in spatial = 0.6855 \; Grey contrast = 23.9111 \\ Information capacity = 6.1768 \; Deep evaluator = 6.1851 \\
        Gradient in spatial = 0.576 \; Grey contrast = 23.8313 \\ Information capacity = 7.3566 \; Deep evaluator = 6.6593 \\
        Gradient in spatial = 23.1059 \; Grey contrast = 0.6294 \\ Information capacity = 3.6167 \; Deep evaluator = 6.0966 \\
        } \\
        \specialrule{0em}{4pt}{4pt}
        2019 & \cite{250} & Similarity explosion & Custom & \makecell[l]{MMCBNU-6000 \\ SDUMLA-HMT} & \makecell[l]{EER = 0.74\% \\ EER = 2.37\%} \\
        \specialrule{0em}{4pt}{4pt}
        2019 & \cite{128} & Verification & DBN & Private (IN:3000 SN:1000) & ACC = 97.4\% \\
        \specialrule{0em}{4pt}{4pt}
        2019 & \cite{143} & Verification & LSTM & HKPU & EER = 0.95\% \\
        \specialrule{0em}{4pt}{4pt}
        2019 & \cite{185} & Verification & Custom & \makecell[l]{(Train) MMCBNU-6000 \\ (Test) Private (IN:9888 SN:103) } & ACC = 86.49\% \; AUC = 0.9337 \\
        \specialrule{0em}{4pt}{4pt}
        2019 & \cite{131} & Verification & Siamese CNN & \makecell[l]{MMCBNU-6000 \\ FV-USM \\ SDUMLA-HMT \\ SCUT} & \makecell[l]{EER = 0.12\% \\ EER = 0.30\% \\ EER = 0.66\% \\ EER = 2.78\%} \\
        \specialrule{0em}{4pt}{4pt}
        2019 & \cite{103} & Verification & AlexNet & Private (IN:100 SN:12) & \makecell[l]{ACC = 91.67\% \\ AUC = 0.942} \\
        \specialrule{0em}{4pt}{4pt}
        2019 & \cite{216} & Segmentation & \makecell[l]{U-Net \\ RefineNet \\ SegNet} & UTFVP & \makecell[l]{EER = 0.64\% \; FMR = 1.85\% \;  ZFMR = 3.47\% \\ EER = 1.76\% \; FMR = 4.12\% \; ZFMR = 6.34\% \\ EER = 2.21\% \; FMR = 6.20\% \; ZFMR = 11.25\%} \\
        \specialrule{0em}{4pt}{4pt}
        
        \bottomrule
    \end{tabular}}
\end{sidewaystable}

\begin{sidewaystable}[htbp]
    \renewcommand{\thetable}{5}
    \caption{Continued: Summary of the FVR based on deep neural networks.}
    \centering
    \resizebox{\linewidth}{!}{\begin{tabular}{ccllll}
        \toprule
        \textbf{Year} & \textbf{Reference} & \textbf{Task} & \textbf{Network} & \textbf{Dataset} & \textbf{Performance} \\
        \specialrule{0em}{4pt}{4pt}
        \hline
        \specialrule{0em}{4pt}{4pt}
        2019 & \cite{191} & Verification & Custom & \makecell[l]{Private (IN:500 \; SN:50)} & \makecell[l]{ACC = 98.64\% \\ ACC = 98.95\%} \\
        \specialrule{0em}{4pt}{4pt}
        2019 & \cite{120} & Verification & GAN & \makecell[l]{SDUMLA-HMT \\ THU-FVFDT2} & \makecell[l]{EER = 0.94\% \\ EER = 1.12\%} \\
        \specialrule{0em}{4pt}{4pt}
        2019 & \cite{262} & Image enhancement & GAN & HKPU & EER = 0.87\% \\
        \specialrule{0em}{4pt}{4pt}
        2019 & \cite{172} & Verification & GNN & \makecell[l]{MMCBNU-6000 \\ SDUMLA-HMT} & \makecell[l]{ACC = 99.98\% \\ ACC = 99.98\%} \\
        \specialrule{0em}{4pt}{4pt}
        2019 & \cite{124} & Verification & CAE & \makecell[l]{FV-USM \\ SDUMLA-HMT} & \makecell[l]{ACC = 99.95\% \; EER = 0.12\% \\ ACC = 99.78\% \; EER = 0.21\%} \\
        \specialrule{0em}{4pt}{4pt}
        2019 & \cite{181} & Verification & Custom & \makecell[l]{FV-USM \\ SDUMLA-HMT \\ THU-FVFDT2} & \makecell[l]{ACC = 99.49\% \\ ACC = 98.19\% \\ ACC = 100\%} \\
        \specialrule{0em}{4pt}{4pt}
        2019 & \cite{116} & Verification & SqNet & \makecell[l]{MMCBNU-6000 \\ SDUMLA-HMT} & \makecell[l]{EER = 1.89\% \\ EER = 4.91\%} \\
        \specialrule{0em}{4pt}{4pt}
        2019 & \cite{209} & Image enhancement & CAE & Private (IN:5850 \; 585) & EER = 0.16\% \\
        \specialrule{0em}{4pt}{4pt}
        2019 & \cite{222} & Segmentation & LadderNet & \makecell[l]{SDUMLA-HMT \\ MMCBNU-6000} & \makecell[l]{ACC = 92.44\% \; AUC = 91.56\% \\ ACC = 93.93\% \; AUC = 92.91\%} \\
        \specialrule{0em}{4pt}{4pt}
        2019 & \cite{177} & Verification & Custom & FV-USM & ACC = 82.45\% (T.S. 1) \; ACC = 99.52\% (T.S. 2) \\
        \specialrule{0em}{4pt}{4pt}
        \bottomrule
    \end{tabular}}
\end{sidewaystable}

\begin{sidewaystable}[htbp]
    \renewcommand{\thetable}{5}
    \caption{Continued: Summary of the FVR based on deep neural networks.}
    \centering
    \resizebox{\linewidth}{!}{\begin{tabular}{ccllll}
        \toprule
        \textbf{Year} & \textbf{Reference} & \textbf{Task} & \textbf{Network} & \textbf{Dataset} & \textbf{Performance} \\
        \specialrule{0em}{3pt}{3pt}
        \hline
        \specialrule{0em}{4pt}{4pt}
        2019 & \cite{222} & Segmentation & LadderNet & \makecell[l]{SDUMLA-HMT \\ MMCBNU-6000} & \makecell[l]{ACC = 92.44\% \; AUC = 91.56\% \\ ACC = 93.93\% \; AUC = 92.91\%} \\
        \specialrule{0em}{4pt}{4pt}
        2019 & \cite{186} & Verification & Custom & \makecell[l]{MMCBNU-6000 \\ FV-USM \\ SDUMLA-HMT \\ HKPU} & \makecell[l]{EER = 0.11\% \\ EER = 0.57\% \\ EER = 1.09\% \\ EER = 1.67\%} \\
        \specialrule{0em}{4pt}{4pt}
        2019 & \cite{259} & Verification & DenseNet & SDUMLA-HMT & EER = 0.54\% \\
        \specialrule{0em}{4pt}{4pt}
        2019 & \cite{161} & Verification & DenseNet & \makecell[l]{SDUMLA-HMT \\ HKPU} & \makecell[l]{EER = 2.35\% \\ EER = 0.33\%} \\
        \specialrule{0em}{4pt}{4pt}
        2019 & \cite{188} & Verification & Custom & \makecell[l]{Private (3D Image IN:8526 SN:203) \\ FV-USM \\ SDUMLA-HMT \\ HKPU} & \makecell[l]{EER =  2.84\% \\ EER = 0.94\% \\ EER = 1.69\% \\ EER = 2.40\%} \\
        \specialrule{0em}{4pt}{4pt}
        2020 & \cite{165} & Verification & ShuffleNet & \makecell[l]{SDUMLA-HMT \\ FV-USM \\ MMCBNU-6000} & \makecell[l]{EER = 0.37\% \\ EER = 0.31\% \\ EER = 0.05\%} \\
        \specialrule{0em}{4pt}{4pt}
        2020 & \cite{192} & Verification & Custom & \makecell[l]{FV-USM \\ SDUMLA-HMT \\ THU-FVFDT2} & \makecell[l]{ACC = 96.15\% \\ ACC = 99.48\% \\ ACC = 99.56\%} \\
        \specialrule{0em}{4pt}{4pt}
        2020 & \cite{125} & Verification & CAE & \makecell[l]{FV-USM \\ SDUMLA-HMT} & \makecell[l]{ACC = 98.02\% (CAE) \; ACC = 98.88\% (CAE+ELM) \\ ACC = 97.86\% (CAE) \; ACC = 98.58\% (CAE+ELM)} \\
        \specialrule{0em}{4pt}{4pt}
        \bottomrule
    \end{tabular}}
\end{sidewaystable}

\begin{sidewaystable}[htbp]
    \renewcommand{\thetable}{5}
    \caption{Continued: Summary of the FVR based on deep neural networks.}
    \centering
    \resizebox{\linewidth}{!}{\begin{tabular}{ccllll}
        \toprule
        \textbf{Year} & \textbf{Reference} & \textbf{Task} & \textbf{Network} & \textbf{Dataset} & \textbf{Performance} \\
        \specialrule{0em}{3pt}{3pt}
        \hline
        \specialrule{0em}{4pt}{4pt}
        2020 & \cite{179} & Verification & Custom & \makecell[l]{Private (IN:3000 SN:1000) \\ FV-USM} & \makecell[l]{ACC = 98.4\% \\ ACC = 98.0\%} \\
        \specialrule{0em}{4pt}{4pt}
        2020 & \cite{190} & Verification & Custom & FV-USM & ACC = 95.1\% \; EER = 0.0373\% \\
        \specialrule{0em}{4pt}{4pt}
        2020 & \cite{145} & Verification & LSTM \; Custom CNN & Private (IN:9000 SN:100) & ACC = 99.13\% \\
        \specialrule{0em}{4pt}{4pt}
        2020 & \cite{201} & Verification & GAN & \makecell[l]{HKPU \\ SDUMLA-HMT} & \makecell[l]{EER = 1.814\% \\ EER = 3.934\%} \\
        \specialrule{0em}{4pt}{4pt}
        2020 & \cite{251} & Segmentation & U-Net & \makecell[l]{SDUMLA-HMT \\ MMCBNU-6000 \\ HKPU} & \makecell[l]{EER = 5.827\% \\ EER = 0.364\% \\ EER = 2.372\%} \\
        \specialrule{0em}{4pt}{4pt}
        2020 & \cite{223} & Segmentation & U-Net & \makecell[l]{SDUMLA-HMT \\ MMCBNU-6000} & \makecell[l]{EER = 0.1546\% \\ EER = 0.1556\%} \\
        \specialrule{0em}{4pt}{4pt}
        2020 & \cite{230} & PAD & Custom CNN & \makecell[l]{IDIAP \\ SCUT} & \makecell[l]{EER = 5.61\% (Recognition) \; HTER = 0.00\% (PAD) \\ EER = 2.18\% (Recognition) \; HTER = 0.00\% (PAD)} \\
        \specialrule{0em}{4pt}{4pt}
        2020 & \cite{203} & Image enhancement & GAN & \makecell[l]{MMCBNU-6000 \\ FV-USM} & \makecell[l]{EER = 5.66\% \\ EER = 2.37\%} \\
        \specialrule{0em}{4pt}{4pt}
        2020 & \cite{248} & Image Q-Assessment & Custom & MMCBNU-6000 & ACC = 98.3\% \\
        \specialrule{0em}{4pt}{4pt}
        2020 & \cite{242} & M-Biometric recognition & Custom CNN & SDUMLA-HMT & ACC = 99.49\% \\
        \specialrule{0em}{4pt}{4pt}
        2020 & \cite{148} & Verification & ZFNet & SDUMLA-HMT & ACC = 86\% \\
        \specialrule{0em}{4pt}{4pt}
        2020 & \cite{253} & ROI extraction & VGG-16 & THU-FVFDT1  UTFVP HKPU & EER = 0.1987\% \; IOU = 94.95\% \\
        \specialrule{0em}{4pt}{4pt}
        2020 & - & Verification & Custom CNN & Private & ACC = 100\% \\
        \specialrule{0em}{4pt}{4pt}
        \bottomrule
    \end{tabular}}
\end{sidewaystable}

\begin{sidewaystable}[htbp]
    \renewcommand{\thetable}{5}
    \caption{Continued: Summary of the FVR based on deep neural networks.}
    \centering
    \resizebox{\linewidth}{!}{\begin{tabular}{ccllll}
        \toprule
        \textbf{Year} & \textbf{Reference} & \textbf{Task} & \textbf{Network} & \textbf{Dataset} & \textbf{Performance} \\
        \specialrule{0em}{3pt}{3pt}
        \hline
        \specialrule{0em}{4pt}{4pt}
        2020 & \cite{243} & M-Biometric recognition & VGG-16 & SDUMLA-HMT & ACC = 99.39\% (FLF) \; ACC = 100\% (SLF) \\
        \specialrule{0em}{4pt}{4pt}
        2020 & \cite{239} & M-Biometric recognition & \makecell[l]{AlexNet \\ VGG-16 \\ ResNet-50} & SDUMLA-HMT &\makecell[l]{
        ACC = 79.03\% (Softmax) \; 76.77\% (SVM) \; EER = 0.2925\% (Softmax) \; 0.322\% (SVM) \\
        ACC = 85.04\% (Softmax) \; 79.96\% (SVM) \; EER = 0.1490\% (Softmax) \; 0.204\% (SVM) \\
        ACC = 98.58\% (Softmax) \; 93.34\% (SVM) \; EER = 0.0142\% (Softmax) \; 0.066\% (SVM)} \\
        \specialrule{0em}{4pt}{4pt}
        2020 & \cite{162} & Verification & DenseNet & \makecell[l]{HKPU \\ SDUMLA-HMT} & \makecell[l]{EER = 0.05\% \\ EER = 1.65\%} \\
        \specialrule{0em}{4pt}{4pt}
        2020 & \cite{153} & Verification & Lightweight CNN & \makecell[l]{MMCBNU-6000 \\ FV-USM} & \makecell[l]{ACC = 99.05\% \; EER = 0.503\% \\ ACC = 97.95\% \; EER = 1.070\%} \\
        \specialrule{0em}{4pt}{4pt}
        2020 & \cite{217} & Segmentation & \makecell[l]{U-Net \\ RefineNet \\ SegNet} & UTFVP & \makecell[l]{
        EER = 0.462\% \; FMR = 0.787\% \; ZFMR = 1.388\% \\ 
        EER = 1.437\% \; FMR = 2.870\% \; ZFMR = 4.444\% \\
        EER = 0.686\% \; FMR = 1.388\% \; ZFMR = 2.916\%} \\
        \specialrule{0em}{4pt}{4pt}
        2020 & \cite{107} & Verification & \makecell[l]{U-Net ResNet} & \makecell[l]{SDUMLA-HMT \\ THU-FVFDT2} & \makecell[l]{ACC = 99.53\% \\ ACC = 98.4\%} \\
        \specialrule{0em}{4pt}{4pt}
        2020 & \cite{255} & Image generation & GAN & MMCBNU-6000 & EER = 0.24\% \\
        \specialrule{0em}{4pt}{4pt}
        2021 & \cite{210} & Image enhancement & VGG-16 & \makecell[l]{SDUMLA-HMT \\ FV-USM \\ UTFVP \\ THU-FVFDT1} & \makecell[l]{ACC = 99.84\% \\ ACC = 98.6\% \\ ACC = 99.10\% \\ ACC = 98.9\%} \\
        \specialrule{0em}{4pt}{4pt}
        2021 & \cite{237} & Template protection & Custom CNN & \makecell[l]{SDUMLA-HMT \\ MMCBNU-6000 \\ HKPU \\ FV-USM} & \makecell[l]{
        EER = 2.451\% \; CIR = 95.912\% (LBP) \; EER = 2.137\% \; CIR = 96.698\% (One-D vector) \\
        EER = 0.232\% \; CIR = 99.100\% (LBP) \; EER = 0.090\% \; CIR = 99.667\% (One-D vector) \\
        EER = 0.277\% \; CIR = 99.038\% (LBP) \; EER = 0.312\% \; CIR = 98.397\% (One-D vector) \\
        EER = 0.091\% \; CIR = 99.593\% (LBP) \; EER = 0.214\% \; CIR = 99.187\% (One-D vector)} \\
        \specialrule{0em}{4pt}{4pt}
        2021 & \cite{154} & Verification & Lightweight CNN & \makecell[l]{SDUMLA-HMT \\ PKU-FVD} & \makecell[l]{EER = 1.13\% \; ACC = 99.3\% \\ EER = 0.67\% \; ACC = 99.6\%} \\
        \specialrule{0em}{4pt}{4pt}
        2021 & \cite{155} & Verification & Lightweight CNN & \makecell[l]{SDUMLA-HMT \\ MMCBNU-6000} & \makecell[l]{EER = 2.29\% \\ EER = 0.47\%} \\
        \specialrule{0em}{4pt}{4pt}
        \bottomrule
    \end{tabular}}
\end{sidewaystable}

\begin{sidewaystable}[htbp]
    \renewcommand{\thetable}{5}
    \caption{Continued: Summary of the FVR based on deep neural networks.}
    \centering
    \resizebox{\linewidth}{!}{\begin{tabular}{ccllll}
        \toprule
        \textbf{Year} & \textbf{Reference} & \textbf{Task} & \textbf{Network} & \textbf{Dataset} & \textbf{Performance} \\
        \specialrule{0em}{3pt}{3pt}
        \hline
        \specialrule{0em}{4pt}{4pt}
        2021 & \cite{193} & Verification & Custom CNN & \makecell[l]{SDUMLA-HMT \\ MMCBNU-6000 \\ FV-USM \\ SCUT} & \makecell[l]{EER = 1.18\% \\ EER = 0.23\% \\ EER = 0.49\% \\ EER = 0.86\%} \\
        \specialrule{0em}{4pt}{4pt}
        2021 & \cite{194} & Verification & Custom CNN & \makecell[l]{FV-USM \\ SDUMLA-HMT} & \makecell[l]{ACC = 100\% \\ ACC = 99.82\%} \\ 
        \specialrule{0em}{4pt}{4pt}
        2021 & \cite{158} & Verification & GAN & SDUMLA-HMT  HKPU & \makecell[l]{EER =  0.85\% (Train: SDUMLA-HMT \; Test: HKPU) \\ EER = 3.40\% (Train: HKPU \; Test: SDUMLA-HMT)} \\
        \specialrule{0em}{4pt}{4pt}
        2021 & \cite{254} & ROI extraction & Capsule network & \makecell[l]{FV-USM \\ SDUMLA-HMT} & \makecell[l]{Pixel recognition rate = 99.7\% \\ Pixel recognition rate = 97.5\%} \\
        \specialrule{0em}{4pt}{4pt}
        2021 & \cite{212} & Image enhancement & Custom CNN & \makecell[l]{Private (IN:5850 SN:585) \\ SDUMLA-HMT} & \makecell[l]{
        PSNR = 15.9569dB \; SSIM = 0.8896 \; Scoot = 0.7102 \\
        PSNR = 13.5929dB \; SSIM = 0.7947 \; Scoot = 0.8083} \\
        \specialrule{0em}{4pt}{4pt}
        2021 & \cite{100} & Verification & AlexNet & \makecell[l]{FV-USM \\ SDUMLA-HMT} & \makecell[l]{ACC = 97.93\% \\ ACC = 97.71\%} \\
        \specialrule{0em}{4pt}{4pt}
        2021 & \cite{117} & Verification & SqNet & \makecell[l]{FV-USM \\ SDUMLA-HMT} & \makecell[l]{ACC = 89.35\% \\ ACC = 99.81\%} \\
        \specialrule{0em}{4pt}{4pt}
        2021 & \cite{44} & Verification & AlexNet & \makecell[l]{FV-USM \\ SDUMLA-HMT} & \makecell[l]{ACC = 97.35\% \\ ACC = 92.29\%} \\
        \specialrule{0em}{4pt}{4pt}
        2021 & \cite{241} & M-Biometric recognition & AlexNet & \makecell[l]{FV-USM \\ SDUMLA-HMT} & \makecell[l]{EER = 0.050\% \; ACC = 99.85\% \\ EER = 2.610\% \; ACC = 94.87\%} \\
        \specialrule{0em}{4pt}{4pt}
        \bottomrule
    \end{tabular}}
\end{sidewaystable}

\begin{sidewaystable}[htbp]
    \renewcommand{\thetable}{5}
    \caption{Continued: Summary of the FVR based on deep neural networks.}
    \centering
    \resizebox{\linewidth}{!}{\begin{tabular}{ccllll}
        \toprule
        \textbf{Year} & \textbf{Reference} & \textbf{Task} & \textbf{Network} & \textbf{Dataset} & \textbf{Performance} \\
        \specialrule{0em}{3pt}{3pt}
        \hline
        \specialrule{0em}{4pt}{4pt}
        2021 & \cite{144} & Verification & LSTM & Private & ACC = 99.93\% \; EER = 0.07\% \\
        \specialrule{0em}{4pt}{4pt}
        2021 & \cite{260} & Similarity explosion & \makecell[l]{SqNet \\ LCNN \\ ResNet} & \makecell[l]{SDUMLA-HMT \\ UTFVP \\ PLUSVein-FV3 \\ HKPU} & \makecell[l]{
        EER = 2.7\% (SqNet) \; 3.1\% (ResNet) \; 4.9\% (LCNN) \\ 
        EER = 2.5\% (SqNet) \; 3.6\% (ResNet) \; 4.6\% (LCNN) \\ 
        EER = 2.4\% (SqNet) \; 3.2\% (ResNet) \; 4.7\% (LCNN) \\
        EER = 3.7\% (SqNet) \; 5.6\% (ResNet) \; 10.0\% (LCNN)} \\
        \specialrule{0em}{4pt}{4pt}
        2021 & \cite{115} & Verification & \makecell[l]{AlexNet \\ SqNet \\ GoogleNet} & SDUMLA-HMT & \makecell[l]{ACC = 82.17\% \\ ACC = 87.06\% \\ ACC = 92.22\%} \\
        \specialrule{0em}{4pt}{4pt}
        2021 & \cite{214} & Image enhancement & Custom CNN & SDUMLA-HMT & ACC = 94.87\% \\
        \specialrule{0em}{4pt}{4pt}
        2021 & \cite{199} & Image enhancement & Custom CNN & \makecell[l]{(Train) Private1 (IN:427 SN:960) \\ (Test) Private2 (IN:720 SN:72)} & PSNR = 30.42dB \; SSIM = 0.9885 \\
        \specialrule{0em}{4pt}{4pt}
        2021 & \cite{195} & Verification & Custom CNN & FV-USM & ACC = 98.89\% \\
        \specialrule{0em}{4pt}{4pt}
        2021 & \cite{225} & Segmentation & VGG-16 & \makecell[l]{(Train) Private1 (IN:950) \\ (Test) Private2 (IN:150)} & JS = 0.926 \; MBE = 0.077 \; MAE = 16.46 \\
        \specialrule{0em}{4pt}{4pt}
        2021 & \cite{108} & Verification & ResNet & \makecell[l]{FV-USM \\ SDUMLA-HMT \\ HKPU \\ private (IN:8316 SN:1386)} & \makecell[l]{
        ACC = 99.99\% \; EER = 0.03\% (Closed-Set) \\ ACC = 99.79\% \; EER = 0.25\% (Open-Set) \\
        ACC = 99.56\% \; EER = 0.72\% (Closed-Set) \\ ACC = 99.25\% \; EER = 1.53\% (Open-Set) \\
        ACC = 99.60\% \; EER = 0.55\% (Closed-Set) \\ ACC = 99.07\% \; EER = 1.30\% (Open-Set) \\
        ACC = 99.20\% \; EER = 1.59\% (Closed-Set) \\ ACC = 97.90\% \; EER = 3.97\% (Open-Set)} \\
        \bottomrule
    \end{tabular}}
\end{sidewaystable}

\begin{sidewaystable}[htbp]
    \renewcommand{\thetable}{5}
    \caption{Continued: Summary of the FVR based on deep neural networks.}
    \centering
    \resizebox{\linewidth}{!}{\begin{tabular}{ccllll}
        \toprule
        \textbf{Year} & \textbf{Reference} & \textbf{Task} & \textbf{Network} & \textbf{Dataset} & \textbf{Performance} \\
        \specialrule{0em}{3pt}{3pt}
        \hline
        \specialrule{0em}{4pt}{4pt}
        2021 & \cite{110} & Verification & ResNet & FV-USM & ACC = 98.10\% \\
        \specialrule{0em}{4pt}{4pt}
        2021 & \cite{109} & Verification & ResNet & \makecell[l]{UTFVP \\ FV-USM \\ PALMAR \\ SDUMLA-HMT \\ THU-FVFDT1 \\ IDIAP \\ MMCBNU-6000 \\ HKPU} & \makecell[l]{AUC-ROC Score = 0.9992 \\ AUC-ROC Score = 1.0000 \\ AUC-ROC Score = 1.0000 \\ AUC-ROC Score = 1.0000 \\ AUC-ROC Score = 0.9993 \\ AUC-ROC Score = 0.9990 \\ AUC-ROC Score = 0.9997 \\ AUC-ROC Score = 1.0000} \\
        \specialrule{0em}{4pt}{4pt}
        2021 & \cite{233} & Template protection & CAE & UTFVP & \makecell[l]{FMR = 0.0\% \; FNMR = 0.0\% \; EER = 0.0\% (Nomal) \\ FMR = 18.7\% \; FNMR = 16.7\% \; EER = 17.7\%} \\
        \specialrule{0em}{4pt}{4pt}
        2021 & \cite{118} & Verification & \makecell[l]{SqNet \\ DenseNet} & \makecell[l]{PLUSVein-FV3 \\ PMMDB} & \makecell[l]{~ \\ ~} \\
        \specialrule{0em}{4pt}{4pt}
        2021 & \cite{252} & Feature extraction & Custom CNN & \makecell[l]{FV-USM \\ HKPU \\ SDUMLA-HMT \\ MMCBNU-6000 \\ THU-FVFDT3} & \makecell[l]{EER = 0.81\% \\ EER = 1.59\% \\ EER = 0.31\% \\ EER = 1.43\% \\ EER = 0.39\%} \\
        \specialrule{0em}{4pt}{4pt}
        2021 & \cite{207} & Image enhancement & GAN & \makecell[l]{(Train) Private1 (IN:680 IN:68)$\times$10 \\ (Test) Private2 (IN:600 SN:60)$\times$6 } & PSNR = 34.71dB \\
        \specialrule{0em}{4pt}{4pt}
        2021 & \cite{105} & Verification & ResNet & \makecell[l]{HKPU \\ MMCBNU-6000 \\ FV-USM} & \makecell[l]{EER = 1.90\% \\ EER = 0.21\% \\ EER = 0.48\%} \\
        \specialrule{0em}{4pt}{4pt}
        \bottomrule
    \end{tabular}}
\end{sidewaystable}

\begin{sidewaystable}[htbp]
    \renewcommand{\thetable}{5}
    \caption{Continued: Summary of the FVR based on deep neural networks.}
    \centering
    \resizebox{\linewidth}{!}{\begin{tabular}{ccllll}
        \toprule
        \textbf{Year} & \textbf{Reference} & \textbf{Task} & \textbf{Network} & \textbf{Dataset} & \textbf{Performance} \\
        \specialrule{0em}{3pt}{3pt}
        \hline
        \specialrule{0em}{4pt}{4pt}
        2021 & \cite{204} & Image enhancement & GAN & \makecell[l]{HKPU \\ SDUMLA-HMT} & \makecell[l]{EER = 1.65\% \\ EER = 3.17\%} \\
        \specialrule{0em}{4pt}{4pt}
        2021 & \cite{198} & Verification & Custom CNN & \makecell[l]{HKPU \\ FV-USM} & \makecell[l]{ACC = 92.11\% \\ ACC = 94.17\%} \\
        \specialrule{0em}{4pt}{4pt}
        2021 & \cite{232} & Template protection & CAE & UTFVP & \makecell[l]{FMR = 0.5\% \; FNMR = 0.7\% \; EER = 0.6\% (Nomal) \\ FMR = 8.5\% \; FNMR = 10.3\% \; EER = 9.4\%} \\
        \specialrule{0em}{4pt}{4pt}
        2021 & \cite{240} & M-Biometric recognition & Custom CNN & VeinECG & EER = 0.12\% (FLF) \; EER = 1.40\% (SLF) \\
        \specialrule{0em}{4pt}{4pt}
        2021 & \cite{156} & Verification & Lightweight CNN & \makecell[l]{HKPU \\ FV-USM \\ SDUMLA-HMT \\ UTFVP} & \makecell[l]{CIR = 96.98\% \\ CTR = 98.58\% \\ CIR = 97.75\% \\ 98.61\%} \\  
        \specialrule{0em}{4pt}{4pt}
        2021 & \cite{196} & Verification & Custom CNN & \makecell[l]{FV-USM \\ MMCBNU-6000} & \makecell[l]{ACC = 95.59\% \\ ACC = 97.54\%} \\
        \specialrule{0em}{4pt}{4pt}
        2021 & \cite{168} & Verification & Two-stream CNN & \makecell[l]{MMCBN-6000 \\ SDUMLA-HMT} & \makecell[l]{EER = 0.17\% \\ EER = 0.94\%} \\
        \specialrule{0em}{4pt}{4pt}
        2021 & \cite{171} & Verification & RDL & HKPU & ACC = 91.19\% \\
        \specialrule{0em}{4pt}{4pt}
        2021 & \cite{202} & Image enhancement & GAN & \makecell[l]{SDUMLA-HMT \\ HKPU} & \makecell[l]{PSNR = 32.64dB \; SNR = 22.75 \; SSIM = 0.85 \\ PSNR = 27.70dB \; SNR = 20.17 \; SSIM = 0.90} \\
        \bottomrule
    \end{tabular}}
\end{sidewaystable}

\begin{sidewaystable}[htbp]
    \renewcommand{\thetable}{5}
    \caption{Continued: Summary of the FVR based on deep neural networks.}
    \centering
    \resizebox{\linewidth}{!}{\begin{tabular}{ccllll}
        \toprule
        \textbf{Year} & \textbf{Reference} & \textbf{Task} & \textbf{Network} & \textbf{Dataset} & \textbf{Performance} \\
        \specialrule{0em}{3pt}{3pt}
        \hline
        \specialrule{0em}{4pt}{4pt}
        2022 & \cite{228} & PAD & DSCNN & \makecell[l]{IDIAP \\ SCUT} & \makecell[l]{APCER = 0.00\% \; BPCER = 0.00\% \; ACER = 0.00\% \\ ACPER = 0.00\% \; BPCER = 0.00\% \; ACER = 0.00\%} \\
        \specialrule{0em}{4pt}{4pt}
        2022 & \cite{121} & Verification & GAN & \makecell[l]{SDUMLA-HMT \\ FV-USM \\ HKPU \\ Private (IN: 8316 SN:1396)} & \makecell[l]{
        EER = 0.05\% (Closed-set) \; 1.33\% (Open-set) \\
        EER = 0.03\% (Closed-set) \; 0.14\% (Open-set) \\
        EER = 0.15\% (Closed-set) \; 0.40\% (Open-set) \\
        EER = 0.06\% (Closed-set) \; 1.34\% (Open-set)} \\
        \specialrule{0em}{4pt}{4pt}
        2022 & \cite{129} & Verification & DSCNN & \makecell[l]{SDUMLA-HMT \\ THU-FVFDT2} & \makecell[l]{ACC = 98.5\% \\ ACC = 90\%} \\
        \specialrule{0em}{4pt}{4pt}
        2022 & \cite{184} & Verification & Custom CNN & FV-USM & AAC =  98.53\% \\
        \specialrule{0em}{4pt}{4pt}
        2022 & \cite{132} & Verification & Custom CNN & \makecell[l]{SCUT \\ LFMB-3DPVFV} & \makecell[l]{EER = 2.61\% \\ EER = 2.81\%} \\
        \specialrule{0em}{4pt}{4pt}
        2022 & \cite{218} & Segmentation & \makecell[l]{Mask RCNN \\ CCNet \\ HRNet} & \makecell[l]{HKPU \\ PLUSVein \\ PMMDB-FR \\ PMMDB-FV3 \\ UTFVP} & \makecell[l]{
        IOU = 0.92 (Mask RCNN) \; 0.67 (CCNet) \; 0.73 (HRNet) \\
        IOU = 0.59 (Mask RCNN) \; 0.57 (CCNet) \; 0.75 (HRNet) \\
        IOU = 0.95 (Mask RCNN) \; 0.86 (CCNet) \; 0.87 (HRNet) \\
        IOU = 0.92 (Mask RCNN) \; 0.91 (CCNet) \; 0.84 (HRNet) \\
        IOU = 0.96 (Mask RCNN) \; 0.96 (CCNet) \; 0.98 (HRNet) \\
        IOU = 0.86 (Mask RCNN) \; 0.79 (CCNet) \; 0.83 (HRNet)} \\
        \specialrule{0em}{4pt}{4pt}
        2022 & \cite{122} & Verification & GAN & THU-FVFDT2 & \textbf{-} \\
        \specialrule{0em}{4pt}{4pt}
        2022 & \cite{200} & Image enhancement & GAN & Private & PSNR = 50.173dB (P-oil dirt) \; 32.978dB (P-finger molting) \\
        \bottomrule
    \end{tabular}}
\end{sidewaystable}

\section{Challenges and potential directions}\label{sec6}

Although the finger vein is stable and challenging in forgery due to the finger vein being deep inside the skin, the recognition process of FVR still faces some challenges. In this section, we summarize the challenges related to FVR and analyze existing research on these challenges. Meanwhile, the ANN technology represented by DL has been popular in recent years. In this section, we present the potential development directions of FVR based on cutting-edge knowledge in the field of ANN and the properties of finger veins as biological tissue. 

\subsection{Challenges of FVR}

\subsubsection{Finger posture changes}
During the finger vein image acquisition, the user may place the finger on the finger vein collector in different positions, which may result in misaligned images. Meanwhile, the skin's thickness on the finger's surface is hardly perfectly uniform. Some light angular rotations of the finger during the capture process can make the image quality change, and severe rotation of the finger resulting in postural changes can lead to significant differences in the captured vein structure from the registered images. \cite{170, 254} uses capsule networks for processing finger vein images to mitigate the effects of finger pose changes on FVR by taking advantage of the capsule network's ability to store information on feature location changes. However, the training time of the capsule network is too long, and too much redundant information will lead to the capsule shedding problem~\cite{278}. Moreover, the capsule network is still in development and not yet mature. \cite{188, 132} construct 3D images using multiple 2D finger vein images, and then use the constructed 3D finger vein images for recognition instead of traditional 2D images. The 3D finger vein images are shown in Fig.~\ref{CC1}. The 3D finger vein image can fully render the entire finger's vein structure instead of the localized structure captured on one side in the 2D image. The vein structure in the final 3D finger vein image is consistent regardless of finger posture variations. However, the 3D modeling process is computationally complex compared to the traditional FVR approach using 2D images. Meanwhile, the multiple 2D finger vein images used to construct the 3D model need to be captured from different angles using multiple NIR cameras. This method of capturing images is even more costly. There is also a lack of large public datasets of 3D finger veins for other network models to be trained. In conclusion, more cost-effective and convenient methods need to be developed to solve the challenge of image misalignment caused by finger posture changes in FVR.

\subsubsection{Illumination blur}
Illumination conditions can affect the finger vein image quality during the image acquisition phase~\cite{213}. Affected by uneven illumination condition, the brightness of the central part of the finger vein image is too high, and both sides are too low, resulting in the loss of some of the vein pattern information. Meanwhile, the obtained finger vein images are frequently damaged due to light attenuation in biological tissue~\cite{208}. To address the degradation phenomenon of image quality caused by uneven illumination, most research works employ cumbersome image pre-processing methods to improve image contrast. However, the restoration effect of this method is hardly significant when the image degradation is severe. To enhance the quality of finger vein images damaged by uneven illumination, \cite{204} restores the vein information damaged by light with the help of residual blocks. However, the robustness of this model is insufficient, and the recovery of uneven illumination is unsatisfactory for general scenes. To address this challenge, on the software side, the network model needs to improve robustness without increasing complexity as much as possible, and it also needs to optimize the image enhancement methods further. In terms of hardware, creating an image acquisition environment that is as free from light as possible is necessary.

\subsubsection{Dynamic recognition}
In a typical biometric recognition system, the user usually need some time to pick up biometric information from the body during the image acquisition process. It is undoubtedly more satisfying if the feature acquisition process can be completed without dwell. However, image capture while the user is in motion will cause image blurring, and these blurred images can affect the recognition performance of ANNs. \cite{202} uses a DeblurGAN based on an improved loss function to focus the motion blurred finger vein image restoration. The approach performs image enhancement, but the matching process still has failed cases because of the difficulty in ROI extraction of motion blurred finger vein images. To more thoroughly address the challenge of dynamic verification for FVR, \cite{145} uses an array of multiple cameras for image capture, and these cameras can collect multiple finger vein videos at different exposures during user movement. The LSTM and CNN are used to process the captured images. However, this system is designed with a non-end-to-end network structure, and the processing is complex. In addition, as with several other solutions to the challenge, the image acquisition method for this method is costly. FVR  still needs further exploration in dynamic recognition.

\subsubsection{Other challenges}
FVR faces several challenges that haphazardly interfere with its performance and application capabilities of FVR. Temperature changes may impact blood flow in the veins, affecting the contrast of the vein picture. Furthermore, various thicknesses of finger bones and muscles will produce irregular shading, which can detect ambiguous or incomplete vessel structures. Besides, although the finger vein information is claimed to be only detected on aliveness, many recognition systems can be cheated by forged vein patterns printed on distinctive paper. The excellent performance, privacy, and stability of FVR are already widely appreciated, and if researchers can overcome these above challenges, the FVR technology will be more powerful.

\subsection{Potential development direction of FVR}

\subsubsection{Knowledge distillation}
In addition to training network models, how to deploy the models on terminals with limited hardware conditions is a critical issue for FVR. In general, the trained and deployed models are usually the same in FVR tasks, and networks with excellent recognition performance in FVR tasks usually have complex structures. These complex network models are difficult to deploy on mobile terminals due to their slow inference speed and high computational resource requirements. To enable the network models with excellent performance to be deployed on mobile terminals to enhance the application of capabilities of FVR, the model compression methods are used. Traditional compression methods mainly focus on designing lightweight network structures. The lightweight networks can achieve comparable performance to complex networks on easy tasks through a refined design, but their structure is shallow, resulting in limited feature extraction capability that fails to solve the tasks that have large data volumes~\cite{268}. Knowledge distillation~\cite{267} can enable lightweight network models to have the better recognition performance comparable to that of deep networks, thus improving the application capability of FVR.

\begin{figure}[htbp]
    \centering
    \includegraphics[width = 1.0\textwidth]{ 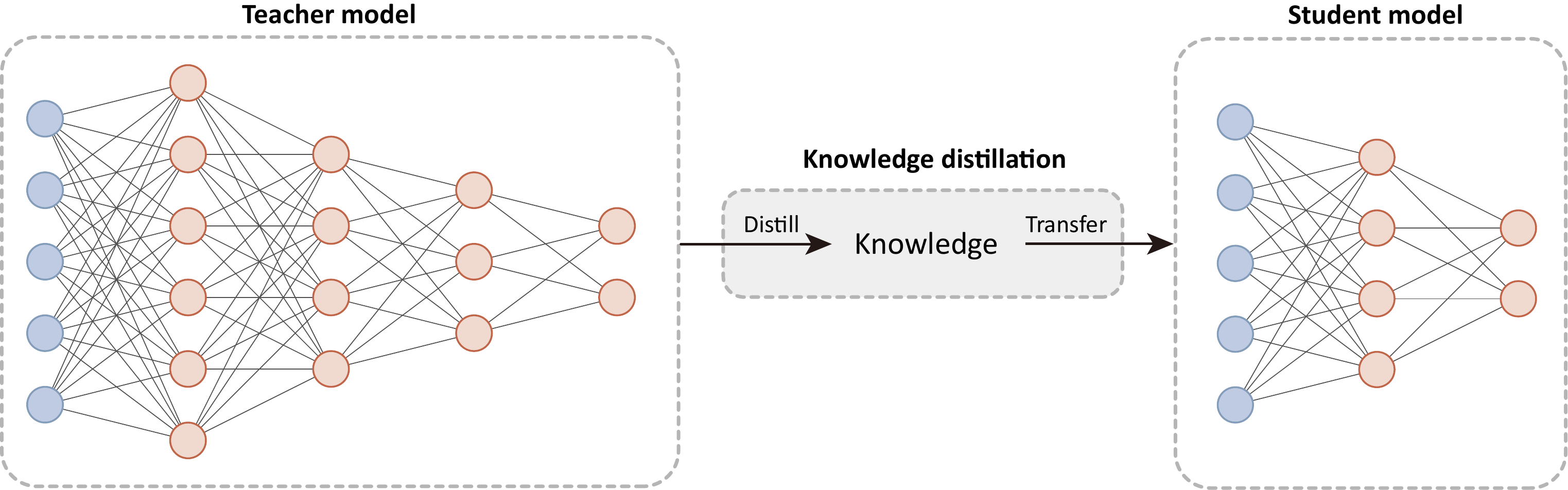}
    \caption{The knowledge distillation based on “teacher-student” network~\cite{267}.}
    \label{KD}
\end{figure}

Knowledge distillation is a model compression algorithm and employs a training method based on a “teacher-student” network. The schematic diagram is shown in Fig.~\ref{KD}. This approach uses the knowledge of larger networks with better performance to supervise the training process of the lightweight network to allow the lightweight network to achieve better performance. This large network is called the “teacher” model, and the lightweight model is called the “student” model. In the case of using knowledge distillation, the teacher model only serves as a guide, and it is the student model that is really deployed on terminals.

Knowledge distillation can be divided into response-based distillation and feature-based distillation according to the different distillation methods. In the classification task, the network uses a Softmax function at the final layer to obtain the probabilities of the positive and the negative samples. The response-based distillation uses the output of Softmax from the teacher model to assist the training of the student model, allowing the student model to learn the generalization ability of the teacher model by directly combining the labels of samples~\cite{270}. In this strategy, the output of Softmax is called soft target, and the true labels of samples are called hard target. Due to all data in the same distribution having similarities~\cite{91}, and negative samples with higher scores have more similarities to positive samples, in addition to positive samples, a large number of negative samples contain abundant information inferred from the teacher model. However, using the output of Softmax directly as soft target makes its contribution to the loss function minor. As shown in~\myref{EQKD}, the response-based distillation amplifies the effect of information carried by negative samples on model training by setting the parameter $T$ in the Softmax function. $p_i^T$ represents the value of the Softmax output of the teacher model on class $i$ at the temperature equal to $T$, and $q_i^T$ represents the student model. $N$ represents the total number of samples.

\begin{equation}\label{EQKD}
    \begin{array}{l}
p_i^T = \frac{{\exp ({v_i}/T)}}{{\sum\nolimits_k^N {\exp ({v_k}/T)} }}\\
\\
q_i^T = \frac{{\exp ({z_i}/T)}}{{\sum\nolimits_k^N {\exp ({z_k}/T)} }}
\end{array}
\end{equation}

\begin{figure}[htbp]
    \centering
    \includegraphics[width = 1.0\textwidth]{ 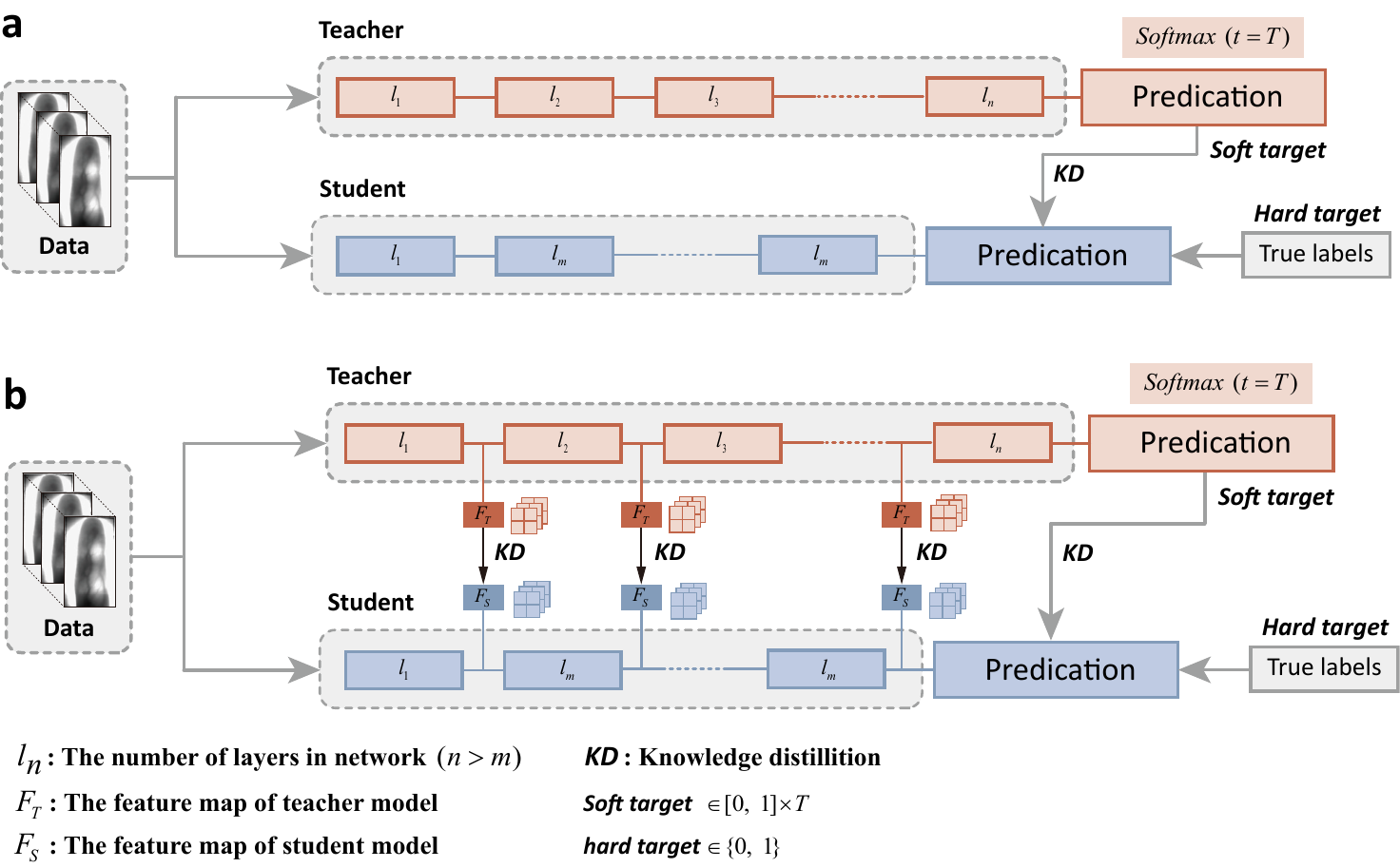}
    \caption{The response-based distillation and the feature-based distillation. \textbf{(a)} The response-based distillation. \textbf{(b)} The feature-based distillation.}
    \label{KD2}
\end{figure}

In addition to the output of the final layer in the network, the feature maps output from the middle layers of the network in the teacher model can also be the knowledge for supervising the training of the student model~\cite{271}. The feature-based distillation is an extension of response-based distillation, and the difference between the two methods is shown in Fig.~\ref{KD2}. The feature-based knowledge distillation leans not only on the output results in the teacher model but also on the features extracted from hidden layers in the teacher model.

According to the literature we surveyed, most of the model compression methods involved in the FVR are to design lightweight networks~\cite{150, 151, 152, 153, 154, 155, 156}, and knowledge distillation has never been widely used. Therefore, applying knowledge distillation in FVR is yet to be explored. The development of this technology will take FVR to a new level.

\subsubsection{Transformer}
\begin{figure}
    \centering
    \includegraphics[width = 0.5\textwidth]{ 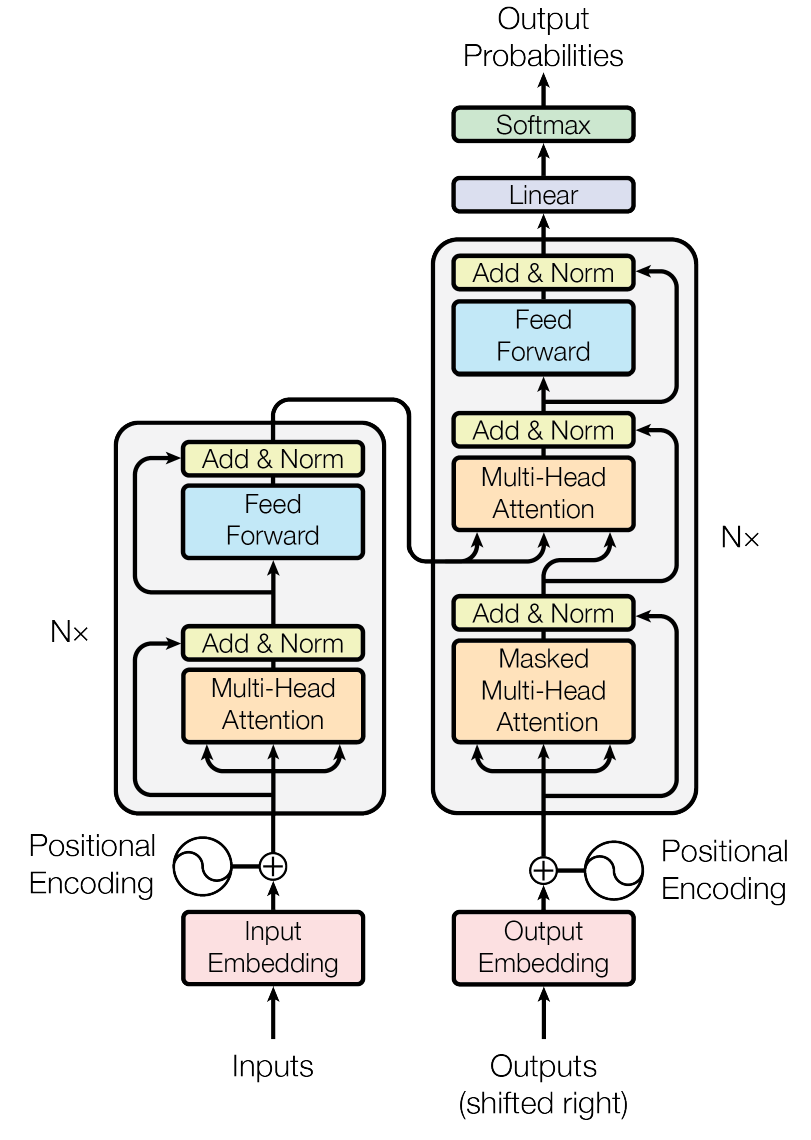}
    \caption{The Encoder-Decoder structure based on self-attention~\cite{272}.}
    \label{TF1}
\end{figure}

Transformer is a landmark model proposed by Google in 2017~\cite{272}, which has created a revolution in the field of \emph{Nature Language Process} (NLP). Transformer uses an Encoder-Decoder structure based on self-attention, and the specific structure is shown in Fig.~\ref{TF1}. In NLP tasks, RNNs process data in a serial manner. In contrast to RNNs, The innovation of Transformer processes data in a parallel manner since the attention mechanism allows the model to consider the interrelationship between any two words regardless of their position in the text sequence. Relying on this advantage, Transformer has achieved great success in NLP tasks~\cite{273, 274}. At the same time, several studies have pioneered the introduction of Transformer into computer vision with excellent performance, including image classification~\cite{275}, object detection~\cite{276}, image segmentation tasks~\cite{277}, etc. The \emph{Vision Transformer} (VIT) and the \emph{SEgmentation TRansformer} (SETR) are two representative Transformer-based models in the CV field, based on which Transformer can make a breakthrough in the verification task and segmentation task of FVR.

\subsubsection{The similarity of finger veins}
One of the conveniences of FVR is that even if one finger is in an accident, the other fingers can still be used for identification. In addition, it is undoubtedly excellent to lock the identity of a suspect in forensic identification with the veins of any one finger. Nevertheless, registering ten fingers simultaneously in an identification system is a hassle for users. Therefore, it is necessary to explore whether the finger veins of the ten fingers of the same individual are similar in future FVR research work. If there are some connections between different finger veins of the same person and they can be identified by FVR systems, it will take the convenience of FVR systems to a new level. Although~\cite{259, 260} focus on this problem, they still have some limitations. \cite{259} considers the connection between the veins of different fingers of the same person is too weak to perform the recognition. \cite{260} utilizes the triplet loss with hard triplet online mining for FVR. This strategy successfully verified that symmetric fingers (the same sort of finger from opposite hands in the same individual) have enough similarities to be recognized. The similarities of other asymmetric fingers are also proved in~\cite{260}, but the proposed recognition system can still not effectively identify these asymmetric finger veins. Therefore, related work can still be further explored in the future.

\section{Conclusion}\label{sec7}
This paper provides a comprehensive survey of the ANN-based FVR and compensates for the lack of a comprehensive survey related to ANN in the field of FVR. A total of 149 papers have been collected to support this work. The purpose of this paper is to discuss the FVR tasks based on ANNs. In this paper, some FVR-related information is first presented, including the background of FVR in Sec.~\ref{sec1}, representative network structures in Sec.~\ref{sec2}, and commonly used public datasets in Sec.~\ref{sec3}. The literature review section follows this paper's most important work. In Sec.~\ref{sec4}, we discuss the application of classical neural networks in FVR from the perspective of the biometric recognition process, including pre-processing, feature extraction, and matching. The classical neural networks are mainly used to perform the classification task in FVR. The MLP and ANFIS are the most widely used in these networks. In Sec.~\ref{sec5}, we discuss the application of deep neural networks in FVR from tasks of the papers. In these papers, the typical networks such as AlexNet, VGGNet, ResNet, etc., and some custom networks demonstrate the advantages of DL methods on FVR. The GAN and U-Net show their excellent performance on finger vein image enhancement tasks and segmentation tasks, respectively. In addition, the PAD and template protection tasks on FVR are researched, and these researches achieve outstanding performance. The end of this survey summarizes some typical challenges in FVR and suggests potential directions for FVR. This content can inspire future research in the field of FVR.

\section{Acknowledgements}
This work is supported by the “Scientific Research Project of Education Department of Hunan Province” (No. 21C0839). We also thank Mr. Shanjun Chen and Mr. Rui Pan for their important discussion in this work. Renye Zhang is the co-first author of this paper.

\bibliography{sn-bibliography}
\end{document}